\documentclass[11pt, a4paper, logo, onecolumn, nonumbering]{style_gdm/googledeepmind}

\pdftrailerid{redacted}


\definecolor{GoogleBlue}{HTML}{1A73E8}
\usepackage[colorlinks=true, allcolors=GoogleBlue]{hyperref}
\usepackage{kantlipsum, lipsum}
\usepackage{dsfont}
\usepackage{multirow}
\usepackage{array}
\usepackage{amssymb} %
\usepackage{pifont} %
\usepackage{makecell}
\usepackage{algorithm}
\usepackage{algpseudocode}
\usepackage{amsmath}
\usepackage{multicol}
\usepackage{float}
\usepackage{wrapfig}
\usepackage{siunitx}
\usepackage{tabularx}
\usepackage{booktabs}
\usepackage{enumitem}
\usepackage{etoolbox}

\usepackage{colortbl}
\definecolor{GoogleBlue}{HTML}{1A73E8}
\definecolor{GoogleBlueLight}{HTML}{e8f0fe}
\definecolor{GoogleBlueDark}{HTML}{174ea6}
\definecolor{GoogleRed}{HTML}{D93025}
\definecolor{GoogleGray}{HTML}{767676}
\definecolor{figtextcolor}{HTML}{333333}
\definecolor{perceptioncol}{HTML}{1A73E8}
\definecolor{modelingcol}{HTML}{925BCF}
\definecolor{manipulationcol}{HTML}{C440A7}
\definecolor{reasoningcol}{HTML}{DA2F77}
\definecolor{tablelightgray}{gray}{0.7} 

\usepackage{subcaption}
\usepackage{opensans}
\DeclareCaptionFont{subfigcapsize}{\fontsize{8pt}{8pt}\selectfont}  %
\newcommand{\nanobanana}{Gemini-3.1-Flash-Image-Preview}
\newcommand{\omnione}{Omni~1.1}
\newcommand{\geminiomnione}{Gemini \omnione{}}
\newcommand{\gptone}{GPT-1}
\newcommand{\gpttwo}{GPT-2}
\newcommand{\gptthree}{GPT-3}
\newcommand{\seedancetwofive}{Seedance~2.5}
\newcommand{\vpctsketch}{VPCT-sketch}
\newcommand{\vpctblender}{VPCT-blender}

\newcommand{\chromaticscrambleone}{\texttt{Chromatic scramble~v1}}
\newcommand{\chromaticscrambletwo}{\texttt{Chromatic scramble~v2}}
\newcommand{\chromaticscramblethree}{\texttt{Chromatic scramble~v3}}

\newcommand{\twofigwidth}{0.49}
\newcommand{\threefigwidth}{0.325}

\usepackage{xurl}

\usepackage[capitalize]{cleveref}
\crefname{section}{Sec.}{Secs.}  %
\crefname{subsection}{Sec.}{Secs.}  %
\crefname{subsubsection}{Sec.}{Secs.}  %
\crefname{table}{Tab.}{Tabs.}
\crefname{figure}{Fig.}{Figs.}
\crefname{appendix}{App.}{Apps.}  %
\crefname{subappendix}{App.}{Apps.}  %

\makeatletter
\@ifpackageloaded{cvpr}{%
  \newcommand{\website}[1]{}%
}{%
  \newcommand{\website}[1]{#1}%
}
\makeatother

\usepackage{graphicx}
\graphicspath{{figures/}}

\usepackage[breakable, skins]{tcolorbox}
\newcommand{\callout}[1]{%
    \begin{tcolorbox}[colframe=GoogleBlue, colback=GoogleBlueLight, coltext=GoogleBlue!30!black, width=\columnwidth, left=2mm, right=2mm, top=1mm, bottom=1mm, boxrule=.5mm, before skip=6pt, after skip=6pt]
        #1
    \end{tcolorbox}
}
\newcounter{takeaway}
\newcommand{\takeaway}[1]{%
    \stepcounter{takeaway}
    \callout{\textbf{Takeaway\;\thetakeaway}\hspace{1em}#1}
}

\makeatletter
\expandafter\let\csname iftok\endcsname\if
\AtBeginDocument{%
  \def\NAT@spacechar{~}%
  \def\NAT@space{~}%
  \def\NAT@cmt{,~}%
  \def\NAT@penalty{\penalty\@M}%
  \renewcommand\NAT@citenum[3]{%
    \ifNAT@swa
      \NAT@mbox{\NAT@@open\iftok*#2*\else#2\NAT@spacechar\fi}%
      #1\iftok*#3*\else\NAT@cmt#3\fi\NAT@@close
    \else
      #1%
    \fi\endgroup
  }%
  \let\@cite\NAT@citenum
  \g@addto@macro\NAT@set@cites{%
    \def\NAT@spacechar{~}%
    \def\NAT@space{~}%
    \def\NAT@cmt{,~}%
    \def\NAT@penalty{\penalty\@M}%
    \let\@cite\NAT@citenum
  }%
}
\makeatother

\newtcolorbox{promptbox}[1]{
    enhanced jigsaw,            %
    colback=gray!20,
    coltext=black!80,
    arc=5pt,
    boxrule=0pt,
    coltitle=black!80,
    fonttitle=\small\bfseries,
    colbacktitle=gray!30,
    breakable,                  %
    fontupper=\small\ttfamily,  %
    title={#1},
}

\usepackage{tikz}
\usetikzlibrary{calc, arrows.meta}
\usepackage{pdfrender}
\newlength{\tikzwidth}
\newlength{\boxpad}
\newlength{\boxgap}
\newlength{\boxtextheight}
\definecolor{plotspinecolor}{HTML}{222222}

\newcommand{\figfont}{\IfFileExists{T1DejaVuSans-TLF.fd}{\fontencoding{T1}\fontfamily{DejaVuSans-TLF}\selectfont}{\sffamily}}

\newcommand{\reftt}[1]{\scalebox{0.96}[1.22]{\fontfamily{lmtt}\fontseries{m}\upshape\selectfont #1}}

\newsavebox{\refttprotobox}
\newcommand{\reflabeltext}[2]{%
  {\fontsize{6.5pt}{7.8pt}\selectfont\reftt{#2}}%
}

\newcommand{\framebadge}[1]{%
  \mbox{%
    \figfont\mdseries\fontsize{5.6pt}{6.5pt}\selectfont
    \rlap{\kern-0.65pt\raisebox{0.00pt}{\textcolor{white}{#1}}}%
    \rlap{\kern 0.65pt\raisebox{0.00pt}{\textcolor{white}{#1}}}%
    \rlap{\kern 0.00pt\raisebox{-0.65pt}{\textcolor{white}{#1}}}%
    \rlap{\kern 0.00pt\raisebox{0.65pt}{\textcolor{white}{#1}}}%
    \rlap{\kern-0.46pt\raisebox{-0.46pt}{\textcolor{white}{#1}}}%
    \rlap{\kern 0.46pt\raisebox{-0.46pt}{\textcolor{white}{#1}}}%
    \rlap{\kern-0.46pt\raisebox{0.46pt}{\textcolor{white}{#1}}}%
    \rlap{\kern 0.46pt\raisebox{0.46pt}{\textcolor{white}{#1}}}%
    \rlap{\kern-0.60pt\raisebox{-0.25pt}{\textcolor{white}{#1}}}%
    \rlap{\kern 0.60pt\raisebox{-0.25pt}{\textcolor{white}{#1}}}%
    \rlap{\kern-0.60pt\raisebox{0.25pt}{\textcolor{white}{#1}}}%
    \rlap{\kern 0.60pt\raisebox{0.25pt}{\textcolor{white}{#1}}}%
    \rlap{\kern-0.25pt\raisebox{-0.60pt}{\textcolor{white}{#1}}}%
    \rlap{\kern 0.25pt\raisebox{-0.60pt}{\textcolor{white}{#1}}}%
    \rlap{\kern-0.25pt\raisebox{0.60pt}{\textcolor{white}{#1}}}%
    \rlap{\kern 0.25pt\raisebox{0.60pt}{\textcolor{white}{#1}}}%
    \rlap{\kern-0.32pt\raisebox{0.00pt}{\textcolor{white}{#1}}}%
    \rlap{\kern 0.32pt\raisebox{0.00pt}{\textcolor{white}{#1}}}%
    \rlap{\kern 0.00pt\raisebox{-0.32pt}{\textcolor{white}{#1}}}%
    \rlap{\kern 0.00pt\raisebox{0.32pt}{\textcolor{white}{#1}}}%
    \textcolor{plotspinecolor}{#1}%
  }%
}

\tikzset{
  frameborder/.style={draw=plotspinecolor, line width=0.45pt},
  paneltitle/.style={
    anchor=south west,
    inner sep=0pt,
    outer sep=0pt,
    font=\figfont\fontsize{8.2pt}{9.8pt}\selectfont,
    text=plotspinecolor
  },
  promptlabel/.style={
    anchor=south west,
    inner sep=0pt,
    outer sep=0pt,
    font=\figfont\itshape\fontsize{6.5pt}{7.8pt}\selectfont,
    text=plotspinecolor
  },
  promptlabeltop/.style={
    anchor=north west,
    inner sep=0pt,
    outer sep=0pt,
    align=justify,
    font=\figfont\itshape\fontsize{6.5pt}{7.8pt}\selectfont,
    text=plotspinecolor
  },
  reflabel/.style={
    anchor=north,
    inner sep=0pt,
    outer sep=0pt,
    font=\fontsize{6.5pt}{7.8pt}\selectfont,
    text=plotspinecolor
  },
  badgelabel/.style={
    anchor=north west,
    inner sep=0pt,
    outer sep=0pt
  }
}

\newcommand{\frameboxsixteennine}[4]{%
  \begin{scope}%
    \clip (#1) rectangle ++(#2, -#3);%
    \fill[black] (#1) rectangle ++(#2, -#3);%
    \node[inner sep=0pt, outer sep=0pt] at ($(#1) + (0.5*#2, -0.5*#3)$) {%
      \includegraphics[width=#2, height=#3, keepaspectratio]{#4}%
    };%
  \end{scope}%
  \draw[frameborder] (#1) rectangle ++(#2, -#3);%
}

\newcommand{\frameboxcover}[4]{%
  \begin{scope}%
    \clip (#1) rectangle ++(#2, -#3);%
    \fill[black] (#1) rectangle ++(#2, -#3);%
    \node[inner sep=0pt, outer sep=0pt] at ($(#1) + (0.5*#2, -0.5*#3)$) {%
      \includegraphics[height=#3]{#4}%
    };%
  \end{scope}%
  \draw[frameborder] (#1) rectangle ++(#2, -#3);%
}

\newcommand{\drawrefvideofilmstrip}[6]{%
  \pgfmathsetmacro{\vrailw}{3.2}%
  \pgfmathsetmacro{\vimgw}{#3 - 2.0*\vrailw}%
  \pgfmathsetmacro{\vimgh}{#3 * 9.0 / 16.0}%
  \fill[plotspinecolor] (#1, #2) rectangle ++(#3, -\vimgh);%
  \pgfmathsetmacro{\vimgx}{#1 + \vrailw}%
  \frameboxcover{\vimgx, #2}{\vimgw pt}{\vimgh pt}{#4}%

  \pgfmathsetmacro{\vholew}{1.5}%
  \pgfmathsetmacro{\vholeh}{3.1}%
  \pgfmathsetmacro{\vstep}{\vimgh / 6.0}%
  \foreach \k in {1,...,6} {%
    \pgfmathsetmacro{\vholey}{#2 - (\k - 0.5)*\vstep + 0.5*\vholeh}%
    \pgfmathsetmacro{\vlholex}{#1 + 0.5*(\vrailw - \vholew)}%
    \pgfmathsetmacro{\vrholex}{#1 + #3 - 0.5*(\vrailw + \vholew)}%
    \fill[white, rounded corners=0.4pt] (\vlholex, \vholey) rectangle ++(\vholew, -\vholeh);%
    \fill[white, rounded corners=0.4pt] (\vrholex, \vholey) rectangle ++(\vholew, -\vholeh);%
  }%
  \draw[frameborder] (#1, #2) rectangle ++(#3, -\vimgh);%
  \node[reflabel] at (#1 + 0.5*#3, #2 - \vimgh - 1.6) {\reflabeltext{#6}{#5}};%
}

\newsavebox{\refimgbox}
\newcommand{\drawrefimagefit}[7]{%
  \sbox{\refimgbox}{\includegraphics[width=#3 pt, height=#7 pt, keepaspectratio]{#4}}%
  \pgfmathsetmacro{\refimgw}{\wd\refimgbox}%
  \pgfmathsetmacro{\refimgh}{\ht\refimgbox + \dp\refimgbox}%
  \pgfmathsetmacro{\refleftx}{#1 + 0.5*(#3 - \refimgw)}%
  \pgfmathsetmacro{\reftopy}{#2 + 0.5*\refimgh}%
  \pgfmathsetmacro{\dogear}{0.22*min(\refimgw*9.0/16.0, \refimgh)}%
  \begin{scope}%
    \clip (\refleftx, \reftopy) -- ++(\refimgw - \dogear, 0) -- ++(\dogear, -\dogear) -- ++(0, -\refimgh + \dogear) -- ++(-\refimgw, 0) -- cycle;%
    \fill[white] (\refleftx, \reftopy) rectangle ++(\refimgw, -\refimgh);%
    \node[inner sep=0pt, outer sep=0pt] at (\refleftx + 0.5*\refimgw, #2) {\usebox{\refimgbox}};%
  \end{scope}%
  \draw[frameborder, line join=round] (\refleftx, \reftopy) -- ++(\refimgw - \dogear, 0) -- ++(\dogear, -\dogear) -- ++(0, -\refimgh + \dogear) -- ++(-\refimgw, 0) -- cycle;%
  \filldraw[frameborder, fill=gray!25, line join=round] (\refleftx + \refimgw - \dogear, \reftopy) -- ++(0, -\dogear) -- ++(\dogear, 0) -- cycle;%
  \node[reflabel] at (\refleftx + 0.5*\refimgw, \reftopy - \refimgh - 1.6) {\reflabeltext{#6}{#5}};%
}

\newcommand{\drawvideofilmstrip}[5]{%
  \pgfmathsetmacro{\vrailw}{3.6}%
  \pgfmathsetmacro{\vimgw}{#3 - 2.0*\vrailw}%
  \pgfmathsetmacro{\vimgh}{\vimgw * 9.0 / 16.0}%
  \fill[plotspinecolor] (#1, #2) rectangle ++(#3, -\vimgh);%
  \pgfmathsetmacro{\vimgx}{#1 + \vrailw}%
  \frameboxsixteennine{\vimgx, #2}{\vimgw pt}{\vimgh pt}{#4}%

  \pgfmathsetmacro{\vholew}{1.7}%
  \pgfmathsetmacro{\vholeh}{3.5}%
  \pgfmathsetmacro{\vstep}{\vimgh / 6.0}%
  \foreach \k in {1,...,6} {%
    \pgfmathsetmacro{\vholey}{#2 - (\k - 0.5)*\vstep + 0.5*\vholeh}%
    \pgfmathsetmacro{\vlholex}{#1 + 0.5*(\vrailw - \vholew)}%
    \pgfmathsetmacro{\vrholex}{#1 + #3 - 0.5*(\vrailw + \vholew)}%
    \fill[white, rounded corners=0.45pt] (\vlholex, \vholey) rectangle ++(\vholew, -\vholeh);%
    \fill[white, rounded corners=0.45pt] (\vrholex, \vholey) rectangle ++(\vholew, -\vholeh);%
  }%
  \draw[frameborder] (#1, #2) rectangle ++(#3, -\vimgh);%
  \node[badgelabel] at (\vimgx + 2.0, #2 - 2.0) {\framebadge{#5}};%
}

\newsavebox{\iclpromptbox}
\newcommand{\drawsinglelineprompt}[3]{%
  \node[promptlabel] at (#1, #2) {%
    \sbox{\iclpromptbox}{\figfont\itshape\fontsize{6.5pt}{7.8pt}\selectfont #3}%
    \pgfmathsetmacro{\promptw}{\wd\iclpromptbox}%
    \pgfmathsetmacro{\promptxscale}{min(1.0, \halfw / \promptw)}%
    \scalebox{\promptxscale}[1.0]{\usebox{\iclpromptbox}}%
  };%
}

\newcommand{\drawiclgroup}[6]{%
  \drawsinglelineprompt{#1}{#2 + 3.0}{#3}%
  \pgfmathsetmacro{\refimghmax}{\refcolw * 9.0 / 16.0}%
  \pgfmathsetmacro{\refcy}{#2 - 0.5*\refimghmax}%
  \drawrefimagefit{#1}{\refcy}{\refcolw}{#4}{@image1}{44.43}{\refimghmax}%
  \drawvideofilmstrip{#1 + \refcolw + \refgap}{#2}{\vidslotw}{#5}{Query}%
  \drawvideofilmstrip{#1 + \refcolw + \refgap + \vidslotw + \vidgap}{#2}{\vidslotw}{#6}{Output}%
}

\newcommand{\drawiclgroupvideoref}[6]{%
  \drawsinglelineprompt{#1}{#2 + 3.0}{#3}%
  \drawrefvideofilmstrip{#1}{#2}{\refcolw}{#4}{@video1}{44.43}%
  \drawvideofilmstrip{#1 + \refcolw + \refgap}{#2}{\vidslotw}{#5}{Query}%
  \drawvideofilmstrip{#1 + \refcolw + \refgap + \vidslotw + \vidgap}{#2}{\vidslotw}{#6}{Output}%
}

\newcommand{\drawquantblock}[3]{%
  \node[promptlabeltop, text width=\halfw pt] (qprompt) at (#1, #2) {%
    #3%
  };%
  \path (qprompt.south west);
  \pgfgetlastxy{\qpx}{\qpy}%
  \pgfmathsetmacro{\qvisy}{\qpy - 4.2}%
  \pgfmathsetmacro{\minirefhgap}{\refgap}%
  \pgfmathsetmacro{\minirefw}{(\refcolw - \minirefhgap) / 2.0}%
  \pgfmathsetmacro{\minirefh}{\minirefw * 9.0 / 16.0}%
  \pgfmathsetmacro{\minirefvstep}{\vidboxh - \minirefh}%
  \pgfmathsetmacro{\toprefcy}{\qvisy - 0.5*\minirefh}%
  \pgfmathsetmacro{\botrefcy}{\toprefcy - \minirefvstep}%
  \drawrefimagefit{#1}{\toprefcy}{\minirefw}{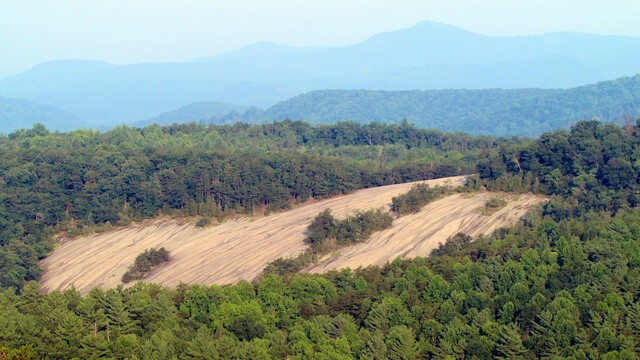}{@image1}{44.43}{\minirefh}%
  \drawrefimagefit{#1 + \minirefw + \minirefhgap}{\toprefcy}{\minirefw}{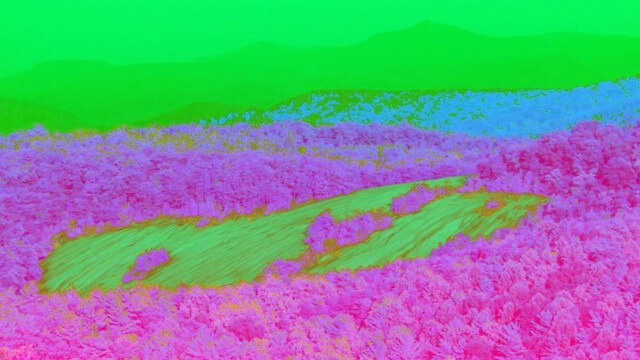}{@image2}{44.43}{\minirefh}%
  \drawrefimagefit{#1}{\botrefcy}{\minirefw}{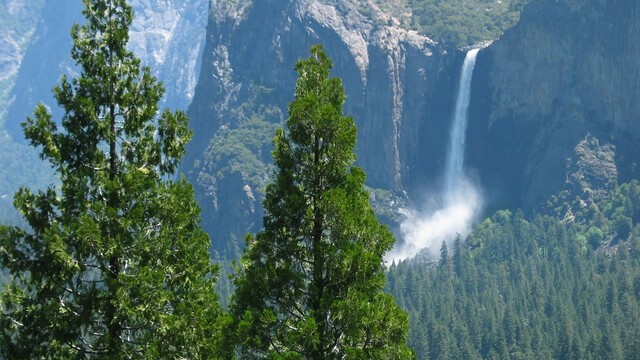}{@image3}{44.43}{\minirefh}%
  \drawrefimagefit{#1 + \minirefw + \minirefhgap}{\botrefcy}{\minirefw}{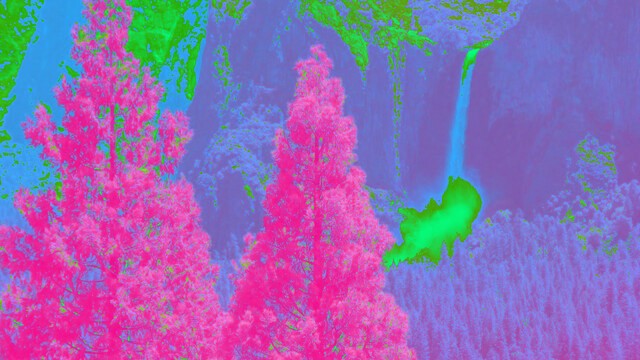}{@image4}{44.43}{\minirefh}%
  \drawvideofilmstrip{#1 + \refcolw + \refgap}{\qvisy}{\vidslotw}{\quantinframe}{Query}%
  \drawvideofilmstrip{#1 + \refcolw + \refgap + \vidslotw + \vidgap}{\qvisy}{\vidslotw}{\quantoutframe}{Output}%
  \pgfmathsetmacro{\quantblockboty}{\qvisy - \vidboxh - 6.8}%
}

\definecolor{markercolor}{HTML}{9C356D}
\definecolor{markeredge}{HTML}{1A1A1A}

\newcommand{\taskillustration}[4]{%
  \begin{tikzpicture}[x=1pt, y=1pt, baseline=0pt]
    \pgfmathsetmacro{\pw}{69.43}%
    \pgfmathsetmacro{\vrailw}{3.6}%
    \pgfmathsetmacro{\vimgw}{\pw - 2.0*\vrailw}%
    \pgfmathsetmacro{\ph}{\vimgw * 9.0 / 16.0}%
    \pgfmathsetmacro{\pgap}{4.0}%
    \node[anchor=west, inner sep=0pt, outer sep=0pt, text width=67.4pt, align=left,
          font=\figfont\itshape\fontsize{6.5pt}{7.4pt}\selectfont, text=plotspinecolor]
      at (0, 0.5*\ph) {#1};%
    \pgfmathsetmacro{\xquery}{\pw + \pgap}%
    \ifnum#4=1\relax
      \drawvideofilmstrip{\xquery}{\ph}{\pw}{#2}{Query}%
    \else
      \pgfmathsetmacro{\qimgx}{\xquery + \vrailw}%
      \frameboxsixteennine{\qimgx, \ph}{\vimgw pt}{\ph pt}{#2}%
      \node[badgelabel] at (\qimgx + 2.0, \ph - 2.0) {\framebadge{Query}};%
    \fi
    \pgfmathsetmacro{\xright}{2.0 * (\pw + \pgap)}%
    \drawvideofilmstrip{\xright}{\ph}{\pw}{#3}{Output}%
  \end{tikzpicture}%
}

\DeclareRobustCommand{\omnilinemarker}{%
  \tikz[baseline=-0.76ex]{
    \draw[draw=markercolor, line width=0.85pt, dash pattern=on 1.9pt off 1.1pt] (-5.8pt,0) -- (5.8pt,0);
    \filldraw[fill=markercolor, draw=markeredge, line width=0.55pt] (0,0) circle (2.2pt);
  }%
}
\DeclareRobustCommand{\seedancelinemarker}{%
  \tikz[baseline=-0.76ex]{
    \draw[draw=markercolor, line width=0.85pt, dash pattern=on 1.9pt off 1.1pt] (-5.8pt,0) -- (5.8pt,0);
    \filldraw[fill=white, draw=markercolor, line width=0.95pt, rounded corners=0.85pt] (-2.1pt,-2.1pt) rectangle (2.1pt,2.1pt);
  }%
}

\robustify\bfseries
\newcolumntype{R}{S[table-format = +2.1, retain-explicit-plus, detect-weight = true, detect-family = true, mode = text]}

\colorlet{shotpos}{green!70!black}
\colorlet{shotneg}{red!75!black}

\ExplSyntaxOn
\fp_const:Nn \c_shot_max_intensity_fp { 30 }

\fp_new:N \g_shot_curr_min_fp
\fp_new:N \g_shot_curr_max_fp
\fp_new:N \g_shot_table_min_fp
\fp_new:N \g_shot_table_max_fp
\fp_gset:Nn \g_shot_table_min_fp { -1e-9 }
\fp_gset:Nn \g_shot_table_max_fp {  1e-9 }

\cs_new_protected:Npn \shotsetminmax #1#2 {
  \fp_gset:Nn \g_shot_table_min_fp { \fp_min:nn {#1} { -1e-9 } }
  \fp_gset:Nn \g_shot_table_max_fp { \fp_max:nn {#2} {  1e-9 } }
}

\cs_new_protected:Npn \shottrack #1 {
  \tl_if_blank:nF {#1} {
    \fp_gset:Nn \g_shot_curr_min_fp  { \fp_min:nn { \g_shot_curr_min_fp }  {#1} }
    \fp_gset:Nn \g_shot_curr_max_fp  { \fp_max:nn { \g_shot_curr_max_fp }  {#1} }
    \fp_gset:Nn \g_shot_table_min_fp { \fp_min:nn { \g_shot_table_min_fp } {#1} }
    \fp_gset:Nn \g_shot_table_max_fp { \fp_max:nn { \g_shot_table_max_fp } {#1} }
  }
}

\fp_new:N \g_shot_white_fp
\fp_new:N \g_shot_green_fp

\cs_new_protected:Npn \shotcolormap #1#2 {
  \fp_gset:Nn \g_shot_white_fp {#1}
  \fp_gset:Nn \g_shot_green_fp {#2}
}

\cs_new:Npn \shotnumcell #1 {
  \fp_compare:nNnTF { (#1 - \g_shot_white_fp) / (\g_shot_green_fp - \g_shot_white_fp) } > { 0 }
    {
      \tex_unskip:D \cellcolor {
        shotpos ! \fp_eval:n {
          round(
            \fp_min:nn { \c_shot_max_intensity_fp } {
              \c_shot_max_intensity_fp * (#1 - \g_shot_white_fp) / (\g_shot_green_fp - \g_shot_white_fp)
            }, 1
          )
        }
      } #1
    }
    { #1 }
}

\cs_new:Npn \shotoldnumcell #1#2 {
  \fp_compare:nNnTF {#2} = { 0 }
    { #1 #2 }
    {
      \fp_compare:nNnTF {#2} > { 0 }
        {
          \tex_unskip:D \cellcolor {
            shotpos ! \fp_eval:n {
              round(
                \fp_min:nn { \c_shot_max_intensity_fp } {
                  \c_shot_max_intensity_fp * (#2) / \g_shot_table_max_fp
                }, 1
              )
            }
          } #1 #2
        }
        {
          \tex_unskip:D \cellcolor {
            shotneg ! \fp_eval:n {
              round(
                \fp_min:nn { \c_shot_max_intensity_fp } {
                  \c_shot_max_intensity_fp * (#2) / \g_shot_table_min_fp
                }, 1
              )
            }
          } #1 #2
        }
    }
}

\cs_new:Npn \shotsaveminmax {
  \token_to_str:N \shotsetminmax
    { \fp_to_tl:N \g_shot_curr_min_fp }
    { \fp_to_tl:N \g_shot_curr_max_fp }
}
\ExplSyntaxOff

\definecolor{tabhlcolor}{HTML}{1F2328}
\newcommand{\tabrowmark}[1]{\noalign{\tikz[remember picture,overlay]\coordinate (#1);}}
\newcommand{\tabrowmarktwo}[2]{\noalign{\tikz[remember picture,overlay]{\coordinate (#1);\coordinate (#2);}}}
\newcommand{\tabcolmarks}[1]{%
  \noalign{\vskip-\arraystretch\dimexpr\ht\strutbox+\dp\strutbox+\extrarowheight\relax}%
  & &
    \multicolumn{1}{@{}l@{}}{\tikz[remember picture,overlay]\coordinate(#1g1L);} & & \multicolumn{1}{@{}r@{}}{\tikz[remember picture,overlay]\coordinate(#1g1R);} &
    \multicolumn{1}{@{}l@{}}{\tikz[remember picture,overlay]\coordinate(#1g2L);} & & \multicolumn{1}{@{}r@{}}{\tikz[remember picture,overlay]\coordinate(#1g2R);} &
    \multicolumn{1}{@{}l@{}}{\tikz[remember picture,overlay]\coordinate(#1g3L);} & & \multicolumn{1}{@{}r@{}}{\tikz[remember picture,overlay]\coordinate(#1g3R);} &
    \multicolumn{1}{@{}l@{}}{\tikz[remember picture,overlay]\coordinate(#1g4L);} & & \multicolumn{1}{@{}r@{}}{\tikz[remember picture,overlay]\coordinate(#1g4R);} \\%
}

\newlength{\taboneleftw}
\newlength{\tabonecolsep}
\newlength{\taboneacrossep}
\newlength{\tabhlxshift}
\makeatletter
\@ifpackageloaded{cvpr}{%
  \setlength{\taboneleftw}{364.0pt}%
  \setlength{\tabonecolsep}{1.00pt}%
  \setlength{\taboneacrossep}{5.0pt}%
  \setlength{\tabhlxshift}{1.8pt}%
  \newcommand{\tabonebodypt}{6.5pt}%
  \newcommand{\tabonesubpt}{5.4pt}%
  \newcommand{\tabonesubskip}{6.5pt}%
}{%
  \setlength{\taboneleftw}{344.0pt}%
  \setlength{\tabonecolsep}{0.85pt}%
  \setlength{\taboneacrossep}{4.0pt}%
  \setlength{\tabhlxshift}{1.3pt}%
  \newcommand{\tabonebodypt}{6.15pt}%
  \newcommand{\tabonesubpt}{5.1pt}%
  \newcommand{\tabonesubskip}{6.2pt}%
}
\makeatother
\newcommand{\tabonesubfont}{\fontsize{\tabonesubpt}{\tabonesubskip}\selectfont}

\newcommand{\hlcellgroup}[4]{%
  \begin{tikzpicture}[remember picture, overlay]
    \coordinate (NW) at ($(#1g#2L |- #1r#3T)+(\tabhlxshift,-0.65pt)$);
    \coordinate (SE) at ($(#1g#2R |- #1r#3B)+(0,0.65pt)$);
    \coordinate (SW) at (NW |- SE);
    \draw[white, line width=0.90pt, line join=round] ($(NW)+(-1.8pt,0)$) rectangle (SE);
    \fill[tabhlcolor] ($(NW)+(3.4pt,0)$) {[rounded corners=1.1pt] -- ($(NW)+(-1.8pt,0)$) -- ($(SW)+(-1.8pt,0)$)} -- ($(SW)+(3.4pt,0)$) -- cycle;
    \draw[tabhlcolor, line width=0.75pt, rounded corners=1.1pt] ($(NW)+(-1.8pt,0)$) rectangle (SE);
    \node[inner sep=0pt, outer sep=0pt, text=white, font=\figfont\bfseries\fontsize{5.8pt}{5.8pt}\selectfont]
      at ($0.5*(NW)+0.5*(SW)+(0.8pt, 0)$) {#4};
  \end{tikzpicture}%
}

\DeclareRobustCommand{\tabbadge}[1]{%
  \smash{\tikz[baseline=-2.4pt]{%
    \node[fill=tabhlcolor, text=white, font=\figfont\bfseries\fontsize{6.0pt}{6.0pt}\selectfont,
          rounded corners=1.1pt, inner xsep=1.8pt, inner ysep=1.2pt, minimum height=7.2pt] {#1};%
  }}%
}

\DeclareRobustCommand{\taboutlined}[1]{%
  \smash{\tikz[baseline=(X.base)]{%
    \node[draw=tabhlcolor, line width=0.75pt, rounded corners=1.1pt,
          inner xsep=1.6pt, inner ysep=1.1pt, outer sep=0pt] (X) {#1};%
  }}%
}

\providecommand{\tabonefontvariant}{3}
\newcommand{\tabfigfont}{\figfont}

\newsavebox{\tableftcolbox}
\newsavebox{\tabcheaderbox}
\newdimen\tabaheaderdp

\newcommand{\subtabheader}[4]{%
  \sbox{\tabcheaderbox}{\parbox[t]{\linewidth}{%
    \tabfigfont\fontsize{6.5pt}{7.9pt}\selectfont\color{plotspinecolor}%
    {\phantomsubcaption\label{#1}}%
    \strut\textbf{#2}\hspace{0.55em}\textbf{#3}#4\strut\par\vspace{0pt}%
  }}%
  \ifstrequal{#2}{A}{\global\tabaheaderdp=\dp\tabcheaderbox}{}%
  \par\noindent\usebox{\tabcheaderbox}\par
}

\newcommand{\tabonebodyfont}{%
  \ifnum\tabonefontvariant=3\relax
    \IfFileExists{T1DejaVuSans-TLF.fd}{\def\sfdefault{DejaVuSans-TLF}}{\def\sfdefault{phv}}%
    \figfont\fontsize{\tabonebodypt}{9.5pt}\selectfont\color{plotspinecolor}%
    \def\scriptsize{\figfont\tabonesubfont}%
  \fi
}

\newcommand{\tabcpregap}{\nointerlineskip\vspace{\dimexpr\tabaheaderdp-\dp\tabcheaderbox+4.0pt\relax}}
\newcommand{\tabpanelcheight}{\the\dimexpr\dp\tableftcolbox-\tabaheaderdp-4.0pt\relax}

\definecolor{inpaintyellow}{HTML}{FFEE00}

\newcommand{\drawgridvideofilmstrip}[7]{%
  \pgfmathsetmacro{\vrailw}{3.6}%
  \pgfmathsetmacro{\vimgw}{#3 - 2.0*\vrailw}%
  \pgfmathsetmacro{\vimgh}{\vimgw * 9.0 / 16.0}%
  \pgfmathsetmacro{\cellw}{0.5 * \vimgw}%
  \pgfmathsetmacro{\cellh}{0.5 * \vimgh}%
  \pgfmathsetmacro{\vimgx}{#1 + \vrailw}%
  \fill[plotspinecolor] (#1, #2) rectangle ++(#3, -\vimgh);%
  \frameboxsixteennine{\vimgx, #2}{\cellw pt}{\cellh pt}{#4}%
  \frameboxsixteennine{\vimgx + \cellw, #2}{\cellw pt}{\cellh pt}{#5}%
  \fill[black] (\vimgx, #2 - \cellh) rectangle ++(\cellw, -\cellh);%
  \frameboxsixteennine{\vimgx + \cellw, #2 - \cellh}{\cellw pt}{\cellh pt}{#6}%
  \draw[draw=black, line width=0.45pt] (\vimgx + \cellw, #2) -- ++(0, -\vimgh);%
  \draw[draw=black, line width=0.45pt] (\vimgx, #2 - \cellh) -- ++(\vimgw, 0);%
  \pgfmathsetmacro{\vholew}{1.7}%
  \pgfmathsetmacro{\vholeh}{3.5}%
  \pgfmathsetmacro{\vstep}{\vimgh / 8.0}%
  \foreach \k in {1,...,8} {%
    \pgfmathsetmacro{\vholey}{#2 - (\k - 0.5)*\vstep + 0.5*\vholeh}%
    \pgfmathsetmacro{\vlholex}{#1 + 0.5*(\vrailw - \vholew)}%
    \pgfmathsetmacro{\vrholex}{#1 + #3 - 0.5*(\vrailw + \vholew)}%
    \fill[white, rounded corners=0.45pt] (\vlholex, \vholey) rectangle ++(\vholew, -\vholeh);%
    \fill[white, rounded corners=0.45pt] (\vrholex, \vholey) rectangle ++(\vholew, -\vholeh);%
  }%
  \draw[frameborder] (#1, #2) rectangle ++(#3, -\vimgh);%
  \node[reflabel] at (#1 + 0.5*#3, #2 - \vimgh - 1.6) {\reflabeltext{44.43}{#7}};%
}

\newcommand{\drawrefimagepair}[6]{%
  \pgfmathsetmacro{\halfpairw}{0.5 * #3}%
  \pgfmathsetmacro{\pairh}{\halfpairw * 9.0 / 16.0}%
  \pgfmathsetmacro{\dogear}{0.24 * \pairh}%
  \begin{scope}%
    \clip (#1, #2) -- ++(#3 - \dogear, 0) -- ++(\dogear, -\dogear) -- ++(0, -\pairh + \dogear) -- ++(-#3, 0) -- cycle;%
    \fill[black] (#1, #2) rectangle ++(#3, -\pairh);%
    \node[inner sep=0pt, outer sep=0pt] at (#1 + 0.5*\halfpairw, #2 - 0.5*\pairh) {%
      \includegraphics[width=\halfpairw pt, height=\pairh pt, keepaspectratio]{#4}%
    };%
    \node[inner sep=0pt, outer sep=0pt] at (#1 + 1.5*\halfpairw, #2 - 0.5*\pairh) {%
      \includegraphics[width=\halfpairw pt, height=\pairh pt, keepaspectratio]{#5}%
    };%
    \draw[draw=black, line width=0.45pt] (#1 + \halfpairw, #2) -- ++(0, -\pairh);%
  \end{scope}%
  \draw[frameborder, line join=round] (#1, #2) -- ++(#3 - \dogear, 0) -- ++(\dogear, -\dogear) -- ++(0, -\pairh + \dogear) -- ++(-#3, 0) -- cycle;%
  \filldraw[frameborder, fill=gray!25, line join=round] (#1 + #3 - \dogear, #2) -- ++(0, -\dogear) -- ++(\dogear, 0) -- cycle;%
  \node[reflabel] at (#1 + 0.5*#3, #2 - \pairh - 1.6) {\reflabeltext{44.43}{#6}};%
}

\newcommand{\drawspatialinpaintvideo}[5]{%
  \pgfmathsetmacro{\vrailw}{3.2}%
  \pgfmathsetmacro{\vimgw}{#3 - 2.0*\vrailw}%
  \pgfmathsetmacro{\halfvidw}{0.5 * \vimgw}%
  \pgfmathsetmacro{\vimgh}{\halfvidw * 9.0 / 16.0}%
  \pgfmathsetmacro{\vimgx}{#1 + \vrailw}%
  \fill[plotspinecolor] (#1, #2) rectangle ++(#3, -\vimgh);%
  \frameboxsixteennine{\vimgx, #2}{\halfvidw pt}{\vimgh pt}{#4}%
  \fill[inpaintyellow] (\vimgx + \halfvidw, #2) rectangle ++(\halfvidw, -\vimgh);%
  \draw[draw=black, line width=0.45pt] (\vimgx + \halfvidw, #2) -- ++(0, -\vimgh);%
  \pgfmathsetmacro{\vholew}{1.5}%
  \pgfmathsetmacro{\vholeh}{2.8}%
  \pgfmathsetmacro{\vstep}{\vimgh / 5.0}%
  \foreach \k in {1,...,5} {%
    \pgfmathsetmacro{\vholey}{#2 - (\k - 0.5)*\vstep + 0.5*\vholeh}%
    \pgfmathsetmacro{\vlholex}{#1 + 0.5*(\vrailw - \vholew)}%
    \pgfmathsetmacro{\vrholex}{#1 + #3 - 0.5*(\vrailw + \vholew)}%
    \fill[white, rounded corners=0.4pt] (\vlholex, \vholey) rectangle ++(\vholew, -\vholeh);%
    \fill[white, rounded corners=0.4pt] (\vrholex, \vholey) rectangle ++(\vholew, -\vholeh);%
  }%
  \draw[frameborder] (#1, #2) rectangle ++(#3, -\vimgh);%
  \node[reflabel] at (#1 + 0.5*#3, #2 - \vimgh - 1.6) {\reflabeltext{44.43}{#5}};%
}

\newcommand{\drawtemporalextensionstrips}[8]{%
  \pgfmathsetmacro{\labelpad}{8.5}%
  \pgfmathsetmacro{\stripgap}{6.5}%
  \pgfmathsetmacro{\stripw}{0.5 * (#3 - \labelpad - \stripgap)}%
  \pgfmathsetmacro{\vrailw}{2.8}%
  \pgfmathsetmacro{\vimgw}{\stripw - 2.0*\vrailw}%
  \pgfmathsetmacro{\vimgh}{\vimgw * 9.0 / 16.0}%
  \pgfmathsetmacro{\fstepgap}{3.024}%
  \pgfmathsetmacro{\twostriph}{2.0*\vimgh + \fstepgap}%
  \pgfmathsetmacro{\totstriph}{3.0*\vimgh + 2.0*\fstepgap}%
  \pgfmathsetmacro{\lx}{#1 + \labelpad}%
  \pgfmathsetmacro{\limgx}{\lx + \vrailw}%
  \pgfmathsetmacro{\rx}{\lx + \stripw + \stripgap}%
  \pgfmathsetmacro{\rimgx}{\rx + \vrailw}%
  \pgfmathsetmacro{\segytwo}{#2 - \vimgh - \fstepgap}%
  \pgfmathsetmacro{\segythree}{#2 - 2.0*(\vimgh + \fstepgap)}%
  \node[anchor=east, inner sep=0pt, font=\tabfigfont\fontsize{6.5pt}{7.8pt}\selectfont, text=plotspinecolor]
    at (\lx - 2.0, #2 - 0.5*\vimgh) {$\mathit{t}_1$};%
  \node[anchor=east, inner sep=0pt, font=\tabfigfont\fontsize{6.5pt}{7.8pt}\selectfont, text=plotspinecolor]
    at (\lx - 2.0, \segytwo - 0.5*\vimgh) {$\mathit{t}_2$};%
  \node[anchor=east, inner sep=0pt, font=\tabfigfont\fontsize{6.5pt}{7.8pt}\selectfont, text=plotspinecolor]
    at (\lx - 2.0, \segythree - 0.5*\vimgh) {$\mathit{t}_3$};%
  \fill[plotspinecolor] (\lx, #2) rectangle ++(\stripw, -\totstriph);%
  \frameboxsixteennine{\limgx, #2}{\vimgw pt}{\vimgh pt}{#4}%
  \frameboxsixteennine{\limgx, \segytwo}{\vimgw pt}{\vimgh pt}{#5}%
  \frameboxsixteennine{\limgx, \segythree}{\vimgw pt}{\vimgh pt}{#6}%
  \draw[draw=white, line width=0.65pt, dash pattern=on 1.4pt off 1.1pt]
    (\lx, #2 - \vimgh - 0.5*\fstepgap) -- ++(\stripw, 0);%
  \draw[draw=white, line width=0.65pt, dash pattern=on 1.4pt off 1.1pt]
    (\lx, \segytwo - \vimgh - 0.5*\fstepgap) -- ++(\stripw, 0);%
  \fill[plotspinecolor] (\rx, #2) rectangle ++(\stripw, -\twostriph);%
  \frameboxsixteennine{\rimgx, #2}{\vimgw pt}{\vimgh pt}{#7}%
  \frameboxsixteennine{\rimgx, \segytwo}{\vimgw pt}{\vimgh pt}{#8}%
  \draw[draw=white, line width=0.65pt, dash pattern=on 1.4pt off 1.1pt]
    (\rx, #2 - \vimgh - 0.5*\fstepgap) -- ++(\stripw, 0);%
  \pgfmathsetmacro{\vholew}{1.3}%
  \pgfmathsetmacro{\vholeh}{2.2}%
  \pgfmathsetmacro{\vstep}{\vimgh / 4.0}%
  \pgfmathsetmacro{\llholex}{\lx + 0.5*(\vrailw - \vholew)}%
  \pgfmathsetmacro{\lrholex}{\lx + \stripw - 0.5*(\vrailw + \vholew)}%
  \pgfmathsetmacro{\rlholex}{\rx + 0.5*(\vrailw - \vholew)}%
  \pgfmathsetmacro{\rrholex}{\rx + \stripw - 0.5*(\vrailw + \vholew)}%
  \foreach \s/\sy in {1/#2, 2/\segytwo, 3/\segythree} {%
    \foreach \k in {1,...,4} {%
      \pgfmathsetmacro{\vholey}{\sy - (\k - 0.5)*\vstep + 0.5*\vholeh}%
      \fill[white, rounded corners=0.35pt] (\llholex, \vholey) rectangle ++(\vholew, -\vholeh);%
      \fill[white, rounded corners=0.35pt] (\lrholex, \vholey) rectangle ++(\vholew, -\vholeh);%
    }%
  }%
  \foreach \s/\sy in {1/#2, 2/\segytwo} {%
    \foreach \k in {1,...,4} {%
      \pgfmathsetmacro{\vholey}{\sy - (\k - 0.5)*\vstep + 0.5*\vholeh}%
      \fill[white, rounded corners=0.35pt] (\rlholex, \vholey) rectangle ++(\vholew, -\vholeh);%
      \fill[white, rounded corners=0.35pt] (\rrholex, \vholey) rectangle ++(\vholew, -\vholeh);%
    }%
  }%
  \draw[frameborder] (\lx, #2) rectangle ++(\stripw, -\totstriph);%
  \draw[frameborder] (\rx, #2) rectangle ++(\stripw, -\twostriph);%
  \pgfmathsetmacro{\thirdtopy}{#2 - \twostriph}%
  \draw[draw=plotspinecolor, line width=0.55pt, dash pattern=on 1.0pt off 1.0pt]
    (\rx, \thirdtopy) -- ++(0, -\vimgh - \fstepgap) -- ++(\stripw, 0) -- ++(0, \vimgh + \fstepgap);%
  \draw[draw=plotspinecolor, line width=0.45pt, dash pattern=on 1.0pt off 1.0pt]
    (\rimgx, \segythree) rectangle ++(\vimgw, -\vimgh);%
  \foreach \k in {1,...,4} {%
    \pgfmathsetmacro{\vholey}{\segythree - (\k - 0.5)*\vstep + 0.5*\vholeh}%
    \draw[draw=plotspinecolor!80, line width=0.35pt, dash pattern=on 0.65pt off 0.65pt, rounded corners=0.35pt]
      (\rlholex, \vholey) rectangle ++(\vholew, -\vholeh);%
    \draw[draw=plotspinecolor!80, line width=0.35pt, dash pattern=on 0.65pt off 0.65pt, rounded corners=0.35pt]
      (\rrholex, \vholey) rectangle ++(\vholew, -\vholeh);%
  }%
  \node[reflabel] at (\lx + 0.5*\stripw, #2 - \totstriph - 1.6) {\reflabeltext{44.43}{@video1}};%
  \node[reflabel] at (\rx + 0.5*\stripw, #2 - \totstriph - 1.6) {\reflabeltext{44.43}{@video2}};%
}

\newsavebox{\tabcpromptboxthree}
\newsavebox{\tabcpromptboxfour}
\newsavebox{\tabcpromptboxfive}
\newcommand{\measuretabcprompt}[2]{%
  \sbox{#1}{\parbox[t]{120.8pt}{%
    \raggedright
    \tabfigfont\itshape\fontsize{6.5pt}{7.6pt}\selectfont\color{plotspinecolor}%
    \strut #2\strut\par\vspace{0pt}%
  }}%
}

\newcommand{\tabdesignspacepanelc}[3]{%
  \measuretabcprompt{\tabcpromptboxthree}{#1}%
  \measuretabcprompt{\tabcpromptboxfour}{#2}%
  \measuretabcprompt{\tabcpromptboxfive}{#3}%
  \begin{tikzpicture}[x=1pt, y=1pt]
    \pgfmathsetmacro{\panelw}{120.8}
    \pgfmathsetmacro{\panelh}{\tabpanelcheight}
    \pgfmathsetmacro{\pthreeh}{(\ht\tabcpromptboxthree + \dp\tabcpromptboxthree)/1pt}
    \pgfmathsetmacro{\pfourh}{(\ht\tabcpromptboxfour + \dp\tabcpromptboxfour)/1pt}
    \pgfmathsetmacro{\pfiveh}{(\ht\tabcpromptboxfive + \dp\tabcpromptboxfive)/1pt}
    \useasboundingbox (0, 0) rectangle (\panelw, -\panelh);

    \def\exonein{figures/separate_editing/example1_input.jpg}
    \def\exoneout{figures/separate_editing/example1_solution.jpg}
    \def\extwoin{figures/separate_editing/example2_input.jpg}
    \def\extwoout{figures/separate_editing/example2_solution.jpg}
    \def\qframe{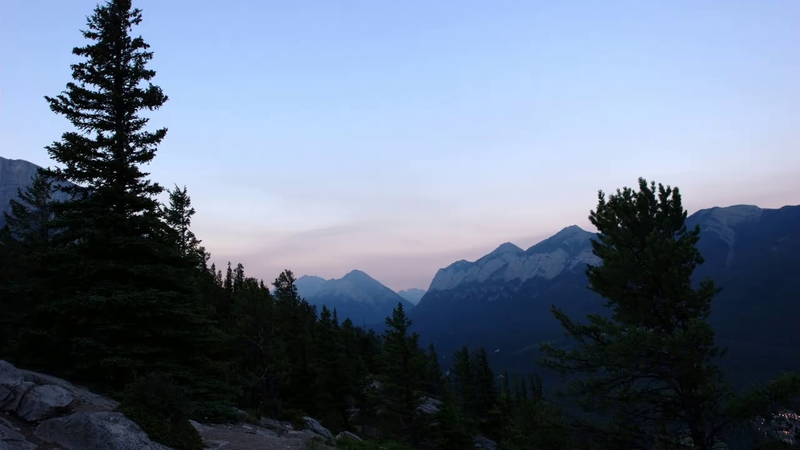}

    \pgfmathsetmacro{\bthreetopy}{0.0}
    \pgfmathsetmacro{\bthreecy}{\bthreetopy - 3.6}
    \pgfmathsetmacro{\bthreevisy}{\bthreetopy - 11.4 - \pthreeh}
    \node[fill=tabhlcolor, text=white, font=\figfont\bfseries\fontsize{6.0pt}{6.0pt}\selectfont,
          rounded corners=1.1pt, inner xsep=1.8pt, inner ysep=1.2pt, minimum height=7.2pt, anchor=west]
      (b3) at (0, \bthreecy) {3};
    \node[anchor=west, inner sep=0pt, outer sep=0pt, font=\tabfigfont\fontsize{6.5pt}{7.8pt}\selectfont, text=plotspinecolor]
      at ([xshift=3.8pt]b3.east) {separate--spatial--inpainting};
    \node[anchor=north west, inner sep=0pt, outer sep=0pt]
      at (0, \bthreetopy - 8.8) {\usebox{\tabcpromptboxthree}};
    \pgfmathsetmacro{\pairgap}{5.2}
    \pgfmathsetmacro{\pairw}{0.5 * (\panelw - \pairgap)}
    \drawrefimagepair{0}{\bthreevisy}{\pairw}{\exonein}{\exoneout}{@image1}
    \drawrefimagepair{\pairw + \pairgap}{\bthreevisy}{\pairw}{\extwoin}{\extwoout}{@image2}
    \drawspatialinpaintvideo{0}{\bthreevisy - 26.6}{\panelw}{\qframe}{@video1}
    \pgfmathsetmacro{\bthreeboty}{\bthreevisy - 64.0}

    \pgfmathsetmacro{\bfivetopy}{-\panelh + 104.2 + \pfiveh}
    \pgfmathsetmacro{\bfivecy}{\bfivetopy - 3.6}
    \pgfmathsetmacro{\bfivevisy}{\bfivetopy - 11.4 - \pfiveh}

    \pgfmathsetmacro{\bfourtot}{81.2 + \pfourh}
    \pgfmathsetmacro{\wsgap}{0.5 * ((\bthreeboty - \bfivetopy) - \bfourtot)}
    \pgfmathsetmacro{\bfourtopy}{\bthreeboty - \wsgap}
    \pgfmathsetmacro{\bfourcy}{\bfourtopy - 3.6}
    \pgfmathsetmacro{\bfourvisy}{\bfourtopy - 11.4 - \pfourh}
    \node[fill=tabhlcolor, text=white, font=\figfont\bfseries\fontsize{6.0pt}{6.0pt}\selectfont,
          rounded corners=1.1pt, inner xsep=1.8pt, inner ysep=1.2pt, minimum height=7.2pt, anchor=west]
      (b4) at (0, \bfourcy) {4};
    \node[anchor=west, inner sep=0pt, outer sep=0pt, font=\tabfigfont\fontsize{6.5pt}{7.8pt}\selectfont, text=plotspinecolor]
      at ([xshift=3.8pt]b4.east) {spatial--output-only--editing};
    \node[anchor=north west, inner sep=0pt, outer sep=0pt]
      at (0, \bfourtopy - 8.8) {\usebox{\tabcpromptboxfour}};
    \drawgridvideofilmstrip{0}{\bfourvisy}{\panelw}{\exoneout}{\extwoout}{\qframe}{@video1}

    \node[fill=tabhlcolor, text=white, font=\figfont\bfseries\fontsize{6.0pt}{6.0pt}\selectfont,
          rounded corners=1.1pt, inner xsep=1.8pt, inner ysep=1.2pt, minimum height=7.2pt, anchor=west]
      (b5) at (0, \bfivecy) {5};
    \node[anchor=west, inner sep=0pt, outer sep=0pt, font=\tabfigfont\fontsize{6.5pt}{7.8pt}\selectfont, text=plotspinecolor]
      at ([xshift=3.8pt]b5.east) {temporal--separate--extension};
    \node[anchor=north west, inner sep=0pt, outer sep=0pt]
      at (0, \bfivetopy - 8.8) {\usebox{\tabcpromptboxfive}};
    \drawtemporalextensionstrips{0}{\bfivevisy}{\panelw}{\exonein}{\extwoin}{\qframe}{\exoneout}{\extwoout}
  \end{tikzpicture}%
}

\newcommand{\drawqualitativevisiongrid}[9]{%
  \def\qvisargA{#1}%
  \def\qvisargB{#2}%
  \def\qvisargC{#3}%
  \def\qvisargD{#4}%
  \def\qvisargE{#5}%
  \def\qvisargF{#6}%
  \def\qvisargG{#7}%
  \def\qvisargH{#8}%
  \def\qvisargI{#9}%
  \drawqualitativevisiongridext
}
\newcommand{\drawqualitativevisiongridext}[3]{%
  \begin{tikzpicture}[x=1pt, y=1pt]
    \pgfmathsetmacro{\figw}{0.84 * \linewidth}
    \pgfmathsetmacro{\labelw}{38.0}
    \pgfmathsetmacro{\colgap}{3.6}
    \pgfmathsetmacro{\rowgap}{3.2}
    \pgfmathsetmacro{\cellw}{(\figw - \labelw - 3.0*\colgap) / 4.0}
    \pgfmathsetmacro{\cellh}{\cellw * 9.0 / 16.0}

    \pgfmathsetmacro{\colAx}{\labelw}
    \pgfmathsetmacro{\colBx}{\labelw + (\cellw + \colgap)}
    \pgfmathsetmacro{\colCx}{\labelw + 2.0*(\cellw + \colgap)}
    \pgfmathsetmacro{\colDx}{\labelw + 3.0*(\cellw + \colgap)}

    \pgfmathsetmacro{\rowAy}{0.0}
    \pgfmathsetmacro{\rowBy}{-(\cellh + \rowgap)}
    \pgfmathsetmacro{\rowCy}{-2.0*(\cellh + \rowgap)}

    \node[anchor=base west, inner sep=0pt, outer sep=0pt, text=plotspinecolor]
      at (\colAx, 4.2) {\figfont\fontsize{6.2pt}{7.4pt}\selectfont\strut\textbf{A}\hspace{0.25em}Surface normal estimation};

    \node[anchor=base west, inner sep=0pt, outer sep=0pt, text=plotspinecolor]
      at (\colBx, 4.2) {\figfont\fontsize{6.2pt}{7.4pt}\selectfont\strut\textbf{B}\hspace{0.25em}Monocular depth estimation};

    \node[anchor=base west, inner sep=0pt, outer sep=0pt, text=plotspinecolor]
      at (\colCx, 4.2) {\figfont\fontsize{6.2pt}{7.4pt}\selectfont\strut\textbf{C}\hspace{0.25em}Human pose estimation};

    \node[anchor=base west, inner sep=0pt, outer sep=0pt, text=plotspinecolor]
      at (\colDx, 4.2) {\figfont\fontsize{6.2pt}{7.4pt}\selectfont\strut\textbf{D}\hspace{0.25em}Video object segmentation};

    \node[anchor=west, inner sep=0pt, outer sep=0pt, align=left, text=plotspinecolor,
          font=\figfont\fontsize{6.8pt}{8.0pt}\selectfont]
      at (0, \rowAy - 0.5*\cellh) {\textbf{Query}};

    \node[anchor=west, inner sep=0pt, outer sep=0pt, align=left, text=plotspinecolor,
          font=\figfont\fontsize{6.8pt}{8.0pt}\selectfont]
      at (0, \rowBy - 0.5*\cellh) {\textbf{Ground}\\\textbf{truth}};

    \node[anchor=west, inner sep=0pt, outer sep=0pt, align=left, text=plotspinecolor,
          font=\figfont\fontsize{6.8pt}{8.0pt}\selectfont]
      at (0, \rowCy - 0.5*\cellh) {\textbf{Few-shot}\\{\fontsize{6.0pt}{7.2pt}\selectfont$k = 5$\textsuperscript{\textasteriskcentered}}};

    \frameboxsixteennine{\colAx, \rowAy}{\cellw pt}{\cellh pt}{\qvisargA}
    \frameboxsixteennine{\colBx, \rowAy}{\cellw pt}{\cellh pt}{\qvisargD}
    \frameboxsixteennine{\colCx, \rowAy}{\cellw pt}{\cellh pt}{\qvisargG}
    \frameboxsixteennine{\colDx, \rowAy}{\cellw pt}{\cellh pt}{#1}

    \frameboxsixteennine{\colAx, \rowBy}{\cellw pt}{\cellh pt}{\qvisargB}
    \frameboxsixteennine{\colBx, \rowBy}{\cellw pt}{\cellh pt}{\qvisargE}
    \frameboxsixteennine{\colCx, \rowBy}{\cellw pt}{\cellh pt}{\qvisargH}
    \frameboxsixteennine{\colDx, \rowBy}{\cellw pt}{\cellh pt}{#2}

    \frameboxsixteennine{\colAx, \rowCy}{\cellw pt}{\cellh pt}{\qvisargC}
    \frameboxsixteennine{\colBx, \rowCy}{\cellw pt}{\cellh pt}{\qvisargF}
    \frameboxsixteennine{\colCx, \rowCy}{\cellw pt}{\cellh pt}{\qvisargI}
    \frameboxsixteennine{\colDx, \rowCy}{\cellw pt}{\cellh pt}{#3}
  \end{tikzpicture}%
}

\makeatletter
\providecommand{\@LN}[2]{}

\newcommand{\appendixtableofcontents}{%
  {\@ifpackageloaded{cvpr}{\def\numberline##1{##1\hskip 1em\relax}}{}%
   \@starttoc{atoc}}%
  \let\orig@addtocontents\addtocontents
  \renewcommand{\addtocontents}[2]{%
    \orig@addtocontents{##1}{##2}%
    \ifstrequal{##1}{toc}{\orig@addtocontents{atoc}{##2}}{}%
  }%
}

\AtEndDocument{\immediate\write\@auxout{\shotsaveminmax}}

\def\bshotcell{\bfseries}

\newcommand{\shot@top}[5]{%
  \iftok\relax\detokenize{#2#3#4}\relax
    \expandafter\@firstoftwo
  \else
    \expandafter\@secondoftwo
  \fi
  {\multicolumn{3}{c}{\raisebox{-0.5\baselineskip}[0pt][0pt]{#5}}}%
  {\ifx#1\bshotcell
     \expandafter\@firstoftwo
   \else
     \expandafter\@secondoftwo
   \fi
   {\shotoldnumcell{\bfseries}{#2} & \shotoldnumcell{\bfseries}{#3} & \shotoldnumcell{\bfseries}{#4}}%
   {\shotoldnumcell{}{#2} & \shotoldnumcell{}{#3} & \shotoldnumcell{}{#4}}}%
}
\newcommand{\shot@bot}[4]{%
  \iftok\relax\detokenize{#1#2#3}\relax
    \expandafter\@firstoftwo
  \else
    \expandafter\@secondoftwo
  \fi
  {\multicolumn{3}{c}{}}%
  {\multicolumn{3}{c}{\footnotesize #4}}%
}
\def\rowlabel#1#2#3#4#5#6#7#8#9{%
  \rowlabel@ii
    {\shottrack{#4}\shottrack{#5}\shottrack{#6}\textbf{#1} & \shot@top{#3}{#4}{#5}{#6}{#7}}%
    {& \shot@bot{#4}{#5}{#6}{#7}}%
    {#7}%
    {#9}%
}
\def\rowlabel@ii#1#2#3#4#5#6#7#8#9{%
  \rowlabel@iii
    {#1\shottrack{#5}\shottrack{#6}\shottrack{#7} & \shot@top{#4}{#5}{#6}{#7}{#8}}%
    {#2 & \shot@bot{#5}{#6}{#7}{#8}}%
    {#3#8}%
}
\def\rowlabel@iii#1#2#3#4#5#6#7#8{%
  \iftok\relax\detokenize{#3#8}\relax
    \expandafter\@firstoftwo
  \else
    \expandafter\@secondoftwo
  \fi
  {\shottrack{#5}\shottrack{#6}\shottrack{#7}#1 & \shot@top{#4}{#5}{#6}{#7}{#8}}%
  {\shottrack{#5}\shottrack{#6}\shottrack{#7}#1 & \shot@top{#4}{#5}{#6}{#7}{#8} \\
   #2 & \shot@bot{#5}{#6}{#7}{#8}}%
}
\makeatother
\newcommand{\itpipe}{\rotatebox[origin=c]{-11}{|}}

\title{Video~models~are~in-context~learners} 
\usepackage[numbers,square,sort&compress]{natbib}
\correspondingauthor{my-email@google.com}

\reportnumber{} %

\author[1]{Andreas~Hochlehnert\textsuperscript{*}}
\author[1]{Thaddäus~Wiedemer\textsuperscript{*}}
\author[1]{Paul~Hayes}
\author[1]{Thomas~Kipf}
\author[1]{Been~Kim}
\author[1]{Robert~Geirhos\textsuperscript{*}}

\affil[]{Google DeepMind}

\begin{abstract}
The transition from zero-shot prompting to few-shot in-context learning was an important milestone in the rise of large language models as general-purpose foundation models. We show for the first time that off-the-shelf generative video models (Gemini Omni 1.1, Seedance 2.5) are in-context learners too, requiring no weight updates, adapters, or architectural changes. Our findings are threefold: (1)~Video models can leverage in-context examples to beat zero-shot performance across a wide variety of spatiotemporal prompt formats. (2)~Across many tested computer vision and reasoning tasks, error rates systematically decrease with more in-context examples. (3)~On a controlled task, in-context learning and best-of-N sampling are super-additive: more examples in context make inference-time scaling increasingly effective. Our findings reveal a deep parallel between frontier language and video models: the ability to infer patterns from in-context examples. This convergence raises a broader question: does intelligence in generative models follow a common developmental path, irrespective of the modality?\website{\\Project page: \url{https://video-in-context-learning.github.io/}}

\end{abstract}

\begin{document}
\maketitle

\section{Introduction}\label{sec:intro}
Natural language processing progressed from task-specific models to general-purpose foundation models in three steps: after generative models initially relied on task-specific fine-tuning~\citep[\gptone{};][]{radford2018improving}, larger-scale models began to show promising zero-shot behavior~\citep[\gpttwo{};][]{radford2019language}, and finally unlocked few-shot \emph{in-context learning}~(ICL)~\citep[\gptthree{};][]{brown2020language}: the ability to specify tasks at inference time via input--output examples without any changes to model weights~\citep{bommasani2021opportunities,dong2022survey,lampinen2024broader,anil2024manyshot}.

Building on advances in diffusion models and generative modeling~\citep{sohl2015deep,ho2020denoising,song2020score}, machine vision is beginning to trace a remarkably similar arc: web-scale generative image and video models learn rich representations of geometry and motion~\citep{huang2025how,nam2025emergent,son2025probing,esmati2026invisible,punzo2026how}. Similar to \gptone{} in NLP, they can be fine-tuned to obtain state-of-the-art vision models~\citep{gabeur2026image,wang2026thfm,wang2026video,chen2025univid,he2026icdepth,zhang2026dvd,jiang2025geo4d,wang2026rgb}. Recently, mirroring \gpttwo{}, generative visual models are emerging as zero-shot learners and reasoners across a wide variety of perception, physical modeling, and visual reasoning tasks~\citep{wiedemer2026video,di2026are,liu2026opensource,tong2025thinking,wang2026very,xu2026vbvr,deng2025video,guo2025video,shrivastava2026point,zeller2026mentisoculi}.

Despite the success of ICL across language and vision-language models~\citep{alayrac2022flamingo,gadetsky2025large,rostami2026incontext}, image and video generation has not had its ``\gptthree{} moment'' yet: here, existing ICL approaches require learned adapters, specific pretraining, architectural modifications, or inference-time tweaks~\citep{liu2024aicl,bian2025videoasprompt,ye2025unic,kao2026unifying,ju2025editverse,zhang2025regionconstraint,acuaviva2025generation,acuaviva2025rethinking,li2025iceffect,li2025vfxmaster,wang2026vicedit,liao2025incontext,nguyen2026universal,li2026viral,gu2024analogist,lin2025realgeneral,oorloff2025stable}, and often struggle to generalize~\citep{halady2026quo}. Without such modifications, frontier models can be steered by their visual prompts~\citep{geirhos2026visual,fang2025invideo}, but not learn from examples.

Yet, unlocking ICL in off-the-shelf models is crucial for transitioning from task-specific models to a single general-purpose model. Zero-shot prompting alone is insufficient, since many visual concepts are hard or impossible to verbalize without examples: from output conventions in dense tasks like surface normal or pose estimation (e.g., colormaps, skeleton topologies, and coordinate systems)~\citep{di2026are} to complex visual, physical, or artistic transformations~\citep{chen2026viper,wang2026vicedit}. Providing such visual examples requires a large multimodal context window, which frontier video models have recently begun to support via image and video references alongside interaction modes like editing and extension~\citep{deepmind2026geminiomni,google2026geminiomni,google2026geminiomni11,seedance2026seedance,seedance2026seedance25,zhang2026vidu}.

\begin{figure*}[t]
    \centering
    \begin{tikzpicture}[x=1pt, y=1pt, trim left=0pt, trim right=\linewidth]
        \pgfmathsetmacro{\figw}{\linewidth}
        \pgfmathsetmacro{\colgap}{16.0}
        \pgfmathsetmacro{\halfw}{(\figw - \colgap) / 2.0}
        \pgfmathsetmacro{\refcolw}{0.2745 * \halfw}
        \pgfmathsetmacro{\refgap}{0.0216 * \halfw}
        \pgfmathsetmacro{\vidgap}{\refgap}
        \pgfmathsetmacro{\vidslotw}{(\halfw - \refcolw - 2.0*\refgap) / 2.0}
        \pgfmathsetmacro{\vidboxh}{((\vidslotw - 7.2) * 9.0 / 16.0)}
        \pgfmathsetmacro{\promptspace}{11.0}
        \pgfmathsetmacro{\rowstep}{60.0}

        \def\quantinframe{figures/fig1/297312812_default_018_k9_attempt_003_orig_query_frame095.png}
        \def\quantoutframe{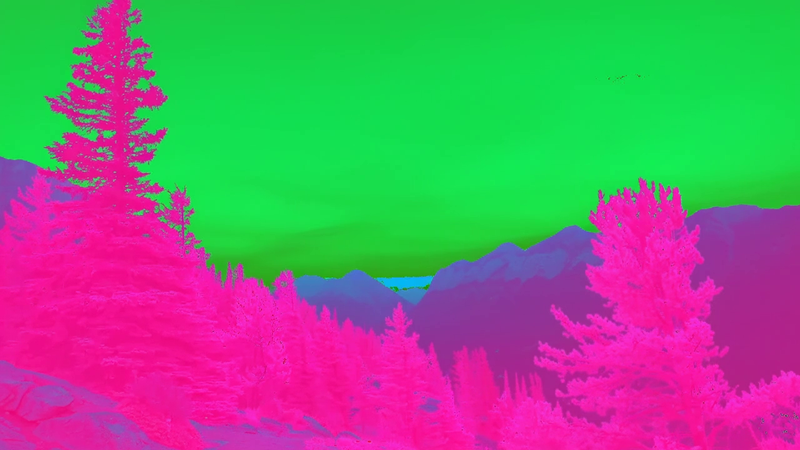}

        \renewcommand{\thesubfigure}{\Alph{subfigure}}%
        \node[paneltitle] at (0, \promptspace + 6.5) {{\phantomsubcaption\label{fig:motivating-examples-a}}\textbf{A}\hspace{0.55em}Qualitative examples of \textbf{one-shot learning}};
        \node[paneltitle] at (\halfw + \colgap, \promptspace + 6.5) {{\phantomsubcaption\label{fig:motivating-examples-b}}\textbf{B}\hspace{0.55em}Quantitative analysis of \textbf{few-shot learning}};

        \drawiclgroup{0}{0.0}{``Perform surface normal estimation by matching the style of \reftt{@image1}.''}{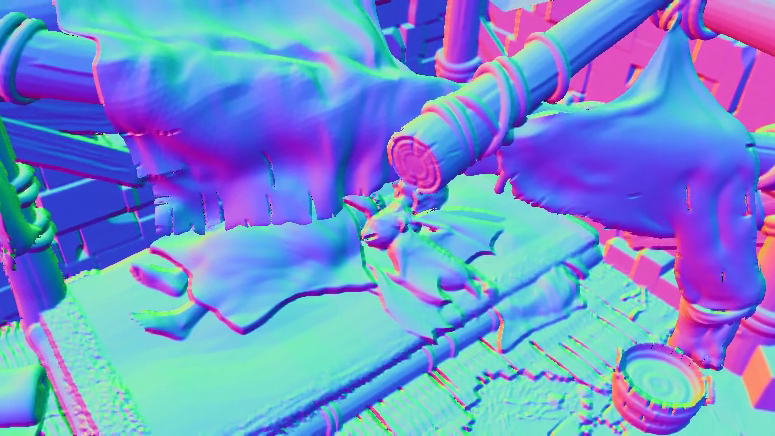}{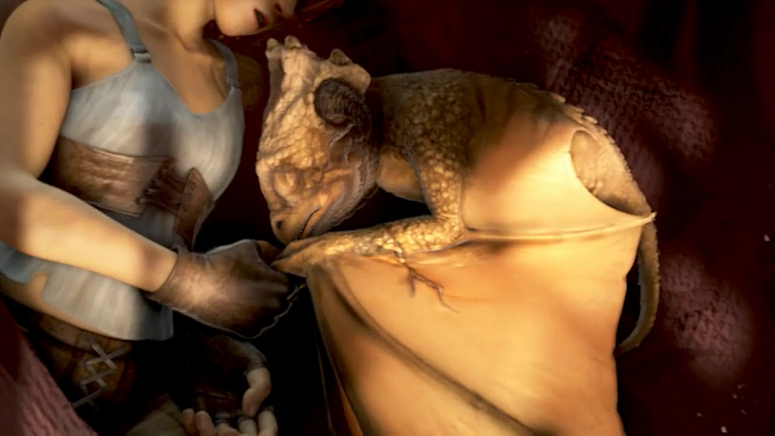}{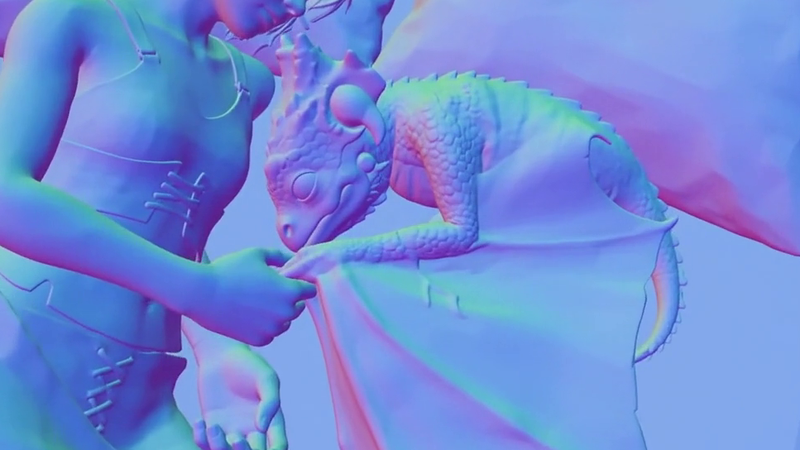}
        \drawiclgroup{0}{-\rowstep}{``Perform depth estimation by matching the style of \reftt{@image1}.''}{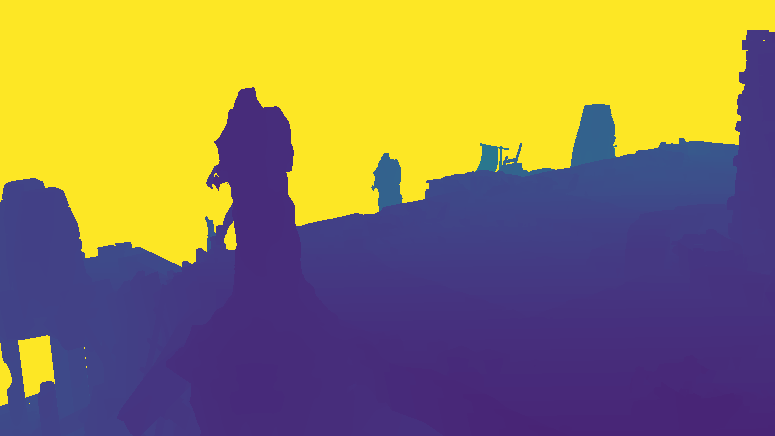}{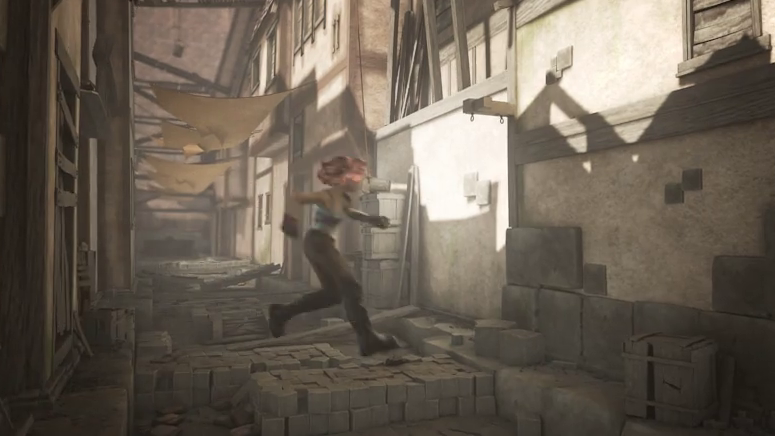}{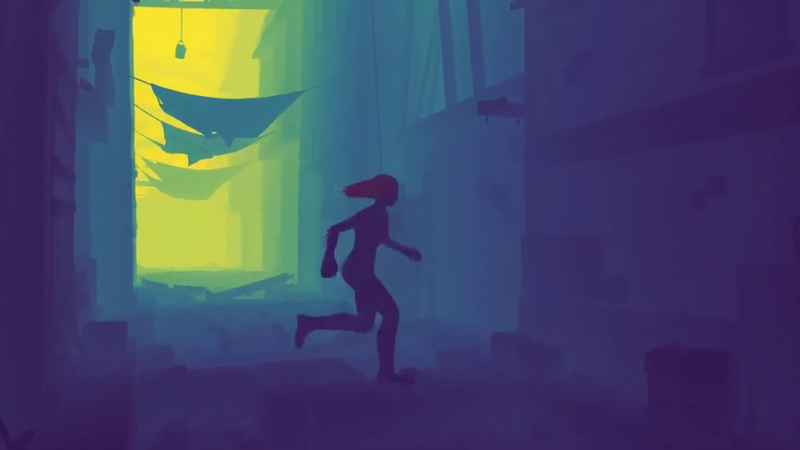}
        \drawiclgroupvideoref{0}{-2.0*\rowstep}{``The person walks to the cyan circle as in \reftt{@video1}, trailing a red line.''}{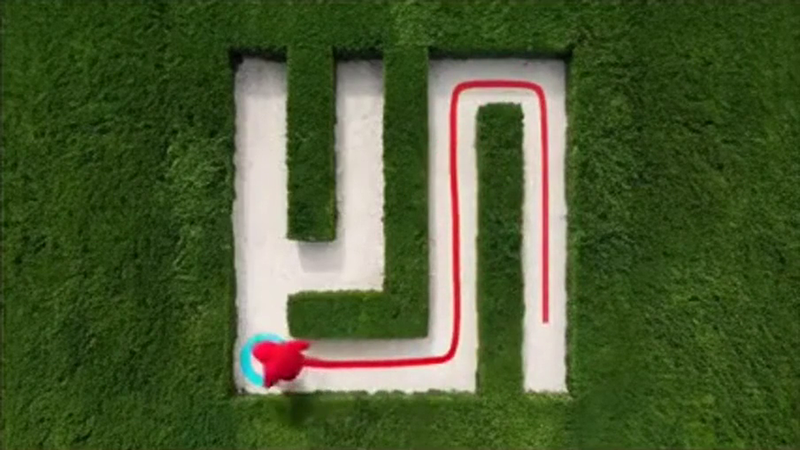}{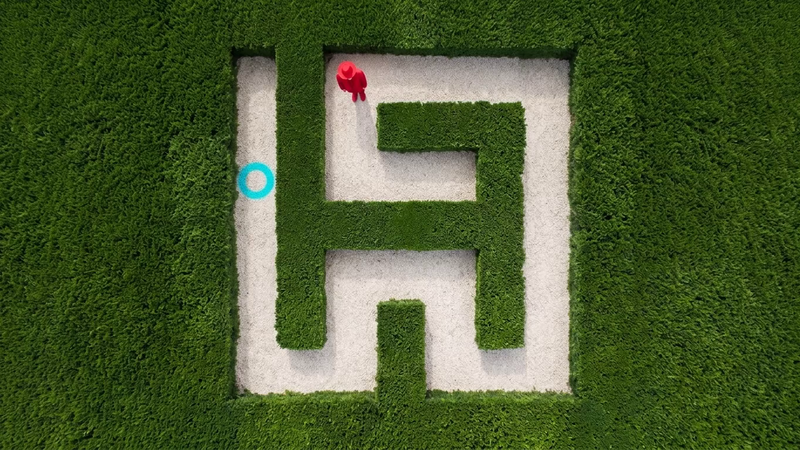}{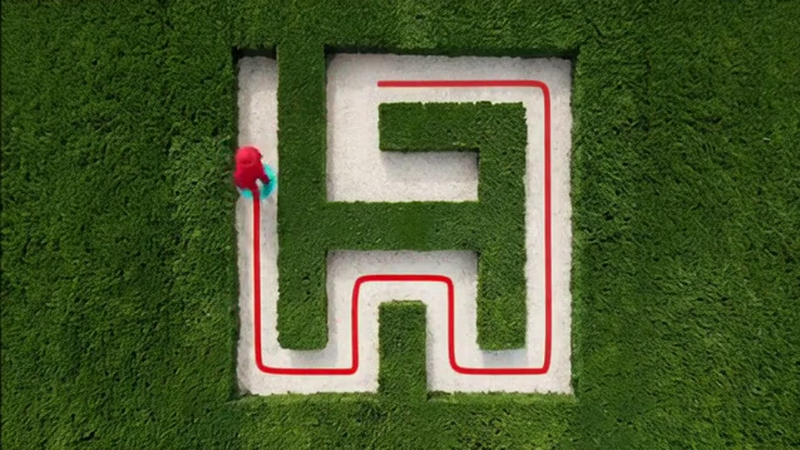}
        \drawiclgroupvideoref{0}{-3.0*\rowstep}{``The ball rolls down in a physically plausible way as in \reftt{@video1}.''}{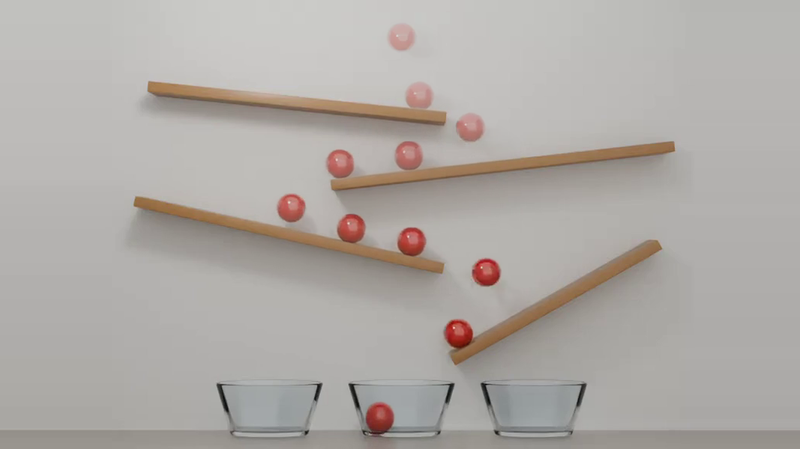}{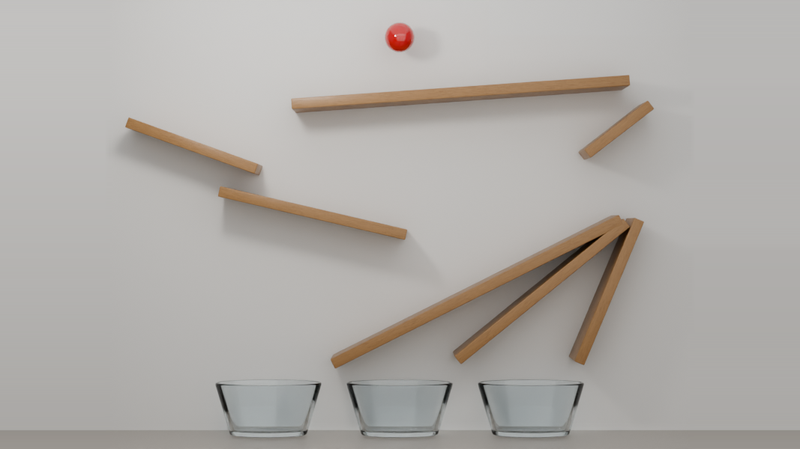}{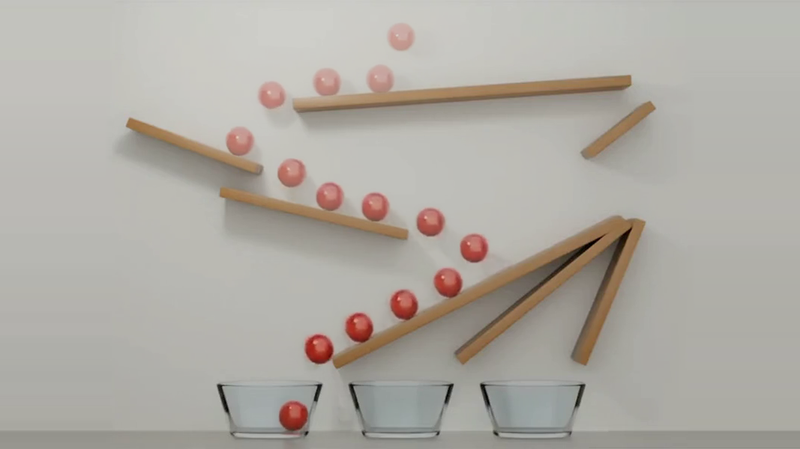}
        \pgfmathsetmacro{\panelaboty}{-3.0*\rowstep - \vidboxh - 4.0}

        \pgfmathsetmacro{\rightx}{\halfw + \colgap}
        \drawquantblock{\rightx}{\promptspace}{``Here are examples with their solutions: input \reftt{@image1} $\rightarrow$ solution \reftt{@image2}, input \reftt{@image3} $\rightarrow$ solution \reftt{@image4}. [\dots] Edit the video as follows: (1.) Red channel: invert the original blue channel (cross-channel inversion); [\dots] Keep everything else the same.''}

        \pgfmathsetmacro{\plottopy}{\quantblockboty - 4.0}
        \pgfmathsetmacro{\plotcentery}{0.5 * (\plottopy + \panelaboty)}

        \node[inner sep=0pt, outer sep=0pt] at (\rightx + 0.5*\halfw, \plotcentery) {%
            \includegraphics[width=\halfw pt]{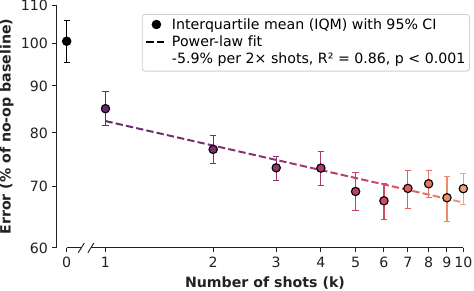}%
        };
    \end{tikzpicture}
    \caption{
        \textbf{Video models are in-context learners}.\website{ See \href{https://video-in-context-learning.github.io/}{website} for videos.}
        \textbf{A}:~With a single in-context example, pretrained unmodified video models like \omnione{} and \seedancetwofive{} can tackle a wide range of video2video and image2video tasks in pixel space.
        \textbf{B}:~We quantify ICL in video models, finding that a single input--output example can substantially reduce \omnione{}'s error, and additional samples can yield systematic further reductions. We illustrate our approach on \chromaticscrambleone{} (\cref{sec:approach}), a carefully constructed and difficult color transformation, and evaluate general computer vision and visual reasoning tasks in \cref{sec:results}. Prompts are paraphrased, details in \cref{app:task_details}.
    }\label{fig:motivating-examples}
\end{figure*}

\noindent
\emph{Despite being trained for aesthetic content creation, can these off-the-shelf video models be repurposed to infer patterns and learn from in-context examples---without weight changes, adapters, or architectural tweaks?}
\callout{
We find that \textbf{video models are in-context learners}:
\begin{enumerate}[leftmargin=*, topsep=3pt, itemsep=1pt]
    \item \textbf{ICL beats zero-shot performance across spatiotemporal prompt formats}: One-shot and few-shot examples systematically reduce error, across 30+ spatiotemporal prompt formats (\cref{sec:design-space}).
    \item \textbf{ICL is a general ability of frontier video models}: Both Gemini Omni 1.1 and Seedance 2.5 successfully infer patterns from in-context examples across many vision and reasoning tasks~(\cref{sec:results}).
    \item \textbf{ICL and best-of-N sampling are super-additive}: For video models, in-context examples make inference-time sampling more effective~(\cref{sec:context_bon_interaction}).
\end{enumerate}
}
\noindent
To be clear: they do not yet rival the ability of bespoke models trained or fine-tuned for these tasks. Nonetheless, the emergence of ICL in off-the-shelf video models raises a question for machine vision: Should we ignore the parallels to NLP, and continue building comparatively cheap, high-performing, yet ``narrow-intelligence'' task-specific models? Alternatively, if intelligence in generative models continues to follow a common developmental trajectory across language and vision, what are the implications of video ICL for building visual general intelligence~\citep{kataoka2026visual}?

\section{Approach}
\label{sec:approach}
\paragraph{Definition} Our goal is to measure the capability for in-context learning in off-the-shelf generative video models. ICL comes in many flavors; we follow the definition of \citet[][p.~2]{lampinen2024broader}: \emph{``Adaptation to properties within a context, yielding lower loss''}. Specifically, for video models with a multimodal context window, we provide an image or video query alongside a text prompt that describes the task, and measure whether placing one-shot or few-shot task examples in context reduces the model's error rate.

\paragraph{Models} We investigate ICL in two distinct frontier models: \geminiomnione{}~\citep{deepmind2026geminiomni} and \seedancetwofive{}~\citep{seedance2026seedance25}. Both were released around a similar time, score high on video leaderboards~\citep{lmarena_leaderboard,aavideo2026leaderboard}, and, unlike most open-weight models, support the large multimodal context window required for ICL. Inference details in~\cref{app:inference-details}.

\paragraph{Tasks} In \cref{sec:results}, we evaluate a range of computer vision and visual reasoning tasks. As for ICL analyses of frontier LLMs, we cannot and do not make assumptions about which tasks may have been part of the training data for video models.
However, we specifically create some evaluation tasks from scratch to rule out task familiarity as a confounder~\citep[cf.][]{halady2026quo,mayilvahanan2025search,mayilvahanan2024does}. We introduce a first task here as a running example: \chromaticscrambleone{} swaps, inverts, and cyclically shifts an input's $r, g, b$ color channels:
\begin{equation}
\label{eq:imagetransform}
    r_\text{new} = 255 - b, g_\text{new} = g, b_\text{new} = \mathrm{mod}((r + 128), 256).
\end{equation}
An example is shown in \cref{fig:motivating-examples-b}. We apply this transform to two datasets to obtain query--target pairs and input--output examples for an image-to-image~(I2I) and a video-to-video~(V2V) version (details and additional controlled tasks in \cref{app:image_and_video_transforms}). We use this task to explain our overall approach, and to investigate different prompt formats~(\cref{sec:design-space}).

\paragraph{Evaluation and statistics}
We prompt the model to edit the query video using a text description of \cref{eq:imagetransform} and $k$ input--output examples provided as separate files. For I2I~(\cref{fig:motivating-examples-b,tab:design_space_image}), the query is a static video of the query image, and we report the mean squared error (MSE) between the target and last frame, normalized by the unmodified query's MSE (no-op baseline corresponding to 100\% error). For V2V~(\cref{tab:design_space_video}), we instead average the normalized MSE over all frames. We aggregate across queries and attempts via the outlier-robust interquartile mean (IQM)~\citep{agarwal2021deep} and fit a power-law over $k \geq 1$ (dashed line in \cref{fig:motivating-examples-b}), reporting goodness of fit $R^2$ and the $p$-value against a flat-line null hypothesis (no improvement with more in-context examples). More details in \cref{app:eval-statistics}.

\paragraph{In-context learning results on a controlled task}  \cref{fig:motivating-examples-b} shows \omnione{} to be an in-context learner: while zero-shot performance ($k=0$) is as bad as not applying any transformation, the model successfully utilizes information from the provided input--output examples to reduce error. A power law fit explains 86\% of variance, and the results provide evidence of both one-shot learning ($k=1$ better than $k=0$) and few-shot learning (larger $k \rightarrow$ lower error).

\section{Which prompt formats enable video ICL?}
\label{sec:design-space}
In language models with a 1D token stream, constructing ICL prompts is straightforward: input--output examples are arranged sequentially. Frontier generative video models, on the other hand, allow arbitrary context arrangements in space and time and across multiple reference files. There is also a choice of interaction mode: the model can generate a new video from scratch, or edit or extend an existing video. Based on the success of visual prompt engineering in video models~\citep{geirhos2026visual}, the prompt format likely plays \emph{some} role in ICL. To measure this sensitivity and to identify robust settings for \cref{sec:results,sec:context_bon_interaction}, we evaluate \chromaticscrambleone{} under a large number of prompt formats.

\paragraph{Prompt structure} Every prompt has three parts:
\begin{itemize}
    \item A description of the query and example layout, referencing media files via tags (e.g., \texttt{@image1}; sometimes this is called reference-guided generation or R2V).
    \item A specification of the file, region, or time slice to modify.
    \item A text description of the transform (e.g., \textit{``Red channel: invert the original blue channel; green channel: identical; blue channel: 180-degree color phase shift on the original red channel. Keep everything else the same.''}). This description is identical across shot counts, so the 0-shot condition is a fully specified text prompt and any improvement at $k \geq 1$ is attributable to the examples.
\end{itemize}

\paragraph{Design space} We explore three dimensions in \cref{tab:design_space}:
\begin{itemize}
    \item \textbf{Interaction mode}. In \emph{editing}, the model edits the query video. In \emph{inpainting}, the model fills in a neon-yellow region (shown next to or after the query). In \emph{extension}, the model appends a new segment. In \emph{generation}, the model generates a new video from scratch.
    \item \textbf{Across-example packing}. Multiple examples are passed as \emph{separate} reference files, tiled \emph{spatially} into a single grid with the query in the bottom-right cell, or concatenated \emph{temporally} into a single video (query = final segment).
    \item \textbf{Within-example packing}. Each example is either \emph{output-only} or an input--output pair, provided as \emph{separate} files, tiled \emph{spatially} side-by-side, or concatenated \emph{temporally} one after the other.
\end{itemize}
In all cases, we score only the spatiotemporal region that the model was instructed to generate or modify. Some impractical combinations (e.g., the model cannot extend only part of the spatially-tiled grid canvas) are left blank in \cref{tab:design_space}. More details and full prompt templates in \cref{app:eval-statistics,app:prompts}.

\begin{table*}[t!]
  \centering
  \setlength{\tabcolsep}{\tabonecolsep}
  \renewcommand{\arraystretch}{0.76}
  \setlength{\extrarowheight}{3.5pt}
  \shotcolormap{0.0}{-22.0}   %
  \caption{\textbf{\omnione{} adapts to in-context examples across a large prompt design space}: The model can be asked to edit, inpaint, or extend an existing video, or to generate a new one from scratch(\emph{interaction mode}, columns 3--6). Within a single in-context example, the example input and output can be packed as separate files, a spatial grid, or a temporal sequence (\emph{within-example packing}, column~2). Multiple examples can be grouped in the same way (\emph{across-example packing}, column~1).\website{ See \href{https://video-in-context-learning.github.io/}{website} for interactive visualization.}
  Percentage point error change on \chromaticscrambleone{} relative to 100\% (no-op baseline); lower is better; details in \cref{sec:approach}. A few \taboutlined{outlined} prompt formats are illustrated on the right; dashes mark infeasible combinations.
  }
  \label{tab:design_space}
  \renewcommand{\thesubtable}{\Alph{subtable}}%
  \noindent%
  \begin{lrbox}{\tableftcolbox}%
  \begin{minipage}[t]{\taboneleftw}
    \begin{subtable}[t]{\linewidth}
      \centering
      \subtabheader{tab:design_space_image}{A}{I2I task}{. \omnione{} struggles 0-shot despite a textual description of the transform; 1-shot reduces the error in 31 of 32 configurations, often with further gains at 4 shots. Providing examples separately generally works best.}
      \vspace{3.0pt}
      \tabonebodyfont
    \begin{tabular*}{\linewidth}{
      @{} l
      @{\hspace{\taboneacrossep}} l
      !{\extracolsep{\fill}} R !{\extracolsep{0pt}} R R
      !{\extracolsep{\fill}} R !{\extracolsep{0pt}} R R
      !{\extracolsep{\fill}} R !{\extracolsep{0pt}} R R
      !{\extracolsep{\fill}} R !{\extracolsep{0pt}} R R @{\hspace{\tabcolsep}}
    }
      \toprule
      \multicolumn{2}{@{}l}{\textbf{Example packing}} & \multicolumn{3}{c}{\textbf{Editing}} & \multicolumn{3}{c}{\textbf{Inpainting}} & \multicolumn{3}{c}{\textbf{Extension}} & \multicolumn{3}{c@{\hspace{\tabcolsep}}}{\textbf{Generation}} \\
      \cmidrule(r{1pt}){1-2} \cmidrule(l{1pt}r{1pt}){3-5} \cmidrule(l{1pt}r{1pt}){6-8} \cmidrule(l{1pt}r{1pt}){9-11} \cmidrule(l{1pt}){12-14}
      Across & Within & {0-shot} & {1-shot} & {4-shot} & {0-shot} & {1-shot} & {4-shot} & {0-shot} & {1-shot} & {4-shot} & {0-shot} & {1-shot} & {4-shot} \\
      \tabcolmarks{a}
      \midrule
      \tabrowmark{ar1T}
      \smash{\makecell[tl]{Separate\\[-0.3ex]\tabonesubfont\emph{as individual}\\[-0.5ex]\tabonesubfont\emph{files}}} & \emph{output-only} & \shotnumcell{+3.9} & \shotnumcell{-10.6} & \shotnumcell{-15.6} & \shotnumcell{+1.1} & \shotnumcell{-10.9} & \shotnumcell{-22.0} & \multicolumn{3}{c}{---} & \shotnumcell{-2.3} & \shotnumcell{-11.0} & \shotnumcell{-17.4} \\
      \tabrowmarktwo{ar1B}{ar2T}
               & separate     & \shotnumcell{+3.9} & \shotnumcell{-13.4} & \shotnumcell{-18.7} & \shotnumcell{+1.1} & \shotnumcell{-10.7} & \shotnumcell{-14.3} & \multicolumn{3}{c}{---} & \shotnumcell{-2.3} & \shotnumcell{-12.6} & \shotnumcell{-19.1} \\
      \tabrowmarktwo{ar2B}{ar3T}
               & spatial            & \shotnumcell{+3.1} & \shotnumcell{-12.2} & \shotnumcell{-18.2} & \shotnumcell{+3.8} & \shotnumcell{-9.8} & \shotnumcell{-20.8} & \multicolumn{3}{c}{---} & \multicolumn{3}{c@{\hspace{\tabcolsep}}}{---} \\
      \tabrowmarktwo{ar3B}{ar4T}
               & temporal           & \shotnumcell{+2.9} & \shotnumcell{-8.5} & \shotnumcell{-16.9} & \shotnumcell{-0.8} & \shotnumcell{-6.2} & \shotnumcell{-15.1} & \shotnumcell{-1.2} & \shotnumcell{-11.1} & \shotnumcell{-21.1} & \multicolumn{3}{c@{\hspace{\tabcolsep}}}{---} \\
      \tabrowmark{ar4B}
      \addlinespace[0.8ex]
      \tabrowmark{ar5T}
      \smash{\makecell[tl]{Spatial\\[-0.3ex]\tabonesubfont\emph{as a grid}}} & \emph{output-only} & \shotnumcell{+3.9} & \shotnumcell{+3.7} & \shotnumcell{+4.8} & \shotnumcell{+1.1} & \shotnumcell{-5.6} & \shotnumcell{-4.2} & \multicolumn{3}{c}{---} & \shotnumcell{-2.3} & \shotnumcell{-7.2} & \shotnumcell{-9.6} \\
      \tabrowmarktwo{ar5B}{ar6T}
               & separate      & \shotnumcell{+3.9} & \shotnumcell{+0.2} & \shotnumcell{-4.4} & \shotnumcell{+1.1} & \shotnumcell{-2.6} & \shotnumcell{-4.5} & \multicolumn{3}{c}{---} & \multicolumn{3}{c@{\hspace{\tabcolsep}}}{---} \\
      \tabrowmarktwo{ar6B}{ar7T}
               & spatial            & \shotnumcell{+3.1} & \shotnumcell{-1.6} & \shotnumcell{-4.0} & \shotnumcell{+3.8} & \shotnumcell{-1.0} & \shotnumcell{+12.6} & \multicolumn{3}{c}{---} & \multicolumn{3}{c@{\hspace{\tabcolsep}}}{---} \\
      \tabrowmarktwo{ar7B}{ar8T}
               & temporal           & \shotnumcell{+2.9} & \shotnumcell{-2.5} & \shotnumcell{-15.4} & \shotnumcell{-0.8} & \shotnumcell{-5.4} & \shotnumcell{-6.9} & \multicolumn{3}{c}{---} & \multicolumn{3}{c@{\hspace{\tabcolsep}}}{---} \\
      \tabrowmark{ar8B}
      \addlinespace[0.8ex]
      \tabrowmark{ar9T}
      \smash{\makecell[tl]{Temporal\\[-0.3ex]\tabonesubfont\emph{as a sequence}}} & \emph{output-only} & \shotnumcell{+3.9} & \shotnumcell{+2.5} & \shotnumcell{-2.6} & \shotnumcell{+1.1} & \shotnumcell{-0.3} & \shotnumcell{-6.3} & \shotnumcell{-0.4} & \shotnumcell{-4.9} & \shotnumcell{-11.6} & \shotnumcell{-2.3} & \shotnumcell{-4.3} & \shotnumcell{-14.8} \\
      \tabrowmarktwo{ar9B}{ar10T}
               & separate     & \shotnumcell{+3.9} & \shotnumcell{-0.6} & \shotnumcell{-6.6} & \shotnumcell{+1.1} & \shotnumcell{-0.4} & \shotnumcell{-11.6} & \shotnumcell{-0.4} & \shotnumcell{-8.6} & \shotnumcell{-18.7} & \multicolumn{3}{c@{\hspace{\tabcolsep}}}{---} \\
      \tabrowmarktwo{ar10B}{ar11T}
               & spatial            & \shotnumcell{+3.1} & \shotnumcell{-1.4} & \shotnumcell{-4.3} & \shotnumcell{+3.8} & \shotnumcell{-8.3} & \shotnumcell{-7.7} & \multicolumn{3}{c}{---} & \multicolumn{3}{c@{\hspace{\tabcolsep}}}{---} \\
      \tabrowmarktwo{ar11B}{ar12T}
               & temporal           & \shotnumcell{+2.9} & \shotnumcell{+0.3} & \shotnumcell{-3.5} & \shotnumcell{-0.8} & \shotnumcell{+1.5} & \shotnumcell{-10.5} & \shotnumcell{-0.4} & \shotnumcell{-4.9} & \shotnumcell{-10.8} & \multicolumn{3}{c@{\hspace{\tabcolsep}}}{---} \\
      \tabrowmark{ar12B}
      \bottomrule
    \end{tabular*}
    \end{subtable}

    \vspace{8.0pt}

    \begin{subtable}[t]{\linewidth}
      \centering
      \subtabheader{tab:design_space_video}{B}{V2V task}{. This variant is harder than I2I, since the transform must be applied consistently over many frames. Many prompt formats show improvements, with editing being the most robust \emph{interaction mode}.}
      \vspace{3.0pt}
      \tabonebodyfont
    \begin{tabular*}{\linewidth}{
      @{} l
      @{\hspace{\taboneacrossep}} l
      !{\extracolsep{\fill}} R !{\extracolsep{0pt}} R R
      !{\extracolsep{\fill}} R !{\extracolsep{0pt}} R R
      !{\extracolsep{\fill}} R !{\extracolsep{0pt}} R R
      !{\extracolsep{\fill}} R !{\extracolsep{0pt}} R R @{\hspace{\tabcolsep}}
    }
      \toprule
      \multicolumn{2}{@{}l}{\textbf{Example packing}} & \multicolumn{3}{c}{\textbf{Editing}} & \multicolumn{3}{c}{\textbf{Inpainting}} & \multicolumn{3}{c}{\textbf{Extension}} & \multicolumn{3}{c@{\hspace{\tabcolsep}}}{\textbf{Generation}} \\
      \cmidrule(r{1pt}){1-2} \cmidrule(l{1pt}r{1pt}){3-5} \cmidrule(l{1pt}r{1pt}){6-8} \cmidrule(l{1pt}r{1pt}){9-11} \cmidrule(l{1pt}){12-14}
      Across & Within & {0-shot} & {1-shot} & {4-shot} & {0-shot} & {1-shot} & {4-shot} & {0-shot} & {1-shot} & {4-shot} & {0-shot} & {1-shot} & {4-shot} \\
      \tabcolmarks{b}
      \midrule
      \tabrowmark{br1T}
      \smash{\makecell[tl]{Separate\\[-0.3ex]\tabonesubfont\emph{as individual}\\[-0.5ex]\tabonesubfont\emph{files}}} & \emph{output-only} & \shotnumcell{-0.5} & \shotnumcell{-3.1} & \shotnumcell{-3.3} & \shotnumcell{-8.1} & \shotnumcell{-8.9} & \shotnumcell{-7.9} & \multicolumn{3}{c}{---} & \shotnumcell{-6.7} & \shotnumcell{-8.0} & \shotnumcell{-9.3} \\
      \tabrowmarktwo{br1B}{br2T}
               & separate          & \shotnumcell{-0.5} & \shotnumcell{-0.8} & \shotnumcell{-2.1} & \shotnumcell{-8.1} & \shotnumcell{-8.5} & \shotnumcell{-9.0} & \multicolumn{3}{c}{---} & \shotnumcell{-6.7} & \shotnumcell{-8.6} & \shotnumcell{-9.1} \\
      \tabrowmarktwo{br2B}{br3T}
               & spatial            & \shotnumcell{-2.8} & \shotnumcell{-9.8} & \shotnumcell{-12.1} & \shotnumcell{-7.0} & \shotnumcell{-1.1} & \shotnumcell{+4.0} & \multicolumn{3}{c}{---} & \multicolumn{3}{c@{\hspace{\tabcolsep}}}{---} \\
      \tabrowmarktwo{br3B}{br4T}
               & temporal           & \shotnumcell{-2.6} & \shotnumcell{-7.5} & \shotnumcell{-6.1} & \shotnumcell{-4.6} & \shotnumcell{-3.5} & \shotnumcell{-6.6} & \shotnumcell{-6.4} & \shotnumcell{-9.1} & \shotnumcell{-10.3} & \multicolumn{3}{c@{\hspace{\tabcolsep}}}{---} \\
      \tabrowmark{br4B}
      \addlinespace[0.8ex]
      \tabrowmark{br5T}
      \smash{\makecell[tl]{Spatial\\[-0.3ex]\tabonesubfont\emph{as a grid}}} & \emph{output-only} & \shotnumcell{-0.5} & \shotnumcell{-7.5} & \shotnumcell{-8.2} & \shotnumcell{-8.1} & \shotnumcell{+1.7} & \shotnumcell{+9.8} & \multicolumn{3}{c}{---} & \shotnumcell{-6.7} & \shotnumcell{-9.6} & \shotnumcell{-4.3} \\
      \tabrowmarktwo{br5B}{br6T}
               & separate          & \shotnumcell{-0.5} & \shotnumcell{-9.9} & \shotnumcell{-7.2} & \shotnumcell{-8.1} & \shotnumcell{-2.8} & \shotnumcell{+9.5} & \multicolumn{3}{c}{---} & \multicolumn{3}{c@{\hspace{\tabcolsep}}}{---} \\
      \tabrowmarktwo{br6B}{br7T}
               & spatial            & \shotnumcell{-2.8} & \shotnumcell{-6.8} & \shotnumcell{-6.4} & \shotnumcell{-7.0} & \shotnumcell{+5.0} & \shotnumcell{+39.7} & \multicolumn{3}{c}{---} & \multicolumn{3}{c@{\hspace{\tabcolsep}}}{---} \\
      \tabrowmarktwo{br7B}{br8T}
               & temporal           & \shotnumcell{-2.6} & \shotnumcell{-6.4} & \shotnumcell{-12.7} & \shotnumcell{-4.6} & \shotnumcell{+24.2} & \shotnumcell{+48.2} & \multicolumn{3}{c}{---} & \multicolumn{3}{c@{\hspace{\tabcolsep}}}{---} \\
      \tabrowmark{br8B}
      \addlinespace[0.8ex]
      \tabrowmark{br9T}
      \smash{\makecell[tl]{Temporal\\[-0.3ex]\tabonesubfont\emph{as a sequence}}} & \emph{output-only} & \shotnumcell{-0.5} & \shotnumcell{-5.3} & \shotnumcell{-4.5} & \shotnumcell{-8.1} & \shotnumcell{-4.0} & \shotnumcell{-11.3} & \shotnumcell{-7.2} & \shotnumcell{-8.0} & \shotnumcell{-8.8} & \shotnumcell{-6.7} & \shotnumcell{-14.6} & \shotnumcell{-8.4} \\
      \tabrowmarktwo{br9B}{br10T}
               & separate          & \shotnumcell{-0.5} & \shotnumcell{-2.1} & \shotnumcell{-3.5} & \shotnumcell{-8.1} & \shotnumcell{-2.7} & \shotnumcell{-7.3} & \shotnumcell{-7.2} & \shotnumcell{-10.4} & \shotnumcell{-10.6} & \multicolumn{3}{c@{\hspace{\tabcolsep}}}{---} \\
      \tabrowmarktwo{br10B}{br11T}
               & spatial            & \shotnumcell{-2.8} & \shotnumcell{-3.9} & \shotnumcell{-4.6} & \shotnumcell{-7.0} & \shotnumcell{+12.7} & \shotnumcell{+9.0} & \multicolumn{3}{c}{---} & \multicolumn{3}{c@{\hspace{\tabcolsep}}}{---} \\
      \tabrowmarktwo{br11B}{br12T}
               & temporal           & \shotnumcell{-2.6} & \shotnumcell{-4.1} & \shotnumcell{-4.8} & \shotnumcell{-4.6} & \shotnumcell{-3.2} & \shotnumcell{-9.6} & \shotnumcell{-7.2} & \shotnumcell{-7.1} & \shotnumcell{-9.4} & \multicolumn{3}{c@{\hspace{\tabcolsep}}}{---} \\
      \tabrowmark{br12B}
      \bottomrule
    \end{tabular*}
    \end{subtable}
  \end{minipage}%
  \end{lrbox}%
  \usebox{\tableftcolbox}%
  \hfill
  \begin{minipage}[t]{120.8pt}
    \begin{subtable}[t]{\linewidth}
      \subtabheader{tab:design_space_formats}{C}{Example formats}{ for 2-shot. \tabbadge{1} is illustrated in \cref{fig:motivating-examples-a}, \tabbadge{2} in \cref{fig:motivating-examples-b}.}
      \tabcpregap
      \noindent\tabdesignspacepanelc
        {``[input\,\itpipe\,solution] examples: \reftt{@image1}, \reftt{@image2}. Edit the yellow right half of \reftt{@video1}.''}%
        {``\reftt{@video1} shows 2 example solutions and the query (bottom-right) in a 2x2 grid. Edit the bottom-right part.''}%
        {``\reftt{@video1}: sequence of example inputs \& query. Extend \reftt{@video2} with the query solution.''}%
    \end{subtable}
  \end{minipage}%
  \hlcellgroup{a}{1}{1}{1}%
  \hlcellgroup{a}{1}{2}{2}%
  \hlcellgroup{a}{2}{3}{3}%
  \hlcellgroup{a}{1}{5}{4}%
  \hlcellgroup{a}{3}{10}{5}%
  \hlcellgroup{b}{1}{1}{1}%
  \hlcellgroup{b}{1}{2}{2}%
  \hlcellgroup{b}{2}{3}{3}%
  \hlcellgroup{b}{1}{5}{4}%
  \hlcellgroup{b}{3}{10}{5}%
\end{table*}

\paragraph{ICL works across spatiotemporal prompt formats}
In language models, few-shot performance is sensitive to the prompt template and example ordering and wording~\citep{zhao2021calibrate,lu2022fantastically,sclar2024quantifying,li2025order}. Visual ICL in image inpainting models likewise depends on the choice and arrangement of examples~\citep{bar2022visual,zhang2023makes,luo2026love}. Given the larger space of spatiotemporal arrangements for video models, one might expect ICL to emerge only in a small subset of prompt formats. Instead, \cref{tab:design_space} shows that on \chromaticscrambleone{} I2I and V2V, \omnione{} 1-shot or 4-shot beats 0-shot across many example packing strategies in all four interaction modes (editing, inpainting, extension, and generation).

\paragraph{1-shot examples resolve task and format ambiguity} Even though the 0-shot prompt explicitly states the transformation rule, \omnione{} fails without visual examples: on I2I, 0-shot error sits above the identity baseline in almost all formats, while a single example lowers error in 31 of 32 cases~(\cref{tab:design_space_image}). Remarkably, a single output-only example is nearly as helpful as an input--output pair, demonstrating that visual examples ground the target distribution and execution of a rule that the model cannot follow from text alone. This mirrors \gptthree{} seeing large boosts from 0 to 1-shot by resolving format ambiguity~\citep{brown2020language}, and parallels \emph{task recognition} in LLM literature: conditioning on the target distribution and format cues disambiguates the task prompt~\citep{min2022rethinking,xie2022explanation,pan2023what,agarwal2024manyshot,gadetsky2025large}.

\paragraph{More examples boost performance, unless they interfere} 4-shot further lowers error over 1-shot in 28 of 32 I2I formats and 20 of 32 V2V formats~(\cref{tab:design_space}). In LLMs, steady few-shot gains on unfamiliar rules are typically associated with \emph{task learning}~\citep{pan2023what,wei2023larger,mao2025survey}, although $k$-shot trends alone cannot separate \emph{task recognition} and \emph{task learning} without controlled ablations~\citep{kossen2024incontext,agarwal2024manyshot,pan2023what}.
Where more examples do not reduce error, we usually see \omnione{} mixing elements from multiple examples, especially in tightly packed formats. In spatially tiled grids, packing $k+1$ cells into a fixed 720p canvas also shrinks each cell's pixel area, which may compound cross-example interference at higher $k$ (though we evaluate all settings at the smallest common resolution for fairness; \cref{app:eval-statistics}). The resulting U-shaped performance has also been observed in related ICL literature~\citep{li2024longiclbench,liu2024lost,rostami2026incontext}.

\takeaway{
\textbf{In-context learning is a general ability} of frontier video models, not limited to a specific choice of prompt formats: in-context examples lower error across spatial, temporal, and separate-file prompt layouts, and whether generating a new video or adapting an existing one (editing, inpainting, extending).
}

\paragraph{Video likelihood scoring: ICL beyond few-shot} Following \citet{lampinen2024broader}, ICL is broader than few-shot learning. A completely different way of adapting to in-context information is likelihood scoring \citep{trinh2018simple,radford2018improving}: evaluating the most likely continuation of an in-context prefix from a fixed set of options. While this article focuses on one-shot and few-shot learning, we demonstrate likelihood scoring as an alternative ICL technique on Wan~2.1 (\cref{app:video_scoring}): the model achieved an overall accuracy of 99.49\% ($p < .001$) on a three-way classification task where it is asked to determine which of three camera positions is the most likely continuation for a given scene.

\begin{figure*}[t!]
    \centering
    \drawqualitativevisiongrid
        {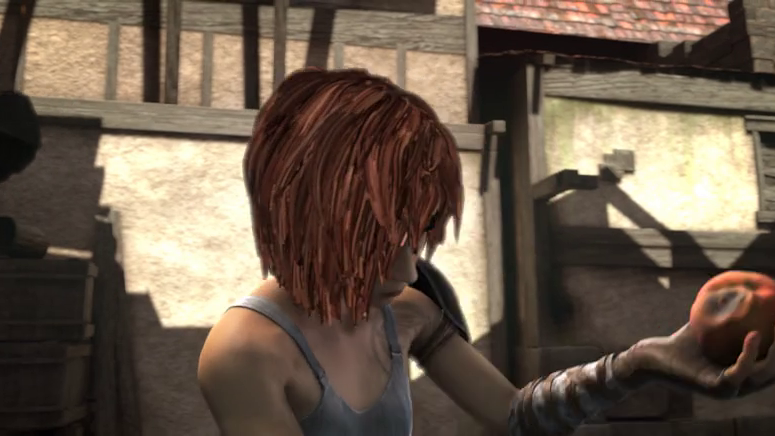}
        {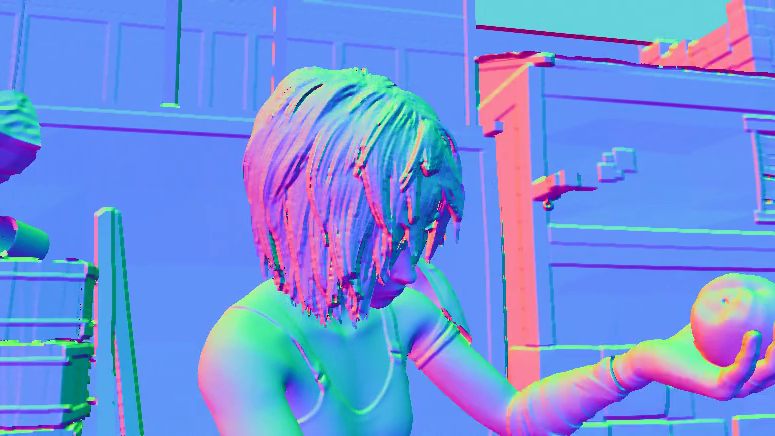}
        {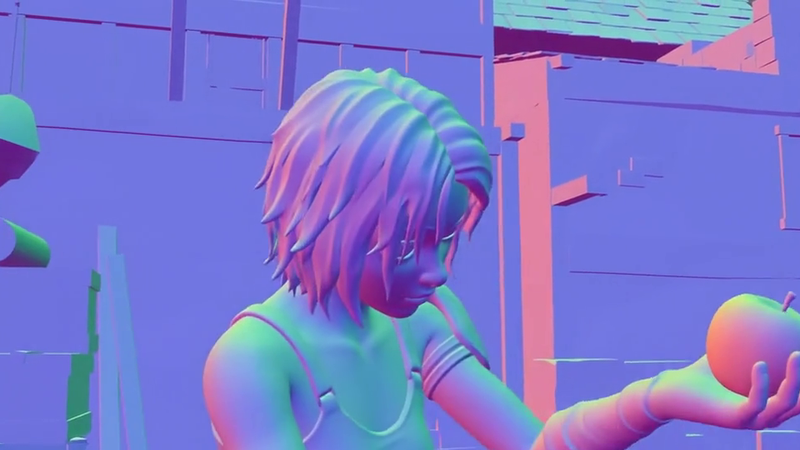}
        {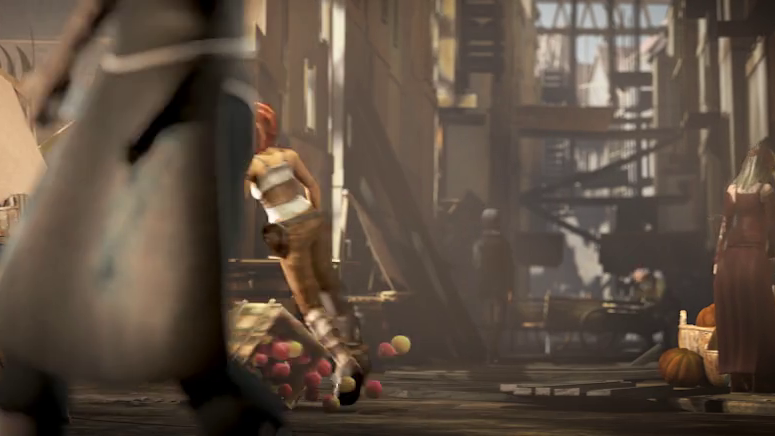}
        {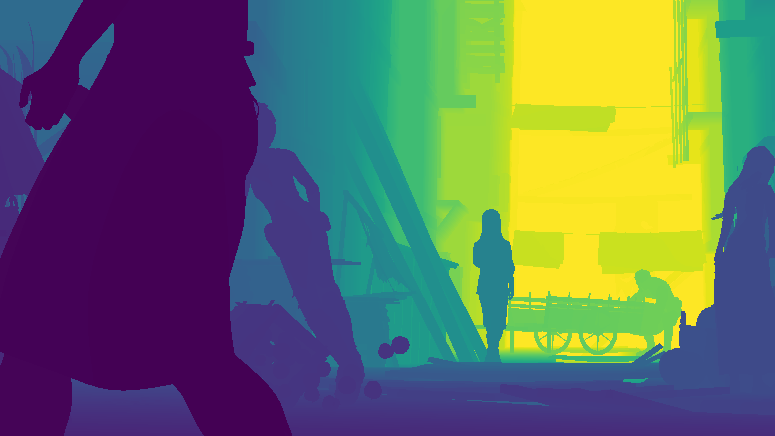}
        {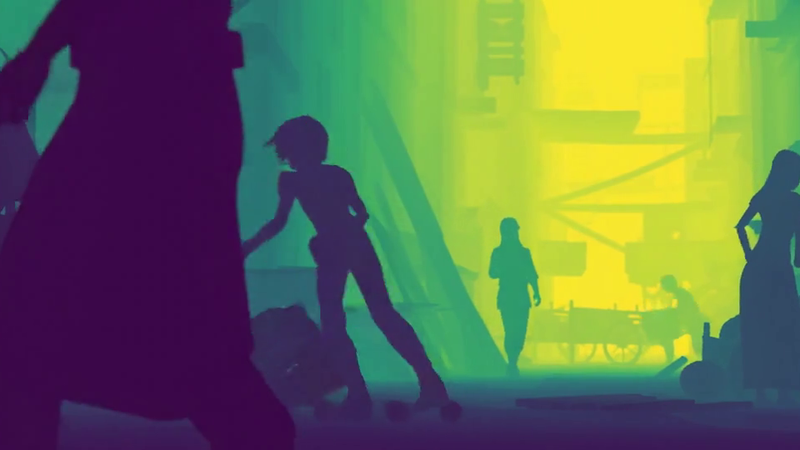}
        {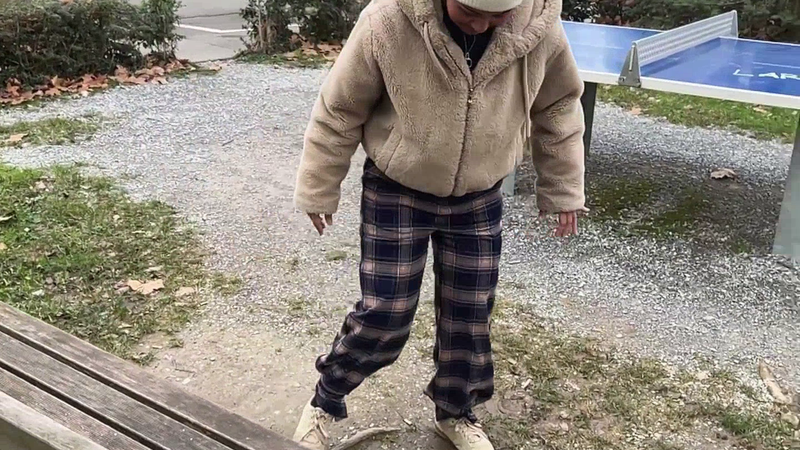}
        {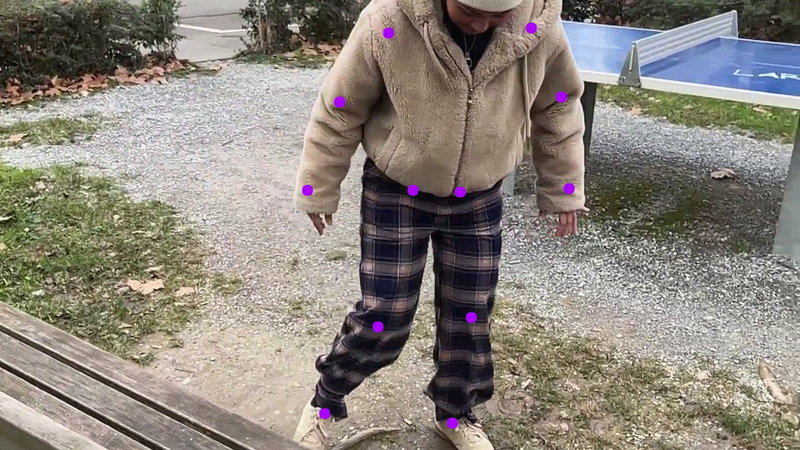}
        {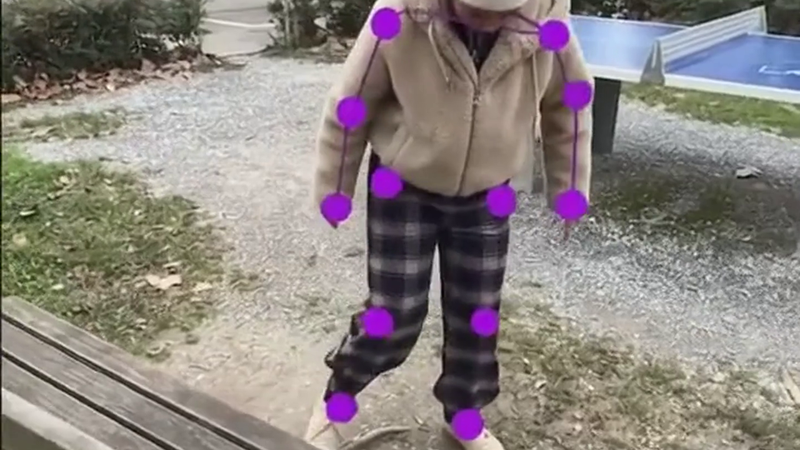}
        {}
        {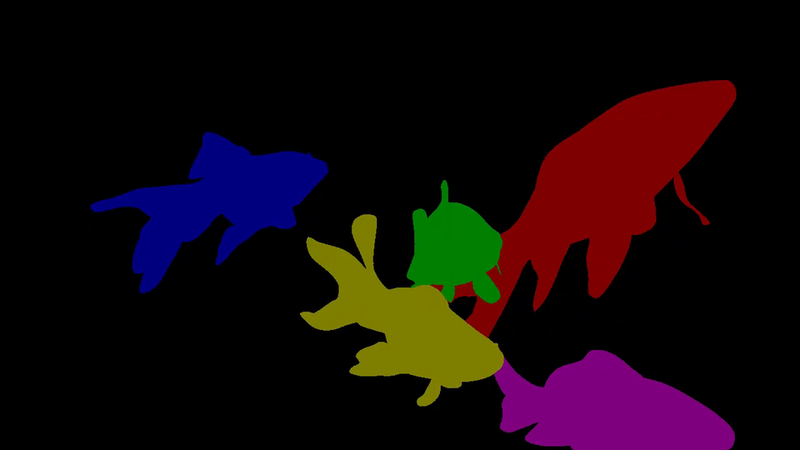}
        {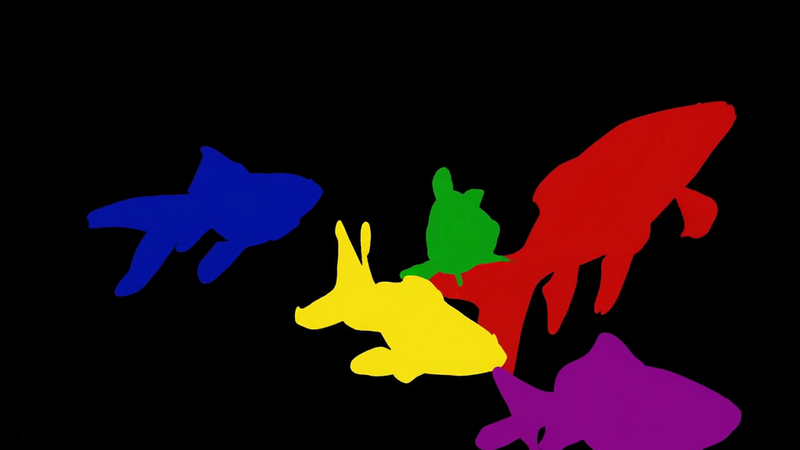}
    \caption{\textbf{Qualitative few-shot predictions alongside ground truth for \omnione{} across the computer vision tasks from \cref{fig:all_tasks_4x4}} ($k=5$ for \textbf{A}--\textbf{C}; \textsuperscript{\textasteriskcentered}$k=1$ for \textbf{D} since the dataset is one-shot only).}
    \label{fig:qualitative_vision_omni}
\end{figure*}

\section{Which vision tasks benefit from video in-context learning?}
\label{sec:results}

\begin{figure*}[t!]
    \centering
    \renewcommand{\thesubfigure}{\Alph{subfigure}}%
    \begin{subfigure}[t]{234.27pt}
        \hbox{\hspace{0.19in}{\phantomsubcaption\label{fig:surface_normal}}\figfont\fontsize{8.2pt}{9.8pt}\selectfont\textcolor{plotspinecolor}{\textbf{A}\hspace{0.55em}Surface normal estimation (MPI~Sintel, V2V)}}
        \vspace{2.0pt}
        \hbox{\hspace{0.19in}\taskillustration{``Perform surface normal estimation, maintaining camera motion.''}{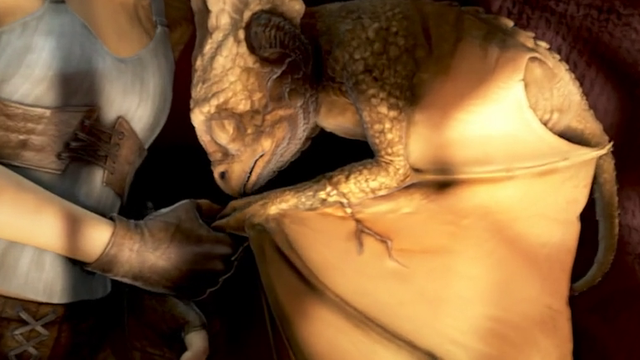}{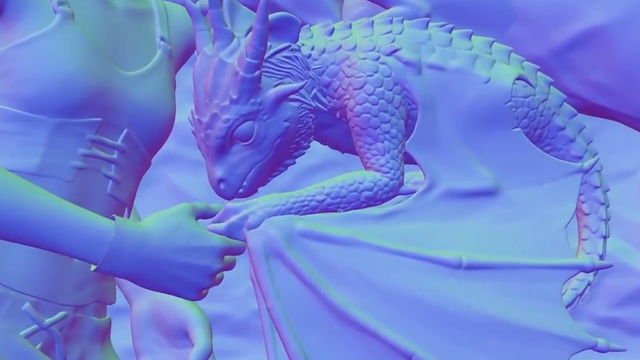}{1}}
        \vspace{3.0pt}
        \hbox to \linewidth{%
            \includegraphics[scale=1.0]{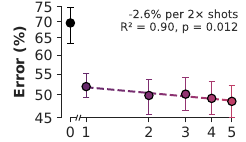}%
            \hfill
            \includegraphics[scale=1.0]{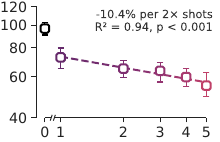}%
        }
    \end{subfigure}
    \hfill
    \begin{subfigure}[t]{222.93pt}
        \hbox{\hspace{0.03in}{\phantomsubcaption\label{fig:depth_estimation}}\figfont\fontsize{8.2pt}{9.8pt}\selectfont\textcolor{plotspinecolor}{\textbf{B}\hspace{0.55em}Monocular depth estimation (MPI~Sintel, V2V)}}
        \vspace{2.0pt}
        \hbox{\hspace{0.03in}\taskillustration{``Perform depth estimation, maintaining camera motion.''}{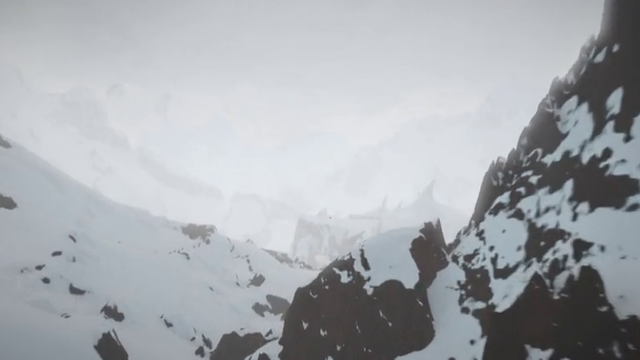}{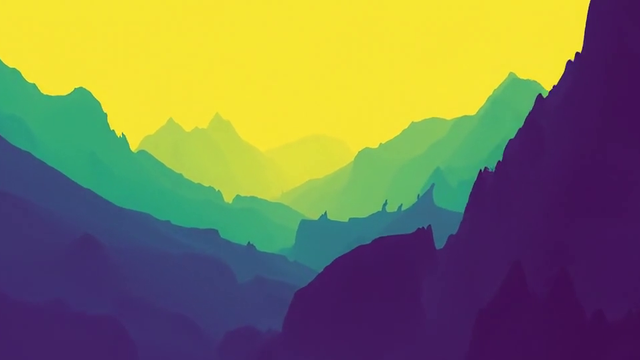}{1}}
        \vspace{3.0pt}
        \hbox to \linewidth{%
            \includegraphics[scale=1.0]{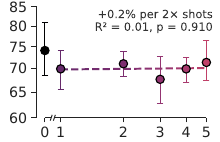}%
            \hfill
            \includegraphics[scale=1.0]{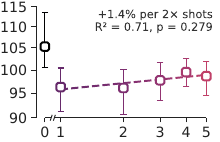}%
        }
    \end{subfigure}
    \par\vspace{4.5pt}
    \begin{subfigure}[t]{234.27pt}
        \hbox{\hspace{0.19in}{\phantomsubcaption\label{fig:pose_estimation}}\figfont\fontsize{8.2pt}{9.8pt}\selectfont\textcolor{plotspinecolor}{\textbf{C}\hspace{0.55em}Human pose estimation (EMDB~1, V2V)}}
        \vspace{2.0pt}
        \hbox{\hspace{0.19in}\taskillustration{``Overlay circles on the 12 body joints, tracking them across frames.''}{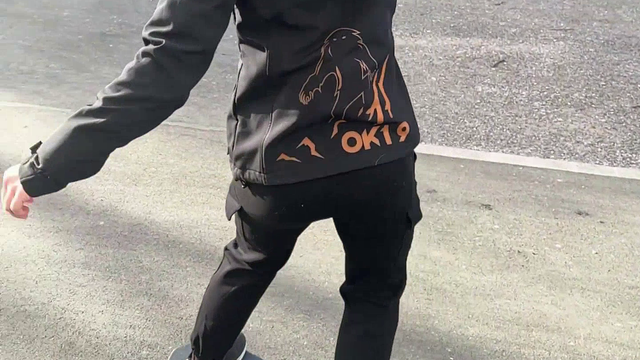}{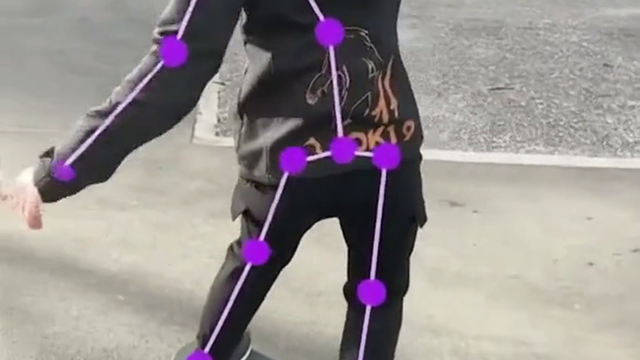}{1}}
        \vspace{3.0pt}
        \hbox to \linewidth{%
            \includegraphics[scale=1.0]{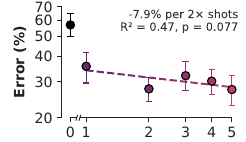}%
            \hfill
            \includegraphics[scale=1.0]{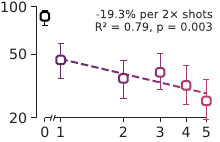}%
        }
    \end{subfigure}
    \hfill
    \begin{subfigure}[t]{222.93pt}
        \hbox{\hspace{0.03in}{\phantomsubcaption\label{fig:segmentation_davis}\label{tab:segmentation_davis}}\figfont\fontsize{8.2pt}{9.8pt}\selectfont\textcolor{plotspinecolor}{\textbf{D}\hspace{0.55em}Video object segmentation (DAVIS 2017, V2V)}}
        \vspace{2.0pt}
        \hbox{\hspace{0.03in}\taskillustration{``Segment the objects, each with its own color. Black background.''}{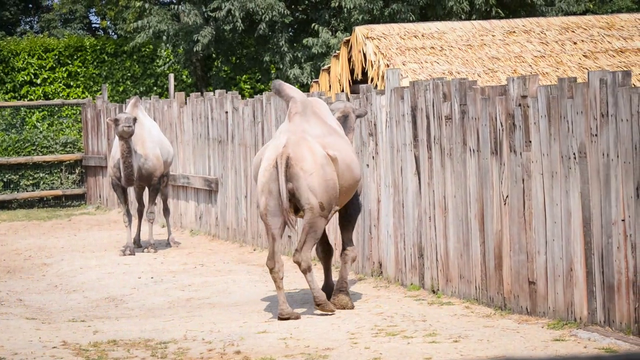}{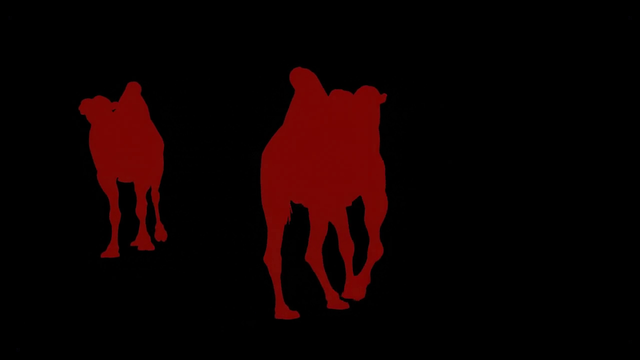}{1}}
        \vspace{3.0pt}
        \hbox to \linewidth{%
            \includegraphics[scale=1.0]{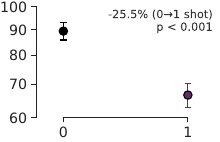}%
            \hfill
            \includegraphics[scale=1.0]{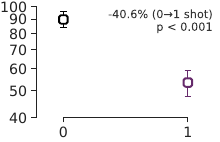}%
        }
    \end{subfigure}
    \par\vspace{4.5pt}
    \begin{subfigure}[t]{234.27pt}
        \hbox{\hspace{0.19in}{\phantomsubcaption\label{fig:maze_solving}}\figfont\fontsize{8.2pt}{9.8pt}\selectfont\textcolor{plotspinecolor}{\textbf{E}\hspace{0.55em}Maze solving, I2V}}
        \vspace{2.0pt}
        \hbox{\hspace{0.19in}\taskillustration{``The person walks through the maze to the cyan circle, trailing a red line.''}{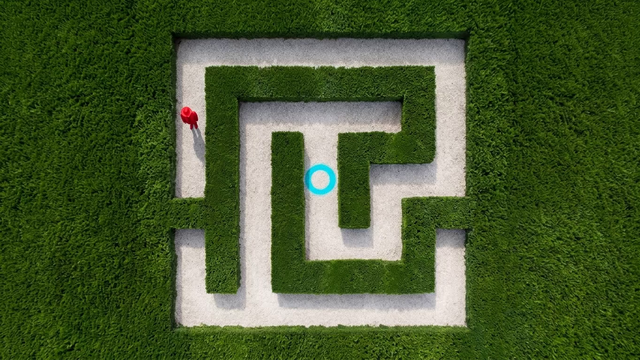}{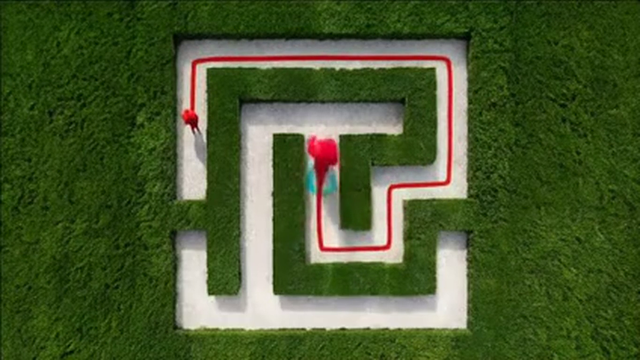}{0}}
        \vspace{3.0pt}
        \hbox to \linewidth{%
            \includegraphics[scale=1.0]{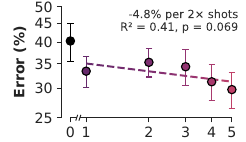}%
            \hfill
            \includegraphics[scale=1.0]{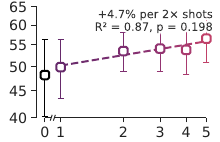}%
        }
    \end{subfigure}
    \hfill
    \begin{subfigure}[t]{222.93pt}
        \hbox{\hspace{0.03in}{\phantomsubcaption\label{fig:vpct}}\figfont\fontsize{8.2pt}{9.8pt}\selectfont\textcolor{plotspinecolor}{\textbf{F}\hspace{0.55em}Physical simulation (VPCT, I2V)}}
        \vspace{2.0pt}
        \hbox{\hspace{0.03in}\taskillustration{``The ball falls and rolls down in a physically plausible way.''}{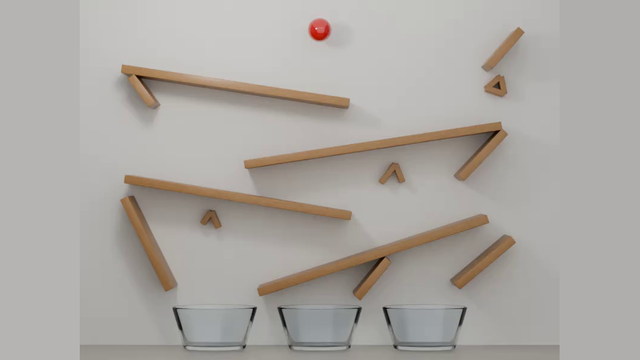}{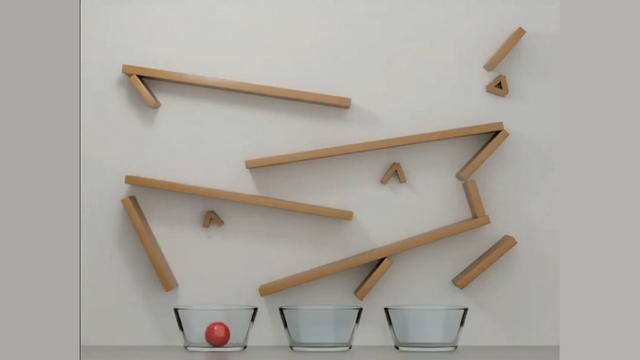}{0}}
        \vspace{3.0pt}
        \hbox to \linewidth{%
            \includegraphics[scale=1.0]{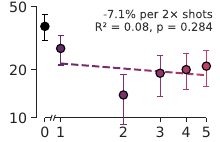}%
            \hfill
            \includegraphics[scale=1.0]{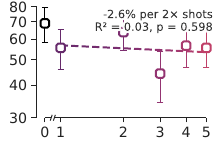}%
        }
    \end{subfigure}
    \par\vspace{4.5pt}
    \begin{subfigure}[t]{234.27pt}
        \hbox{\hspace{0.19in}{\phantomsubcaption\label{fig:vbvr_pro}}\figfont\fontsize{8.2pt}{9.8pt}\selectfont\textcolor{plotspinecolor}{\textbf{G}\hspace{0.55em}Visual reasoning (VBVR-Pro, I2V)}}
        \vspace{2.0pt}
        \hbox{\hspace{0.19in}\taskillustration{E.g., ``fade out all objects not filled with the most frequent color.''}{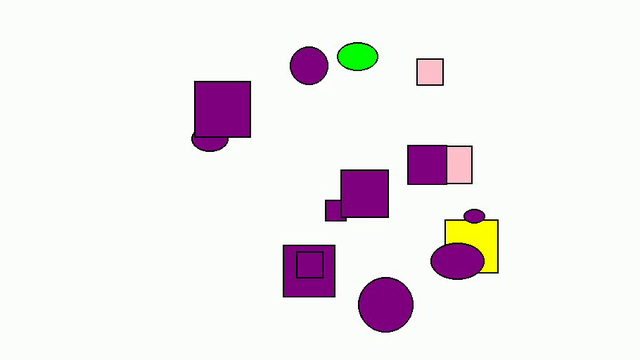}{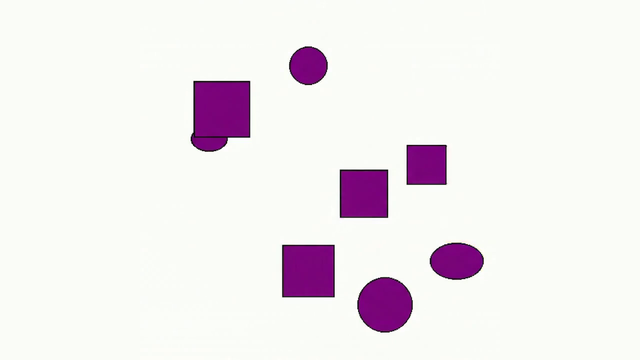}{0}}
        \vspace{3.0pt}
        \hbox to \linewidth{%
            \includegraphics[scale=1.0]{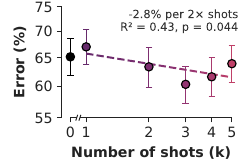}%
            \hfill
            \includegraphics[scale=1.0]{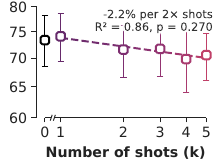}%
        }
    \end{subfigure}
    \hfill
    \begin{subfigure}[t]{222.93pt}
        \hbox{\hspace{0.03in}{\phantomsubcaption\label{fig:cellular_automata}}\figfont\fontsize{8.2pt}{9.8pt}\selectfont\textcolor{plotspinecolor}{\textbf{H}\hspace{0.55em}Cellular automata, I2V}}
        \vspace{2.0pt}
        \hbox{\hspace{0.03in}\taskillustration{E.g., ``over 7 steps, a cell turns blue if its upper or left neighbor was blue.''}{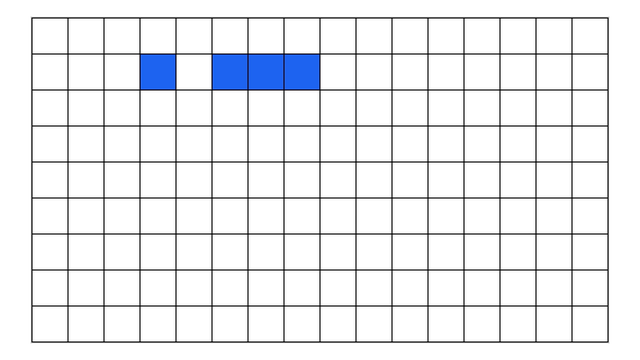}{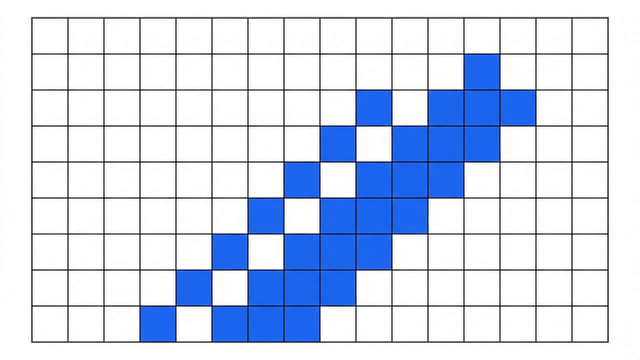}{0}}
        \vspace{3.0pt}
        \hbox to \linewidth{%
            \includegraphics[scale=1.0]{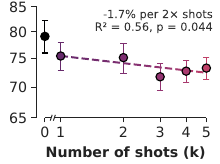}%
            \hfill
            \includegraphics[scale=1.0]{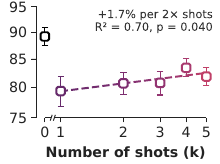}%
        }
    \end{subfigure}
    \caption{\textbf{In-context learning across computer vision and visual reasoning tasks for \omnilinemarker~\omnione{} and \seedancelinemarker~\seedancetwofive{}.} Error (\% of no-op baseline for \textbf{A}--\textbf{D}; percentage error rate for \textbf{E}--\textbf{H}) across $k$ in-context examples. Markers show the interquartile mean (IQM; \textbf{A}--\textbf{C}) or mean (\textbf{D}--\textbf{H}) with within-query clustered $95\%$ confidence intervals; dashed lines show power-law fits for $k \ge 1$. Prompts are paraphrased (\textbf{G}, \textbf{H} span 50 tasks and 5 rules; details in \cref{app:task_details}); illustrations show the query and last Omni-generated frame.\website{ See \href{https://video-in-context-learning.github.io/}{website} for interactive visualization.}}
    \label{fig:all_tasks_4x4}
\end{figure*}

Having explored the prompt design space on our controlled example task with \omnione{}, we now evaluate both \omnione{} and \seedancetwofive{} on four computer vision tasks (\cref{fig:surface_normal,fig:depth_estimation,fig:pose_estimation,fig:segmentation_davis}) and four visual reasoning tasks (\cref{fig:maze_solving,fig:vpct,fig:vbvr_pro,fig:cellular_automata}). For each task, a suitable prompt format is selected from \cref{tab:design_space} (specified in \cref{app:prompts}; task details in \cref{app:task_details}).

\paragraph{Computer vision tasks}
\emph{Surface normal estimation} (\cref{fig:surface_normal}) asks models to edit MPI~Sintel~\citep{butler2012naturalistic} videos into RGB-encoded normal maps whose coordinate convention is only specified through examples, scored by mean angular error (\cref{app:surface_normal}).
\emph{Monocular depth estimation} (\cref{fig:depth_estimation}) asks models to edit the same Sintel videos into Viridis-encoded depth maps, scored by scale- and shift-aligned absolute relative error (\cref{app:depth_estimation}).
\emph{Human pose estimation} (\cref{fig:pose_estimation}) requires tracking and overlaying purple circles on 12 body joints from EMDB~1~\citep{kaufmann2023emdb} videos, scored by a clamped mean joint error that penalizes missed joints (\seedancetwofive{} uses face-blurred clips due to API face-filter refusals on unblurred videos; \cref{app:pose_estimation}).
\emph{Video object segmentation} (\cref{fig:segmentation_davis}) follows the semi-supervised DAVIS~2017~\citep{pont20172017} protocol, where models segment target objects across a video given the query video's own first-frame mask as a single example ($k \le 1$), scored by segmentation error $1 - \mathcal{J}\&\mathcal{F}$ (\cref{app:segmentation}).

\paragraph{Visual reasoning tasks}
\emph{Maze solving} (\cref{fig:maze_solving}) requires animating a person in red navigating from start to goal through $9 \times 9$ mazes~\citep{wiedemer2026video} rendered as photorealistic top-down hedge mazes~\citep{geirhos2026visual}, scored by path error $1 - \text{Path IoU}$ (\cref{app:maze_solving}).
\emph{Physical simulation} (\cref{fig:vpct}) asks models to simulate a ball rolling across angled obstacles into one of three containers on \vpctblender{}, a Blender-rendered version of VPCT~\citep{camelcase2025vpct,geirhos2026visual}, scored by classification error rate (\cref{app:vpct}).
\emph{Visual reasoning} (\cref{fig:vbvr_pro}) evaluates 50 tasks from VBVR-Pro~\citep{wang2026very,xu2026vbvr} using deterministic rule-based verifiers, scored by continuous task error $1 - \text{task score}$ (\cref{app:vbvr_pro}).
\emph{Cellular automata} (\cref{fig:cellular_automata}) tests multi-step simulation of five synthetic 2D binary transition rules over 7 discrete steps on a $9 \times 16$ grid, scored by step-wise rollout error $1 - \text{IoU}$ (\cref{app:cellular_automata}).

\paragraph{Key results} \cref{fig:all_tasks_4x4} shows that both \omnione{} and \seedancetwofive{} are in-context learners: for most tasks, there is an in-context setting that outperforms the zero-shot setting. Sometimes, a single example (one-shot) already achieves a large error reduction; sometimes more in-context examples help further reduce the error. Qualitative examples for the computer vision tasks are shown in \cref{fig:qualitative_vision_omni}.

\paragraph{One-shot vs.\ few-shot learning}
As in LLMs and VLMs, we do not expect examples to reduce error on every task, as ICL is understood to operate within the pretraining support rather than conferring entirely new capabilities~\citetext{\citealp{yadlowsky2023pretraining,li2026littlelearner,halady2026quo}; \citealp[cf.][]{wiedemer2023compositional,wiedemer2024pretraining}}.
Where ICL succeeds (\cref{fig:all_tasks_4x4}), scaling mirrors the empirical regimes known from language models and observed in \cref{sec:design-space}:
Where the main challenge is \emph{specifying} the desired output format, a single example ($k=1$) accounts for most of the error reduction, consistent with \emph{task recognition}: the RGB color convention in surface normals (\cref{fig:surface_normal}; confirmed by rotation-aligned evaluation in \cref{app:surface_normal}), the target object identity in video segmentation (\cref{fig:segmentation_davis}), and the joint marker appearance and side-by-side inpainting layout in pose estimation (\cref{fig:pose_estimation}) are hard to describe purely in text but immediately disambiguated by one visual example.
Where the model must instead infer or refine an underlying rule, error decreases more gradually across multiple shots $k \ge 1$, consistent with \emph{task learning}---as observed for \omnione{} on cellular automata (\cref{fig:cellular_automata}), mazes (\cref{fig:maze_solving}, reaching significance vs.\ $0$-shot at $k=5$), and VBVR-Pro (\cref{fig:vbvr_pro}). Both regimes also co-occur: after a sharp 1-shot drop, additional examples continue to significantly lower error on surface normals and pose estimation, particularly for \seedancetwofive{} (\cref{fig:surface_normal,fig:pose_estimation}).

\paragraph{Computer vision vs.\ visual reasoning}
On three of the four vision tasks (\cref{fig:surface_normal,fig:pose_estimation,fig:segmentation_davis}), examples substantially reduce error for both models (with a smaller reduction on depth estimation; \cref{fig:depth_estimation}). On visual reasoning, gains are less consistent: on VBVR-Pro (\cref{fig:vbvr_pro}), error decreases only slightly, with $95\%$ confidence intervals still overlapping $0$-shot at $k=5$. Rendering reasoning tasks realistically helps: following the realism bias of video models~\citep{geirhos2026visual}, we present mazes and VPCT as photorealistic scenes; both tasks benefit from examples. Cellular automata (\cref{fig:cellular_automata}), by contrast, are abstract sketch-like 2D grids; interestingly, both models still improve over $0$-shot with examples even though the exact transition rule is already stated in the text prompt.
At the same time, reasoning tasks frequently exhibit \emph{U-shaped scaling curves}: error reaches a minimum at two or three examples on VPCT (\cref{fig:vpct}) and rebounds after $k=1$ for \seedancetwofive{} on cellular automata (\cref{fig:cellular_automata}) and across $k$ on mazes (\cref{fig:maze_solving}). Mirroring the cross-example interference in \cref{sec:design-space} and long-context ICL in LLMs~\citep{li2024longiclbench,liu2024lost,rostami2026incontext}, adding examples can hurt when models blend scene layouts or dynamics from multiple references, or, in spatially tiled tasks like mazes and VPCT, when tiling $k+1$ cells into a fixed 720p canvas reduces the query cell's pixel resolution (\cref{app:eval-statistics}).

\paragraph{\omnione{} vs.\ \seedancetwofive{}}
Although \seedancetwofive{} in principle supports more reference files and therefore separate in-context examples than \omnione{}, \omnione{} appears to be the stronger in-context learner overall across the evaluated shot counts. \omnione{} achieves lower absolute error on most tasks (surface normals, depth, mazes, VPCT, VBVR-Pro, and cellular automata) and more consistently improves with additional examples, whereas \seedancetwofive{} is more prone to cross-example interference at higher $k$. Since \seedancetwofive{} often starts near the $100\%$ no-op baseline at $0$-shot (e.g., on surface normals and pose estimation), it sees larger initial 1-shot drops, outperforming \omnione{} on 1-shot video object segmentation (\cref{fig:segmentation_davis}) and reaching comparable error on 5-shot pose estimation (\cref{fig:pose_estimation}). Overall, both models have their respective strengths, and in light of the fact that these models are developed independently, these results provide strong evidence for in-context learning as a general ability of frontier video models.

\takeaway{
Frontier \textbf{video models are in-context learners} across both computer vision and visual reasoning tasks.
}

\section{Combining in-context learning with best-of-N sampling}
\label{sec:context_bon_interaction}

\begin{figure*}[t]
    \centering
    \begin{subfigure}[t]{\twofigwidth\textwidth}
        \caption{\opensans Superadditive scaling: In-context examples and Best-of-N}
    \end{subfigure}
    \hfill
    \begin{subfigure}[t]{\twofigwidth\textwidth}
        \caption{\opensans Power law exponents: More context $\rightarrow$ steeper power law}
    \end{subfigure}\par
    \begin{minipage}[b]{\twofigwidth\textwidth}
        \centering
        \includegraphics[width=\textwidth]{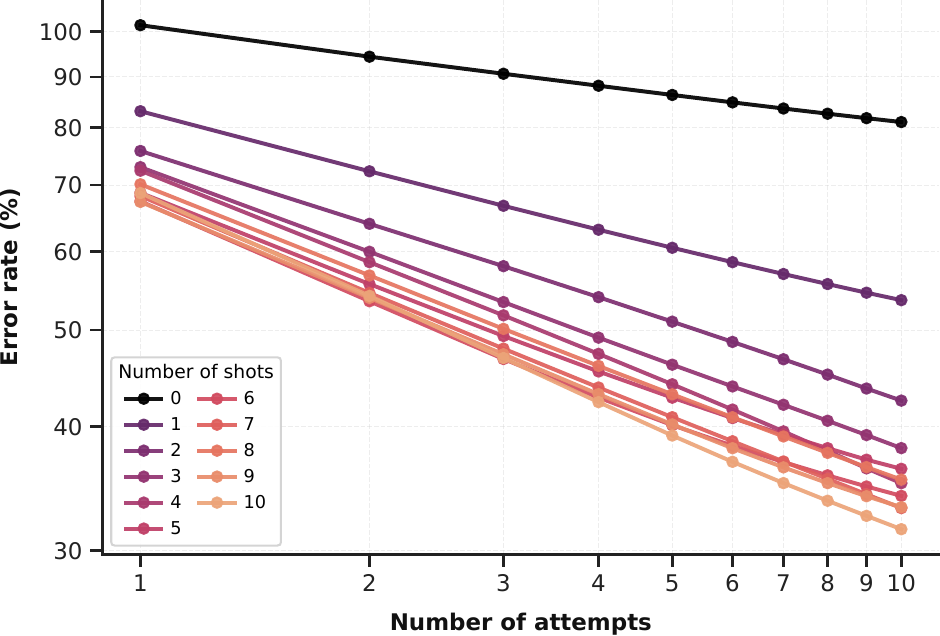}
    \end{minipage}
    \hfill
    \begin{minipage}[b]{\twofigwidth\textwidth}
        \centering
        \includegraphics[width=\textwidth]{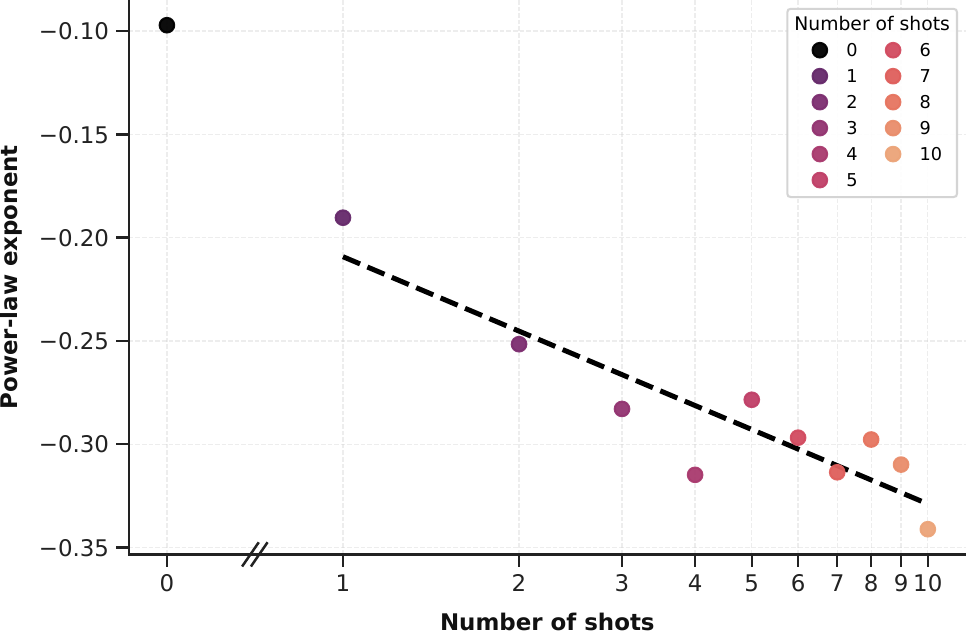}
    \end{minipage}
    \caption{\textbf{Combining in-context learning with Best-of-N leads to superadditive scaling.}\\
    \emph{Left:} Error rates reliably decrease with more attempts (oracle Best-of-N selection, upper bound) and with more in-context examples (= larger number of shots). Results here for \omnione{} on \chromaticscrambleone{}; similar plots for \seedancetwofive{} and another task in \cref{fig:double_scaling_big_panel} and details in \cref{app:best_of_n_methods}. \emph{Right:} When fitting a power law to each of the curves from the left plot, a larger number of shots leads to a steeper power law (lower power-law exponent). Thus, co-scaling attempts and in-context examples results in superadditive scaling: in-context examples make Best-of-N more effective.}
    \label{fig:double_scaling}
\end{figure*}

Whether in language or vision, ICL is a general technique that can help models perform tasks better. Another widely used technique is spending test-time compute to improve results~\citep[e.g.,][]{snell2024scaling}. Here we ask: How compatible are these two approaches for video models? A priori, the best outcome one could hope for is that their effects are additive. If that is the case, then one can expand context size (more in-context samples), and also spend more test-time compute (by sampling more), for the best possible results. On the other hand, it is also conceivable that they interact negatively. In that case, one may have to decide between using more context vs.\ more test-time compute, since their respective benefits might cancel each other out.

\paragraph{Methods} To understand the interaction between in-context learning and test-time compute, we again use the chromatic scramble task as a controlled testbed, since it is hard and by construction not part of the training data. For test-time compute, we evaluate \emph{oracle best-of-N} (BoN), which selects the attempt minimizing MSE against the ground-truth target out of $N$ attempts and averages this expected minimum across queries (\cref{app:best_of_n_methods}). While practical deployment requires a test-time verifier (e.g., a VLM judge or self-consistency), oracle BoN traces the upper bound that no verifier can exceed, making it the most principled choice for our evaluation.

\paragraph{Results} The resulting trends are visualized in \cref{fig:double_scaling}. Error rate predictably decreases with more attempts (best-of-N test-time compute), and also with larger context (more shots). Interestingly, these two approaches interact such that the larger the context, the more effective best-of-N becomes (and vice versa): the resulting pattern is superadditive. Fitting a power law to each shot-count curve (\cref{fig:double_scaling}, right) shows that the scaling exponent becomes steeper with more context across models and datasets (\cref{fig:double_scaling_big_panel}). Thus, the two techniques show a clear synergy, and can be combined to obtain superadditive performance.

\takeaway{On a controlled task (ruling out task familiarity as a confound), \textbf{video in-context learning and best-of-N sampling are superadditive}: the more examples in context, the more effective inference-time sampling becomes.}

\section{Discussion}
\paragraph{Summary} Without weight updates, adapters, or architectural changes, off-the-shelf frontier video models can successfully infer patterns from one-shot and few-shot examples, and apply these patterns to solve visual tasks better than through zero-shot prompting alone. This works across many prompt formats~(\cref{sec:design-space}) and a broad range of perception and reasoning tasks~(\cref{sec:results}). On a controlled task (ruling out task familiarity as a confound), more in-context examples make best-of-N sampling more effective~(\cref{sec:context_bon_interaction}), resulting in superadditive performance. Taken together, both \omnione{} and \seedancetwofive{} can harness their multimodal context windows to improve task performance. Like LLMs, generative video models are in-context learners.

\paragraph{Open questions and future directions} Our finding of in-context learning (ICL) in off-the-shelf video models raises many questions for future work: What is the mechanism behind these results? Are there repeated patterns in the training data that facilitate video ICL, similar to text data~\citep{yan2024understanding,schirrmeister2026unlocking}? Do the same data distributional properties and architecture components~\citep{chan2022data} enable both language- and video-based ICL? Would video ICL emerge in a video-only model, or is the multimodal nature of the models we study key? How will video ICL scale from few-shot to the ``many-shot'' regime~\citep{agarwal2024manyshot}? At which model scale and training stage does video ICL become effective, and how can it be improved? Can datasets be constructed to improve ICL abilities when included in the training mix~\citep{gu2023pre}? If the parallel to language ICL holds, do ICL failures transfer too, such as sensitivity to example ordering~\citep{lu2022fantastically,li2025order}?

\paragraph{Limitations}
Large multimodal context windows are a relatively new development; thus multimodal ICL can only be studied in models that have one, currently excluding several video models. Extrapolating current trends, capable open-weight video models with large multimodal context windows will likely emerge soon; for now, our analysis focuses on two frontier models that push this boundary first. Furthermore, video generation is expensive, and for many tasks we study, capable specialized models (e.g., for depth prediction) are readily available. The motivation for studying video models as general-purpose in-context learners, despite lagging behind specialized models, is that computer vision will never reach a unified foundation model by building only specialized models. Which approach (e.g., video models, VLMs, ...) will lead to a unified vision foundation model remains open, but learning from in-context examples (which most specialized models lack, being limited to training tasks) could be an important step toward that goal.

\paragraph{Outlook}
Our findings of in-context learning in video models provide evidence for a deep parallel between language and video models. This convergence raises a broader question: does intelligence in generative models follow a common developmental path, irrespective of the modality? And if that is indeed the case, which stage of development comes next for video models, now that they've followed in the footsteps of LLMs for zero-shot prompting~\citep[cf.][]{wiedemer2026video} and few-shot in-context learning?

\section*{Acknowledgements}
The authors would like to thank (in alphabetic order): Andrei Zanfir, Andrew Zisserman, Drew Hudson, Eduard Gabriel Bazavan, Effie Li, Joao Carreira, John Hewitt, Jon Shlens, Karsten Roth, Mehdi Sajjadi, Mike Mozer, Misha Andriluka, Rahul Sukthankar, Viorica Patraucean for helpful discussions and feedback; Fernanda Viegas, Martin Wattenberg, and Noah Fiedel for their invaluable support with the project including its budget; the entire Omni team; and Junlin Zhang for PM support.

\bibliographystyle{unsrtnat}
\bibliography{refs}

\newpage
\appendix
\section*{Appendix}
\appendixtableofcontents
\bigskip

\section{Inference details}
\label{app:inference-details}
\citet{aavideo2026leaderboard} reports inference costs of \$9.64 per minute of generated video for \omnione{} and \$40.82 per minute for \seedancetwofive{}. We use the Google Cloud Vertex AI API for \omnione{}~\citep{deepmind2026geminiomni,google2026geminiomni11} and the Replicate API~\citep{aavideo2026leaderboard} for \seedancetwofive{}~\citep{seedance2026seedance25}, where it is billed at \$58.06 per minute. Both models are evaluated at 720p resolution ($1280 \times 720$ for $16{:}9$ tasks) and $24\,\text{fps}$ during October--November 2026. %

\section{General evaluation and statistics details}
\label{app:eval-statistics}
\paragraph{Image queries} For tasks with image inputs (e.g., \chromaticscrambleone{}~I2I in \cref{sec:approach}, other transforms in \cref{app:image_and_video_transforms}, and most visual reasoning tasks), we  present the query image as a static video, either as a standalone file or embedded into a spatial grid or temporal sequence (see \cref{sec:design-space}).

\paragraph{Spatiotemporal query regions} For spatially or temporally packed prompt formats (\cref{sec:design-space}), we extract and score only the spatiotemporal region containing the query that the model was instructed to generate or modify. Since the resolution of this region depends on the exact prompt format and the number of shots (e.g., \cref{tab:design_space} and tasks in \cref{fig:all_tasks_4x4} using spatial packing), we evaluate all settings at the smallest common resolution using Lanczos downscaling to remove evaluation resolution as a confounder. We tile $k+1$ videos into a fixed $1280 \times 720$ spatial grid ($1\times 1$ at $k=0$, $1\times 2$ at $k=1$, $2\times 2$ at $k \in \{2, 3\}$, and $2\times 3$ at $k \in \{4, 5\}$); cell-content is padded with black to maintain its aspect ratio, and empty cells are filled with black. Spatial and temporal packing of reduces the spatiotemporal pixel volume and latent token budget allocated to the query cell during generation, possibly explaining cross-example interference at higher shot counts. Some prompts instruct the model to edit the query region after, e.g., 1 second; this prefix is trimmed during evaluation. In the case of \seedancetwofive{}, video editing requires at least a 4-second input video; if the query video is shorter, the last frame is repeated until this 4-second window is filled (this extended part, of course, is removed again before evaluation).

\paragraph{No-op baseline \& normalization} For MSE in image and video transforms in \cref{sec:approach,sec:design-space,app:image_and_video_transforms} and metrics of other computer vision tasks in \cref{sec:results,app:segmentation,app:surface_normal,app:depth_estimation,app:pose_estimation}, raw errors vary substantially across scenes. For a common, interpretable scale, we normalize prediction errors by the error of the unmodified query (no-op baseline). For multi-frame video evaluations, we normalize frame-by-frame before averaging.

\paragraph{Aggregation across queries and attempts} For unbounded metrics (e.g., normalized MSE in image and video transforms which can be $>100\%$, and \cref{fig:all_tasks_4x4}\textbf{A}--\textbf{C}), we report the interquartile mean~\citep[IQM;][]{agarwal2021deep} over queries (different task samples) and attempts (multiple model calls for the same query with different seeds), i.e., the $25\%$ symmetric trimmed mean that discards the lowest $25\%$ and highest $25\%$ of observations and averages the central $50\%$. The IQM is robust to occasional extreme generation failures in the upper tail while remaining resistant to lucky single-attempt outliers in the lower tail. For metrics bounded in $[0, 1]$ and binary error rates (\cref{fig:all_tasks_4x4}\textbf{D}--\textbf{H}: video object segmentation and all reasoning tasks; plotted as $\%$ of the no-op baseline in \textbf{D}, as $0$--$100\%$ in \textbf{E}--\textbf{H}, and on the raw $[0, 1]$ scale in supplementary figures), we report the standard empirical mean instead, as IQM trims asymmetrically and experiences discrete jumps when data points bunch up at either bound.

\paragraph{Completion rates and missing generations} Across all experiments in \cref{sec:results}, a small fraction of API requests fail to return a video due to transient backend timeouts or automated safety-filter rejections (\cref{tab:completion_rates}). Overall completion rates are high across tasks ($85.3\%$--$100\%$ for \omnione{} and $90.9\%$--$100\%$ for \seedancetwofive{} after face-blurring on EMDB~1), and missing generations are distributed nearly uniformly across shot counts (for example, $82/85$ clips at every $k \in \{0, \dots, 5\}$ for \seedancetwofive{} on pose estimation, $275/300$ at $k=0$ vs.\ $276/300$ at $k=1$ for \omnione{} on video object segmentation, and $96/100$ at $k=5$ vs.\ $100/100$ at $k \in \{0, \dots, 4\}$ for \omnione{} on cellular automata). Consequently, imputing missing generations
with the $100\%$ no-op / worst-case error rate leaves all reported $k$-shot scaling trends and significance conclusions unchanged.

\begin{table}[H]
  \centering\footnotesize
  \setlength{\tabcolsep}{3.5pt}
  \caption{\textbf{Completed / requested video counts across tasks in \cref{fig:all_tasks_4x4}.} All tasks use shots $k \in \{0, \dots, 5\}$, except video object segmentation ($k \in \{0, 1\}$). Missing generations arise from transient API timeouts or safety-filter triggers and occur at similar rates across $k$. \textsuperscript{$\ast$}Face-blurred on EMDB~1.
  }
  \label{tab:completion_rates}
  \begin{tabular}{lrr}
    \toprule
    \textbf{Task} & \textbf{\omnione{}} & \textbf{\seedancetwofive{}} \\
    \midrule
    Surface normal est.       & $274 / 276$ ($99.3\%$) & $276 / 276$ ($100\%$) \\
    Depth estimation          & $1{,}177 / 1{,}380$ ($85.3\%$) & $251 / 276$ ($90.9\%$) \\
    Pose estimation           & $473 / 510$ ($92.7\%$) & \textsuperscript{$\ast$}$492 / 510$ ($96.5\%$) \\
    Video object seg.         & $551 / 600$ ($91.8\%$) & $111 / 120$ ($92.5\%$) \\
    Maze solving ($9\times 9$)& $600 / 600$ ($100\%$)  & $600 / 600$ ($100\%$) \\
    VPCT (\vpctblender{})     & $1{,}164 / 1{,}164$ ($100\%$) & $582 / 582$ ($100\%$) \\
    VBVR-Pro                  & $1{,}500 / 1{,}500$ ($100\%$) & $600 / 600$ ($100\%$) \\
    Cellular automata         & $596 / 600$ ($99.3\%$) & $600 / 600$ ($100\%$) \\
    \bottomrule
  \end{tabular}
\end{table}

\paragraph{Within-query clustered confidence intervals} All error bars show $95\%$ within-query clustered confidence intervals computed via a non-parametric cluster bootstrap (\num{9999} resamples) with Cousineau--Morey centering. Because the same $N$ queries are evaluated across all $C$ plotted shot counts $k$, each bootstrap replicate $b$ resamples the $N$ query clusters with replacement jointly across all $C$ shot counts---retaining all attempts associated with each sampled query---and recomputes the pooled summary statistic $\hat{y}_k^{*(b)}$ (IQM or mean). To remove shared between-query baseline variance so that error bars reflect uncertainty in comparisons across shot counts, we center each replicate's deviation $d_k^{(b)} = \hat{y}_k^{*(b)} - \hat{y}_k$ around its mean across shot counts $\bar{d}^{(b)} = \frac{1}{C}\sum_j d_j^{(b)}$ and apply the Morey small-$C$ bias correction, $\tilde{y}_k^{*(b)} = \hat{y}_k + \sqrt{C / (C - 1)}\,(d_k^{(b)} - \bar{d}^{(b)})$, taking the $2.5$th and $97.5$th percentiles of $\{\tilde{y}_k^{*(b)}\}$ (clamped to $[0, 1]$ for binary error rates).

\paragraph{Power-law scaling fits and hypothesis tests}
To quantify how error scales with the number of in-context examples $k$, we plot positive shot counts $k \geq 1$ on a $\log_2(k)$ horizontal axis and error on a $\log_2(\text{error})$ vertical axis. For conditions evaluated at multiple positive shot counts ($k \geq 1$), we fit an ordinary least squares (OLS) linear regression to the log-transformed per-shot summary values $\hat{y}_k$ (IQM or mean):
\begin{equation}
    \log_2 \hat{y}_k = \alpha + \beta \log_2 k \quad \Longleftrightarrow \quad \hat{y}_k = 2^{\alpha} k^{\beta}.
\end{equation}
From this power-law fit (dashed lines in scaling plots), we report three quantities:
\begin{enumerate}[leftmargin=*, topsep=2pt, itemsep=1pt]
    \item \textbf{Relative error change per $2\times$ shots:} $(2^{\beta} - 1) \times 100\%$, the percentage change in error each time the number of in-context examples is doubled (e.g., $-5.9\%$ per $2\times$ shots in \cref{fig:motivating-examples-b}).
    \item \textbf{Goodness of fit ($R^2$):} The coefficient of determination of the linear fit in $(\log_2 k, \log_2 \hat{y}_k)$ space across evaluated shot counts $k \geq 1$.
    \item \textbf{Cluster-bootstrap slope $p$-value:} A two-sided test of the null hypothesis $H_0: \beta = 0$ (a flat scaling curve, i.e., no improvement with additional in-context examples). To account for query-level correlation across $k$, we estimate the standard error $\operatorname{SE}(\hat{\beta})$ using the same query-level cluster bootstrap ($9{,}999$ resamples): in each bootstrap replicate $b$, we resample the $N$ queries with replacement, recompute the summary statistic $\hat{y}_k^{*(b)}$ at every $k \geq 1$, and refit the OLS slope $\hat{\beta}^{*(b)}$ of $\log_2 \hat{y}_k^{*(b)}$ on $\log_2 k$. We then compute the two-sided $z$-statistic $z = |\hat{\beta}| / \operatorname{SE}(\hat{\beta})$ and corresponding $p$-value $p = 2\big(1 - \Phi(|z|)\big)$.
\end{enumerate}

\section{Prompt formats}
\label{app:prompts}
We refer to prompt formats by their position in \cref{tab:design_space} as \emph{\{across-example packing\}}--\emph{\{within-example packing\}}--\emph{\{interaction mode\}}, e.g., \emph{separate--separate--editing} for input--output example pairs passed as separate files and a query that is edited in-place, or \emph{separate--output-only--editing} for output-only examples as separate files. \Cref{tab:task_prompt_formats} lists the format used for each task in \cref{sec:results}; the full prompts are given in the respective subsections of \cref{app:task_details}. Since the primary goal of \cref{sec:results} is to test whether ICL works across frontier video models rather than to compare \omnione{} and \seedancetwofive{} under an identical prompt format, we use a model-specific format on VBVR-Pro (the only task where the two differ), as inpainting worked best for \omnione{} and editing worked best for \seedancetwofive{}.

\begin{table}[h]
  \centering\small
  \caption{\textbf{Prompt formats used per task in \cref{sec:results}}, named by \{across-example packing\}--\{within-example packing\}--\{interaction mode\} (cf.\ \cref{tab:design_space}).}
  \label{tab:task_prompt_formats}
    \begin{tabular}{ll}
    \toprule
    \textbf{Task} & \textbf{Prompt format} \\
    \midrule
    Surface normal estimation & \textit{separate--output-only--editing} \\
    Depth estimation & \textit{separate--output-only--editing} \\
    Pose estimation & \textit{separate--spatial--inpainting} \\
    Video object segmentation & \textit{separate--output-only--editing} \\
    Maze solving & \textit{spatial--output-only--editing} \\
    VPCT & \textit{spatial--output-only--editing} \\
    VBVR-Pro (\textit{\omnione{}}) & \textit{separate--output-only--inpainting}  \\
    VBVR-Pro (\textit{\seedancetwofive{}}) & \textit{separate--output-only--editing} \\
    Cellular automata & \textit{separate--output-only--editing} \\
    \bottomrule
  \end{tabular}
\end{table}

\paragraph{Design-space prompt templates} Across the 32 feasible formats in \cref{tab:design_space}, prompts are generated programmatically from a unified three-paragraph template (\cref{sec:design-space}): (i)~a \emph{context} paragraph describing the media tags and spatiotemporal layout of the query and $k$ examples, (ii)~an \emph{action} paragraph naming the file, spatial cell, or time slice to edit, inpaint, extend, or generate from scratch, and (iii)~a fixed \emph{task-description} paragraph specifying \chromaticscrambleone{} identically across all shot counts. Representative examples spanning all four interaction modes are shown below (full per-task prompts for all other settings appear in \cref{app:task_details}).

\begin{promptbox}{Representative design-space prompts for \chromaticscrambleone{} (\cref{tab:design_space})}
\textnormal{\textbf{Editing (\textit{separate--output-only--editing}, V2V, $k = 1$):}}\\
Here is an example of a solved task: @video1. @video0 shows the query input.\\[0.5ex]
Edit @video0, replacing it with the solution for the query input.\\[0.5ex]
$\langle\text{task\_description}\rangle$

\medskip
\textnormal{\textbf{Inpainting (\textit{separate--separate--inpainting}, I2I, $k = 1$):}}\\
Here is an example with its solution: input @image0 -> solution @image1. @video0 shows the query input for the first 1s and a solid neon yellow box after that.\\[0.5ex]
Keep seconds 0-1 of @video0 unchanged. Edit seconds 1-2 (neon yellow area) of @video0, replacing it with the solution for the query input shown in seconds 0-1.\\[0.5ex]
$\langle\text{task\_description}\rangle$

\medskip
\textnormal{\textbf{Extension (\textit{temporal--output-only--extension}, V2V, $k = 2$):}}\\
@video0 consists of 3 segments, one after the other. The first 2 segments show example solutions, and the final segment (after the 2s mark) shows the query input.\\[0.5ex]
Extend @video0 with a new segment that shows the solution for the query input.\\[0.5ex]
$\langle\text{task\_description}\rangle$

\medskip
\textnormal{\textbf{Generation (\textit{separate--separate--generation}, V2V, $k = 1$):}}\\
Here is an example with its solution: input @video1 -> solution @video2. @video0 shows the query input.\\[0.5ex]
Generate a new video that shows the solution for the query input in @video0.\\[0.5ex]
$\langle\text{task\_description}\rangle$

\medskip
\textnormal{\textbf{Fixed $\langle\text{task\_description}\rangle$ paragraph (\chromaticscrambleone{}):}}
\begin{itemize}[leftmargin=1.5em, itemsep=0.25ex, topsep=0.5ex]
    \item \textnormal{I2I:} To produce the solution from an input: (1.) Red channel: invert the original blue channel (cross-channel inversion); (2.) green channel: identical; (3.) blue channel: 180-degree color phase shift on the original red channel. Keep everything else the same. Do not include any audio in the output.
    \item \textnormal{V2V:} To produce the solution from an input video: (1.) Red channel: invert the original blue channel (cross-channel inversion); (2.) green channel: identical; (3.) blue channel: 180-degree color phase shift on the original red channel. Apart from this, keep the content, motion and timing of the video exactly the same. Do not include any audio in the output.
\end{itemize}
\end{promptbox}

\section{Video likelihood scoring results for Wan 2.1}
\label{app:video_scoring}
\paragraph{Result} We evaluated whether Wan~2.1 (Wan2.1-T2V-1.3B, open-weight, released under an Apache 2.0 license~\citep{wan2025}) can score video continuations above chance on videos from the Physics-IQ dataset~\citep{motamed2025do}. The dataset consists of 198 videos, split into a conditioning part and a ground truth continuation. An interesting property of the dataset is that for each scenario, the exact same action (e.g., a glass falling down, a chain of dominoes) is filmed from three different perspectives. This means we can take a conditioning video (filmed, e.g., from `perspective-left') and have the video model score which of the three possible continuations (`perspective-left' vs.\ `perspective-center' vs. `perspective-right') is the correct one. The task is challenging since the scene, the action, and even the timing of the action are matched across the three perspectives. The experiment was performed at a resolution of $64 \times 64$ pixels with the last eight frames of the conditioning video and first eight frames of each of the three possible continuation videos. On this task, Wan~2.1 achieved an overall accuracy of 99.49\% ($197/198$, $p < .001$, exact two-sided binomial test against chance $p_0 = 1/3$). This confirms that a video model like Wan 2.1 can successfully decrease its error based on visual context (the conditioning signal), making video scoring one way in which a video model exhibits in-context learning.

\paragraph{Method} We evaluate the conditional likelihood of a candidate video continuation $Y$ given a conditioning prefix video $X$ using a flow-matching video diffusion model. Prefix and candidate frames are spatially resized to a square $r \times r$ frame, with $r$ denoting the resolution. The concatenated video $[X \mathbin{\Vert} Y]$ is mapped into continuous latent space via a 3D Causal Variational Autoencoder (VAE): 

\begin{equation}
    Z = \operatorname{VAE}([X \mathbin{\Vert} Y]) = [Z_{\text{prefix}} \mathbin{\Vert} Z_{\text{cont}}].
\end{equation}

Because the VAE convolutions are strictly causal in time (using left-only temporal padding), prefix latents $Z_{\text{prefix}}$ depend solely on the conditioning frames $X$. Following the flow-matching formulation, we hold the prefix latents clean while diffusing the continuation latents to flow timestep $t \in (0, 1)$ with Gaussian noise $\epsilon \sim \mathcal{N}(0, \mathbf{I})$:

\begin{equation}
    \tilde{Z} = \bigl[Z_{\text{prefix}} \mathbin{\Vert} t Z_{\text{cont}} + (1 - t)\epsilon\bigr], \enspace u^* = \epsilon - Z_{\text{cont}},
\end{equation}

where $u^*$ denotes the ground-truth velocity target pointing from clean data to noise. The model receives a single scalar timestep for the entire latent sequence, i.e., the same noise level is used for all tokens, including the prefix tokens. No text prompt is used, text conditioning is set to an all-zero embedding, and no classifier-free guidance is applied. The conditional surrogate log-likelihood score $S(Y \mid X)$ is defined as the negative velocity prediction mean squared error, evaluated over the continuation latent domain $\Omega_{\text{cont}}$ and averaged over a set of flow timesteps $\mathcal{T} = \{0.3, 0.4, \ldots, 0.8\}$:

\begin{equation}
    S(Y \mid X) = - \frac{1}{|\mathcal{T}|\,|\Omega_{\text{cont}}|} \sum_{t \in \mathcal{T}} \bigl\| \hat{u}_\theta(\tilde{Z}_t, t)_{\text{cont}} - u^*_t \bigr\|_2^2,
\end{equation}

where $\tilde{Z}_t$ and $u^*_t$ are constructed as above with a noise sample $\epsilon_t$ drawn independently for each timestep. Averaging over a range of noise levels avoids making the score depend on a single arbitrary point of the flow path, while skipping the near-clean and near-pure-noise ends, where the velocity target presumably contains little candidate-specific signal. A higher score corresponds to lower reconstruction error, which indicates higher compatibility with the prefix under the model's spatio-temporal prior. When comparing $K$ competing candidate continuations $\{Y_1, \dots, Y_K\}$, standard independent noise sampling introduces large Monte Carlo variance in score differences. To eliminate this sampling noise, we employ Common Random Numbers: for each timestep $t \in \mathcal{T}$, a single noise sample $\epsilon_t \sim \mathcal{N}(0, \mathbf{I})$ is used across all $K$ candidates. Candidates were scored by the model's flow-matching loss on the continuation frames, averaged over six noise levels ($t \in \{0.3, 0.4, \ldots, 0.8\}$) with a single shared noise draw per level so that the three candidates remain directly comparable.

\section{Task details}
\label{app:task_details}

\subsection{Image and video transforms}
\label{app:image_and_video_transforms}

\paragraph{Data} For the image-to-image~(I2I) version, we use OpenImages~\citep{krasin2017openimages,kuznetsova2020open}: 50 randomly selected nature images with \href{https://creativecommons.org/licenses/by/2.0/}{CC-BY-2.0 license} not containing any people from the 1600HQ subset (high resolution), downsampled using Lanczos and center cropped to $1280 \times 720$.
For the video-to-video~(V2V) version, we use DAVIS~2017~\citep[][CC~BY-NC~4.0]{pont20172017}: the first 50 videos across the training and validation splits after excluding 26 manually screened sequences containing minors (19 from \texttt{train} and 7 from \texttt{val}), resized from full resolution to $1280 \times 720$ using Lanczos and clipped to the first 24 frames ($1\,\text{s}$ at $24\,\text{fps}$), with the image transform applied frame-by-frame to obtain the target videos. For both versions, the $k$ in-context examples for a query are sampled without replacement from the remaining 49 queries, so an example never shows the query scene.

\begin{figure*}[ht]
    \centering
    \begin{subfigure}[b]{0.24\textwidth}
        \caption{\opensans Original}
        \includegraphics[width=\textwidth]{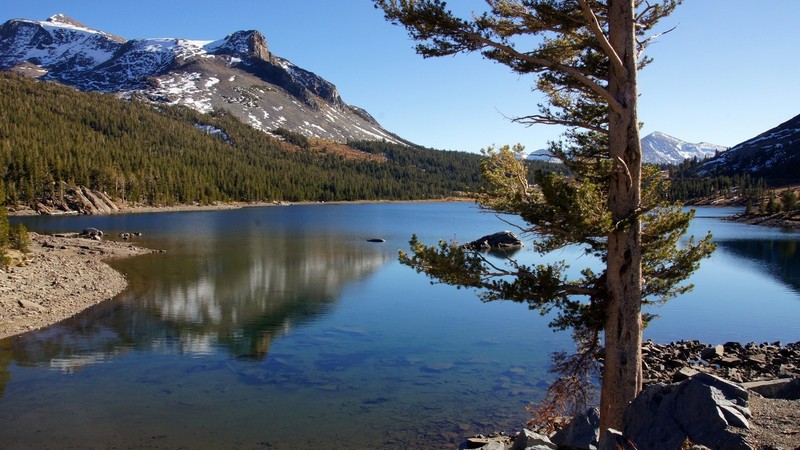}
    \end{subfigure}\hfill
    \begin{subfigure}[b]{0.24\textwidth}
        \caption{\opensans Chromatic scramble v1}
        \includegraphics[width=\textwidth]{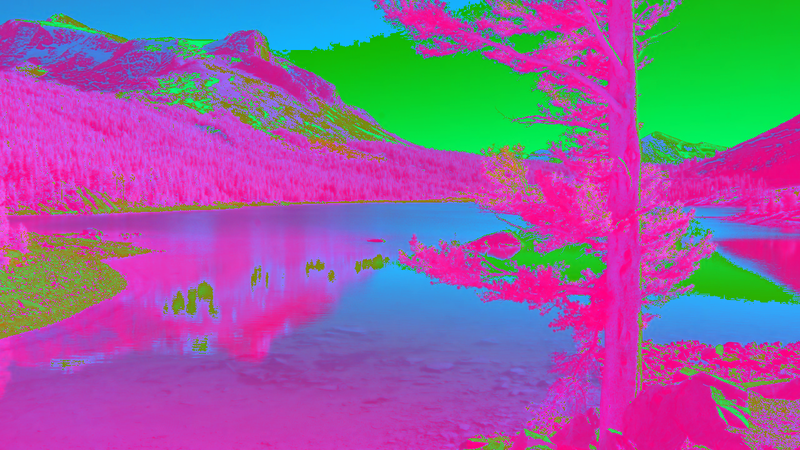}
    \end{subfigure}\hfill
    \begin{subfigure}[b]{0.24\textwidth}
        \caption{\opensans Chromatic scramble v2}
        \includegraphics[width=\textwidth]{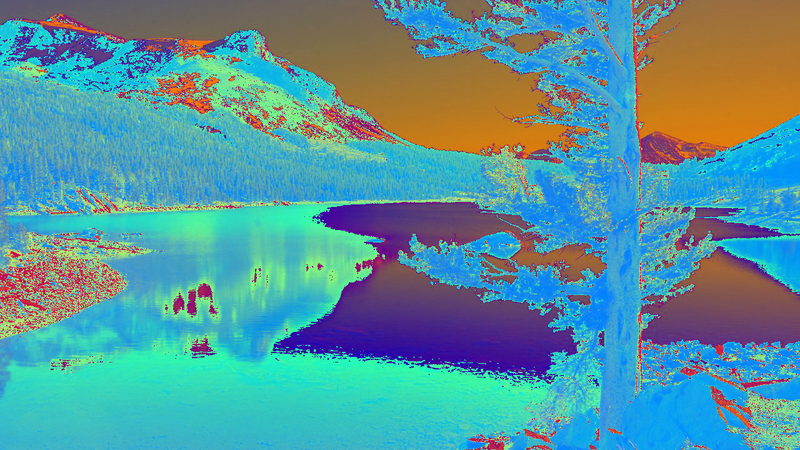}
    \end{subfigure}\hfill
    \begin{subfigure}[b]{0.24\textwidth}
        \caption{\opensans Chromatic scramble v3}
        \includegraphics[width=\textwidth]{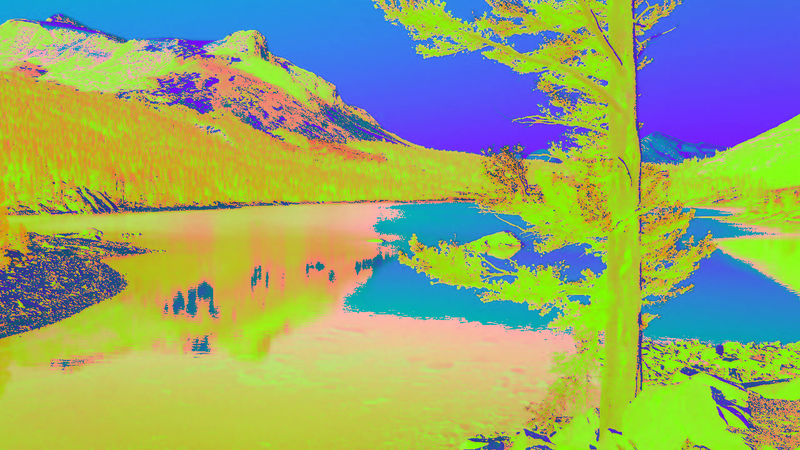}
    \end{subfigure}

    \begin{subfigure}[b]{0.24\textwidth}
        \mbox{}
    \end{subfigure}\hfill
    \begin{subfigure}[b]{0.24\textwidth}
        \caption{\opensans Brightness increase}
        \includegraphics[width=\textwidth]{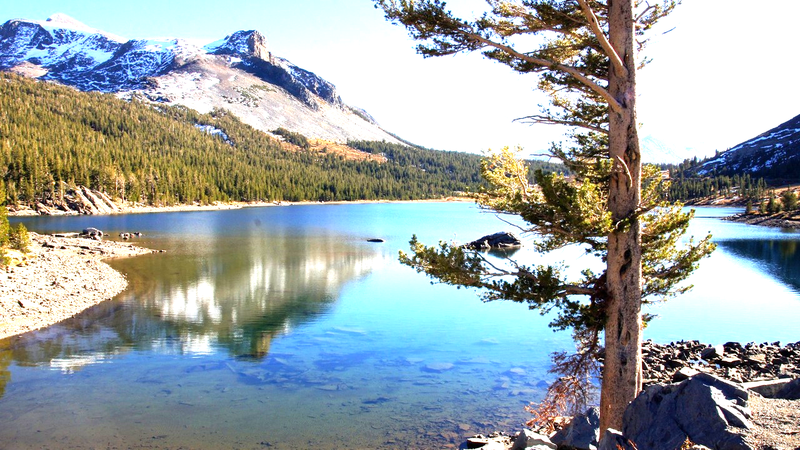}
    \end{subfigure}\hfill
    \begin{subfigure}[b]{0.24\textwidth}
        \caption{\opensans Solarize}
        \includegraphics[width=\textwidth]{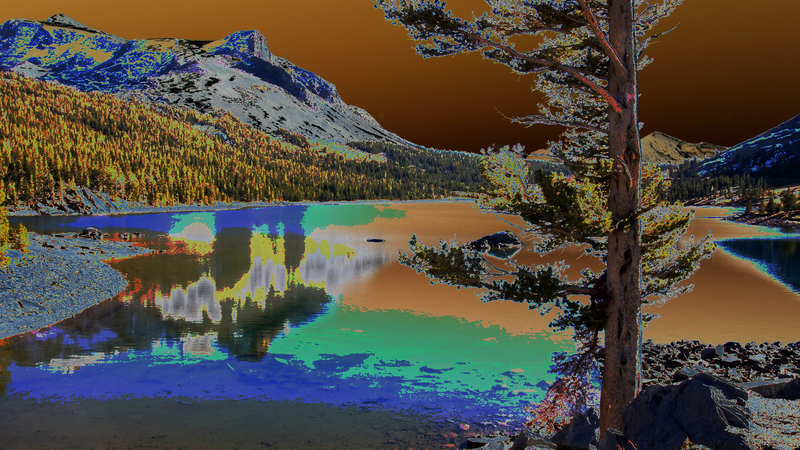}
    \end{subfigure}\hfill
    \begin{subfigure}[b]{0.24\textwidth}
        \caption{\opensans Greyscale (control)}
        \includegraphics[width=\textwidth]{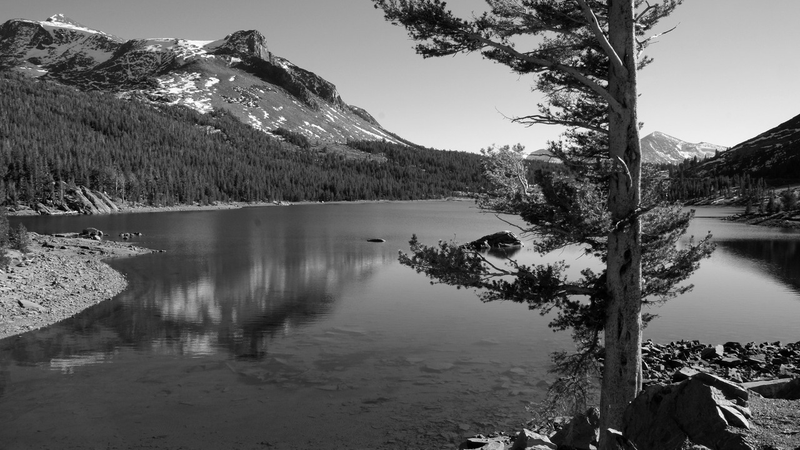}
    \end{subfigure}
    
    \caption{\textbf{Overview of image transformations.}}
    \label{fig:image_transforms_examples}
\end{figure*}

\paragraph{Chromatic scramble transforms} For an input pixel with 8-bit RGB channels $r, g, b \in \{0, \dots, 255\}$, the three chromatic scramble transforms (\cref{fig:image_transforms_examples}, top row) are defined as
\begin{align}
    \text{\makecell[l]{Chromatic\\[-2pt]scramble~v1:}} &\quad \begin{aligned}
        r_\text{new} &= 255 - b, \\[-2pt]
        g_\text{new} &= g, \\[-2pt]
        b_\text{new} &= \mathrm{mod}((r + 128), 256),
    \end{aligned} \label{eq:ood_transform_v1} \\[3pt]
    \text{\makecell[l]{Chromatic\\[-2pt]scramble~v2:}} &\quad \begin{aligned}
        r_\text{new} &= r, \\[-2pt]
        g_\text{new} &= \mathrm{mod}((b + 128), 256), \\[-2pt]
        b_\text{new} &= 255 - g,
    \end{aligned} \label{eq:ood_transform_v2} \\[3pt]
    \text{\makecell[l]{Chromatic\\[-2pt]scramble~v3:}} &\quad \begin{aligned}
        r_\text{new} &= \mathrm{mod}((g + 128), 256), \\[-2pt]
        g_\text{new} &= 255 - r, \\[-2pt]
        b_\text{new} &= b.
    \end{aligned} \label{eq:ood_transform_v3}
\end{align}

\paragraph{Other transforms} The remaining three transformations (\cref{fig:image_transforms_examples}, bottom row) are implemented using Pillow (\texttt{PIL}) on RGB images as
\begin{align}
    &\begin{aligned}
        &\text{Brightness increase:} \\[-2pt]
        &\;\; \texttt{\footnotesize ImageEnhance.Brightness(image)}\\[-3pt]
        &\qquad \texttt{\footnotesize .enhance(2.0)},
    \end{aligned} \label{eq:transform_brightness_increase} \\[2pt]
    &\begin{aligned}
        &\text{Solarize:} \\[-2pt]
        &\;\; \texttt{\footnotesize ImageOps.solarize(image,}\\[-3pt]
        &\qquad \texttt{\footnotesize threshold=80)},
    \end{aligned} \label{eq:transform_solarize} \\[2pt]
    &\begin{aligned}
        &\text{Greyscale (control):} \\[-2pt]
        &\;\; \texttt{\footnotesize image.convert(\textquotesingle L\textquotesingle).convert(\textquotesingle RGB\textquotesingle)}.
    \end{aligned} \label{eq:transform_greyscale}
\end{align}

Samples are visualized in \cref{fig:image_transforms_examples}.

\begin{figure*}[htbp]
    \centering
    \begin{subfigure}[b]{\twofigwidth\textwidth}
        \centering
        \caption{\opensans Chromatic scramble v1}
        \includegraphics[width=\textwidth]{figures/fig1/297312812_k0to10_normalized_stripplot.pdf} %
    \end{subfigure}
    \hfill
    \begin{subfigure}[b]{\twofigwidth\textwidth}
        \centering
        \caption{\opensans Chromatic scramble v2}
        \includegraphics[width=\textwidth]{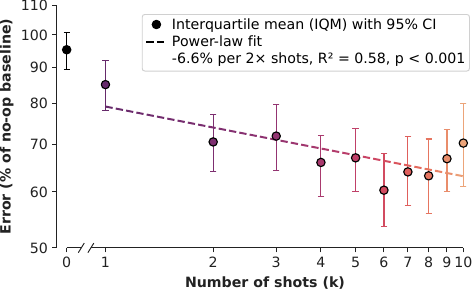}
    \end{subfigure}
    \hfill
    \begin{subfigure}[b]{\twofigwidth\textwidth}
        \centering
        \caption{\opensans Chromatic scramble v3}
        \includegraphics[width=\textwidth]{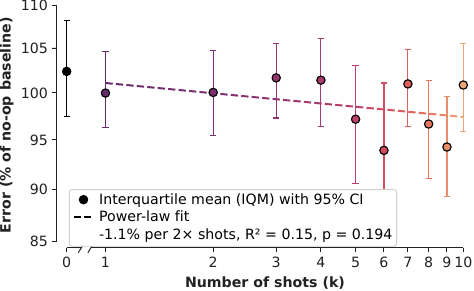}
    \end{subfigure}
    \hfill
    \begin{subfigure}[b]{\twofigwidth\textwidth}
        \centering
        \caption{\opensans Brightness increase}
        \includegraphics[width=\textwidth]{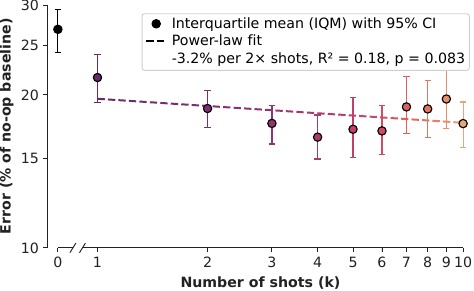}
    \end{subfigure}
    \hfill
    \begin{subfigure}[b]{\twofigwidth\textwidth}
        \centering
        \caption{\opensans Solarize}
        \includegraphics[width=\textwidth]{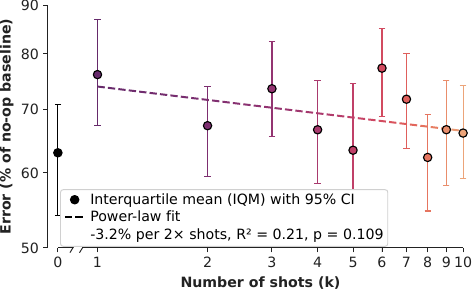}
    \end{subfigure}
    \hfill
    \begin{subfigure}[b]{\twofigwidth\textwidth}
        \centering
        \caption{\opensans Greyscale (control)}
        \includegraphics[width=\textwidth]{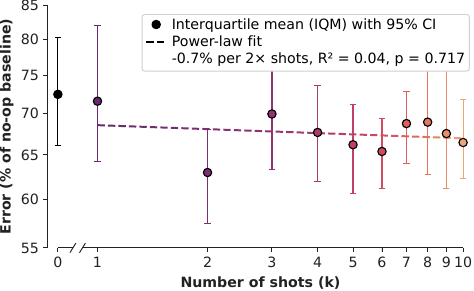}
    \end{subfigure}
    \caption{\textbf{Image transforms on \omnione{}.}}
    \label{fig:image_transforms_results_omni_1p2}
\end{figure*}

\paragraph{Evaluation}
RGB pixel intensities are normalized to $[0, 1]$ by dividing 8-bit channel values by $255$, and prediction accuracy is measured by MSE. 
For I2I, we extract the final frame of the generated video (or of the cropped spatiotemporal query region for packed formats in \cref{tab:design_space}), resize it with Lanczos resampling to match the target (or common evaluation) resolution, and compute the MSE against the transformed target image. For V2V, we temporally align and clip the generated video (or extracted query segment) to the target's $24$-frame ($1\,\text{s}$) duration, resize each frame with Lanczos resampling, and compute per-frame MSE for every frame $t \in \{1, \dots, 24\}$. Following \cref{app:eval-statistics}, errors are normalized against the \emph{identity} baseline---scoring the unmodified query input against the target---as $100 \times \operatorname{MSE}(\hat{I}, I) / \operatorname{MSE}(I_{\text{orig}}, I)$ for I2I and frame-by-frame as $100 \times \frac{1}{T}\sum_{t=1}^{T} \operatorname{MSE}(\hat{I}_t, I_t) / \operatorname{MSE}(I_{\text{orig},t}, I_t)$ for V2V. In the design-space sweep (\cref{tab:design_space}), formats are evaluated at a common downscaled resolution ($316 \times 178$ for V2V, matching the smallest spatially tiled query cell; \cref{app:eval-statistics}). Results in \cref{fig:motivating-examples-b,tab:design_space,fig:image_transforms_results_omni_1p2} are aggregated via the IQM across queries and attempts. The Best-of-N experiments in \cref{fig:double_scaling,fig:double_scaling_big_panel,fig:double_scaling_heatmap} use the empirical mean (\cref{app:best_of_n_methods}).

\paragraph{Experiment details}
In the terminology of \cref{tab:design_space}, the main scaling and multi-transform I2I experiments (\cref{fig:motivating-examples-b,fig:image_transforms_results_omni_1p2}) use the \emph{separate--separate--editing} setting for \omnione{}. Each query image is converted to a $4\,\text{s}$ static video ($96$ frames at $24\,\text{fps}$), and each of the $k \in \{0, \dots, 10\}$ input--output examples is passed as two separate reference images (\texttt{@image0}, \texttt{@image1}, \dots). We evaluate all 50 queries across all 11 shot counts ($k \in \{0, \dots, 10\}$): for \chromaticscrambleone{} (\cref{fig:motivating-examples-b} and \cref{fig:image_transforms_results_omni_1p2}, top-left; also used in \cref{app:best_of_n_methods}), we run 10 attempts per query and shot count; for the five additional transforms in \cref{fig:image_transforms_results_omni_1p2}, we run 2 attempts per query and shot count ($1{,}100$ requested videos per transform. The additional 10-attempt Best-of-N runs on \seedancetwofive{} (\chromaticscrambleone{}, $k \in \{0, 1, 5, 10\}$, $2{,}000/2{,}000$ completed) and \omnione{} (\chromaticscrambletwo{}, $k \in \{0, \dots, 10\}$, $5{,}480/5{,}500$ completed) use the same prompts and are detailed in \cref{app:best_of_n_methods}.

For the prompt design-space sweeps in \cref{tab:design_space}, both I2I (\cref{tab:design_space_image}) and V2V (\cref{tab:design_space_video}) evaluate \chromaticscrambleone{} on \omnione{} across all 32 feasible prompt formats using all 50 queries and 2 attempts per query and shot count ($100$ requested videos per shot count; \cref{tab:design_space} reports $k \in \{0, 1, 4\}$). In \cref{tab:design_space_image}, query images (and video-formatted examples) are converted to $1\,\text{s}$ static videos; in \cref{tab:design_space_video}, queries and examples are the $1\,\text{s}$ ($24$-frame) DAVIS~2017 clips. Inpainting formats fill the target query slot with solid neon yellow (preceded by a $1\,\text{s}$ query-input primer when editing in-place without a separate query input cell). All 32 formats use the unified three-paragraph prompt templates described in \cref{sec:design-space,app:prompts}.

\begin{promptbox}{Prompts for image transforms ($k \in \{0, \dots, 10\}$)}
\textnormal{\textbf{0-shot ($k = 0$):}}\\
Seconds 0-2: nothing changes. Seconds 3-4: Edit the video as follows: $\langle\text{transform\_rule}\rangle$ Keep everything else the same. Do not include any audio in the output.

\medskip
\textnormal{\textbf{1-shot ($k = 1$):}}\\
Here is an example with its solution: input @image0 -> solution @image1.\\[0.5ex]
Seconds 0-2: nothing changes. Seconds 3-4: Edit the video as follows: $\langle\text{transform\_rule}\rangle$ Keep everything else the same. Do not include any audio in the output.

\medskip
\textnormal{\textbf{Few-shot ($k \in \{2, \dots, 10\}$):}}\\
Here are examples with their solutions: input @image0 -> solution @image1, \dots, input @image$\langle 2k{-}2\rangle$ -> solution @image$\langle 2k{-}1\rangle$.\\[0.5ex]
Seconds 0-2: nothing changes. Seconds 3-4: Edit the video as follows: $\langle\text{transform\_rule}\rangle$ Keep everything else the same. Do not include any audio in the output.

\medskip
\textnormal{\textbf{Per-transform $\langle\text{transform\_rule}\rangle$:}}
\begin{itemize}[leftmargin=1.5em, itemsep=0.25ex, topsep=0.5ex]
    \item \chromaticscrambleone{}: (1.) Red channel: invert the original blue channel (cross-channel inversion); (2.) green channel: identical; (3.) blue channel: 180-degree color phase shift on the original red channel.
    \item \chromaticscrambletwo{}: (1.) Red channel: identical; (2.) green channel: 180-degree color phase shift on the original blue channel; (3.) blue channel: invert the original green channel (cross-channel inversion).
    \item \chromaticscramblethree{}: (1.) Red channel: 180-degree color phase shift on the original green channel; (2.) green channel: invert the original red channel (cross-channel inversion); (3.) blue channel: identical.
    \item \textnormal{Brightness increase:} make the image much brighter by doubling the brightness.
    \item \textnormal{Solarize:} solarize the image by inverting all pixel values that are above brightness 80 (out of 255).
    \item \textnormal{Greyscale (control):} convert the image to greyscale.
\end{itemize}
\end{promptbox}

\paragraph{Additional results}
In addition to \chromaticscrambleone{} described in the main text, we here test a total of six image transformations. \chromaticscrambletwo{} and \chromaticscramblethree{} are of similar nature to \chromaticscrambleone{}, except for a permutation of the channel assignment (\cref{eq:ood_transform_v1,eq:ood_transform_v2,eq:ood_transform_v3}). Furthermore, three transformations that are likely in-distribution are tested: \texttt{Greyscale} as a control condition (almost certainly part of the training data), and the \texttt{Brightness increase} and \texttt{Solarize} transforms (also prominent in video data across the web). The results are shown in \cref{fig:image_transforms_results_omni_1p2}. For all three chromatic scramble transforms, the zero-shot baseline performs roughly at chance (near 100\% error), while for likely in-distribution distortions the zero-shot error is much lower already. As expected, the control condition (\texttt{Greyscale}) shows no benefit of ICL; for other transforms there is a general downwards trend (more examples are better than just one example); the effect is most pronounced for two of the chromatic scramble transforms. Due to the high cost of \seedancetwofive{} inference (approx.\ six times the cost of \omnione{}), these additional tasks are only evaluated on \omnione{}.

\subsection{Surface normal estimation}
\label{app:surface_normal}
\paragraph{Data}
We use the 23 training sequences of MPI~Sintel~\citep{butler2012naturalistic}, which is rendered from the open-source animated short film \emph{Sintel} by the Blender Foundation (CC~BY~3.0). Each sequence is converted into a query video at $24\,\text{fps}$ and its native $1024 \times 436$ resolution ($0.83$--$2.08\,\text{s}$, i.e., $20$--$50$ frames). The targets are videos of per-pixel ground-truth surface normals, where each unit normal $\mathbf{n} \in [-1, 1]^3$ is encoded as the RGB color $255 \cdot (\mathbf{n} + 1) / 2$.

For the image examples of the main experiments, we sample one random frame per sequence from the ground-truth normal video. For the video examples of the additional experiment below, we cut a random $24$-frame ($1\,\text{s}$) window from each normal video, skipping the two sequences shorter than $24$ frames. The examples for a query are always drawn from the other sequences, so an example never shows the query scene.

\paragraph{Evaluation}
Each generated frame is resized to the ground-truth resolution, and both the predicted and the ground-truth RGB values are mapped back to $[-1, 1]^3$ via $\mathbf{n} = c / 127.5 - 1$ and normalized to unit length. Pixels where either vector has (near-)zero length are excluded. For every valid pixel $p$ of frame $t$, we compute the angle between the predicted normal $\hat{\mathbf{n}}_{t,p}$ and the ground-truth normal $\mathbf{n}_{t,p}$, and report the mean angular error
\[
e = \frac{1}{T} \sum_{t=1}^{T} \frac{1}{|\Omega_t|} \sum_{p \in \Omega_t} \arccos\!\big(\hat{\mathbf{n}}_{t,p}^\top \mathbf{n}_{t,p}\big)
\]
in degrees (lower is better), where $\Omega_t$ is the set of valid pixels of frame $t$ and $T$ is the number of compared frames. For \omnione{}, generated and ground-truth frames are paired by index. \seedancetwofive{} returns its edit of the input re-timed (e.g., $89$ frames for a $96$-frame input), so we pair every ground-truth frame with the generated frame that shows the same instant of the model input. The \emph{identity} baseline scores the unmodified input video through the same pipeline. As an additional metric, we report the \emph{rotation-aligned} mean angular error, which first rotates all predicted normals of a video by the single rotation $R \in SO(3)$ that best aligns them with the ground truth (solved in closed form via the Kabsch algorithm over all valid pixels of the video) and thus discounts any global mismatch in the coordinate convention.

\paragraph{Experiment details}
In the terminology of \cref{tab:design_space}, the main surface normal experiments in \cref{fig:surface_normal} use the ``separate files, output-only, editing'' setting. The model receives the Sintel query video and $k \in \{0, \dots, 5\}$ ground-truth normal frames as separate reference images, and is prompted to edit the query video in-place (see prompts below). For \seedancetwofive{}, the query video is freeze-padded to $4\,\text{s}$ before inference. We evaluate all 23 sequences with 2 attempts per query and shot count ($276$ requested videos per model; $274$ were generated for \omnione{} and all $276$ for \seedancetwofive{}).

\begin{promptbox}{Prompts for surface normal estimation ($k \in \{0, \dots, 5\}$)}
\textnormal{\textbf{0-shot ($k = 0$):}}\\
Edit @video0. The output should be a video of surface normals. The output video should have the same camera motion and scene content as the input video.

\medskip
\textnormal{\textbf{1-shot ($k = 1$):}}\\
Create the video in the style of the following example: @image0.\\[0.5ex]
Edit @video0. The output should be a video of surface normals. The output video should have the same camera motion and scene content as the input video.

\medskip
\textnormal{\textbf{Few-shot ($k \geq 2$):}}\\
Create the video in the style of the following examples: @image0, \dots, @image$\langle k-1\rangle$.\\[0.5ex]
Edit @video0. The output should be a video of surface normals. The output video should have the same camera motion and scene content as the input video.
\end{promptbox}

\paragraph{Additional results}
The rotation-aligned angular error confirms that a large part of the zero-shot error stems from a mismatched color convention: for \omnione{}, the aligned error is $55.3^\circ$ zero-shot (vs.\ $72.4^\circ$ unaligned) and $47.0^\circ$ at $k=1$ ($p < 0.001$), and stays flat for more examples ($46.1^\circ$ at $k=5$; $\tau = +0.01$, $p = 0.835$). For \seedancetwofive{}, the aligned error decreases from $60.9^\circ$ zero-shot (slightly above the aligned identity baseline of $57.2^\circ$) to $49.3^\circ$ at $k=5$ ($\tau = -0.21$, $p = 0.009$), so its examples improve the predicted geometry beyond the convention. For context, dedicated surface normal predictors such as DSINE~\citep{bae2024dsine} and fine-tuned diffusion priors~\citep{jiang2025geo4d,wang2026rgb,di2026are} achieve $\approx 13^\circ$--$25^\circ$ mean angular error on zero-shot surface normal benchmarks; our evaluation instead isolates whether unmodified video generators can infer the task and coordinate convention purely from in-context examples.

\subsection{Depth estimation}
\label{app:depth_estimation}
\paragraph{Data}
We use the same 23 MPI~Sintel~\citep{butler2012naturalistic} training sequences as for surface normal estimation (\cref{app:surface_normal}), with the RGB frames as query videos at $24\,\text{fps}$ and their native $1024 \times 436$ resolution ($0.83$--$2.08\,\text{s}$, i.e., $20$--$50$ frames), and the per-frame ground-truth depth maps (\texttt{.dpt} files, in meters) as targets.

For the examples, we sample one random frame per sequence (fixed seed) and render its ground-truth depth map with the Viridis colormap, linearly mapping depths in $[0, 70]\,\text{m}$ to the colormap (dark purple for close, bright yellow for far objects) and clipping farther depths to bright yellow. The examples for a query are always drawn from the other 22 sequences, so an example never shows the query scene.

\paragraph{Evaluation}
For each frame of the generated video, we invert the colormap by assigning every pixel the depth value of its nearest (in RGB space) Viridis color, resize the resulting relative depth map to the ground-truth resolution, and pair frames by index with the ground-truth depth maps. Since the encoding only determines depth up to an unknown scale and shift, each predicted frame $\hat{d}$ is aligned to the ground truth $d$ with a robust median-based scale $s$ and shift $t$ computed over the valid pixels $\Omega$ (finite ground-truth depth larger than $1\,\text{mm}$ and below the maximum depth threshold): $s = \operatorname{median}_{\Omega}\big[(d - \operatorname{median}_{\Omega} d) / (\hat{d} - \operatorname{median}_{\Omega} \hat{d})\big]$ and $t = \operatorname{median}_{\Omega}\big[d - s\,\hat{d}\big]$. Negative aligned depths are clipped to zero. We then report the absolute relative error
\[
\text{AbsRel} = \frac{1}{|\Omega|} \sum_{p \in \Omega} \frac{|s\,\hat{d}_p + t - d_p|}{d_p},
\]
averaged over all frames of a video (lower is better). The \emph{identity} baseline scores the unmodified query video through the same pipeline (decoding its RGB frames as if they were Viridis depth maps). For reference, dedicated monocular and video depth estimators (e.g., Depth Anything~\citep{yang2024depthv2,lin2025depth}, Marigold~\citep{ke2024repurposing}, and fine-tuned video depth diffusion models~\citep{jiang2025geo4d,zhang2026dvd,he2026icdepth}) achieve $\text{AbsRel} \approx 0.09$--$0.25$ on MPI~Sintel by predicting metric or affine-invariant depth maps directly, whereas our setup tests pixel-space video editing into a Viridis colormap.

\paragraph{Experiment details}
In the terminology of \cref{tab:design_space}, the main depth experiments in \cref{fig:depth_estimation} use the ``separate files, output-only, editing'' setting. The model receives the Sintel query video and $k$ Viridis depth-map examples as separate reference images, and is prompted to edit the query video in-place (see prompts below). For \seedancetwofive{}, the query video is freeze-padded to $4\,\text{s}$ before inference. We evaluate all 23 sequences for $k \in \{0, \dots, 5\}$, with 10 attempts per query and shot count for \omnione{} ($1{,}177$ of $1{,}380$ requested videos were generated) and 2 attempts for \seedancetwofive{} ($251$ of $276$).

\begin{promptbox}{Prompts for depth estimation ($k \in \{0, \dots, 5\}$)}
\textnormal{\textbf{0-shot ($k = 0$):}}\\
@video0 shows the query input.\\[0.5ex]
Edit @video0, replacing it with the solution for the query input.\\[0.5ex]
The output video should perform depth estimation on the input video, and other than that keep the scene content and camera motion. Depth is encoded with the viridis colormap, where dark purple represents closer objects and bright yellow represents farther objects.

\medskip
\textnormal{\textbf{1-shot ($k = 1$):}}\\
Edit the video to apply the style of the following example: @image0.\\[0.5ex]
Edit @video0. The output video should perform depth estimation on the input video, and other than that keep the scene content and camera motion. Depth is encoded with the viridis colormap, where dark purple represents closer objects and bright yellow represents farther objects.

\medskip
\textnormal{\textbf{Few-shot ($k \geq 2$):}}\\
Edit the video to apply the style of the following examples: @image0, \dots, @image$\langle k-1\rangle$.\\[0.5ex]
Edit @video0. The output video should perform depth estimation on the input video, and other than that keep the scene content and camera motion. Depth is encoded with the viridis colormap, where dark purple represents closer objects and bright yellow represents farther objects.
\end{promptbox}

\subsection{Pose estimation}
\label{app:pose_estimation}
\paragraph{Data}
We use the EMDB~1 subset of the EMDB dataset~\citep{kaufmann2023emdb}, which provides in-the-wild, handheld smartphone videos ($1440 \times 1920$, portrait) of people with accurate per-frame 3D SMPL body poses, camera intrinsics, and a mask of frames with reliable annotations. %
We split each EMDB~1 video into contiguous clips of at most $10\,\text{s}$ (dropping trailing clips shorter than $2\,\text{s}$ and clips without any reliably annotated frame), yielding 85 query clips. For each clip, we project the 3D positions of the 12 major body joints (left and right shoulders, elbows, wrists, hips, knees, and ankles) into the image using the camera intrinsics. The target video overlays a solid purple circle (\texttt{RGB(170, 0, 255)}, radius $1\%$ of the shorter image side, clamped to $4$--$10\,\text{px}$) on every visible joint of the original clip; frames without reliable annotations are left without joints. We chose purple because it is the least common saturated hue in the EMDB~1 videos, which keeps the joint detection in the evaluator free of spurious detections from scene content. As examples, we use the first reliably annotated frame of each clip, both as the plain RGB frame (example input) and with the joints overlaid (example output). The examples for a query are always drawn from other clips. For \seedancetwofive{}, we additionally create a face-blurred version of all clips and examples (with identical targets), since most generation requests with the original clips did not return a video ($61$ of $510$ in an initial run).

\paragraph{Evaluation}
After cropping the query region from the generated video and resizing it to the original clip resolution, the evaluator detects the predicted joints in every frame: it thresholds the frame in HSV space around the joint hue ($\pm 20^\circ$ hue, saturation and value $\geq 150$), extracts connected blobs, and keeps sufficiently circular blobs (circularity $\geq 0.2$) with an area between $3\,\text{px}^2$ and $0.2\%$ of the frame (at least $300\,\text{px}^2$) as joint detections, using their centroids as predicted joint locations. Generated and ground-truth frames are aligned by timestamp. In every reliably annotated frame $t$ of resolution $w \times h$, the detections are matched one-to-one to the $N_t$ visible ground-truth 2D joints $\{\mathbf{p}_{t,j}\}_{j=1}^{N_t}$ via minimum-weight bipartite (Hungarian) matching on Euclidean pixel distance. Because averaging the localization error only over detections matched within a fixed distance threshold suffers from survivorship bias whenever difficult joints are missed, our main metric is the \emph{clamped mean joint error}, which evaluates all $N_t$ ground-truth joints in every frame and assigns any joint that is not found (unmatched) the maximum possible Euclidean pixel distance in the frame, namely the frame diagonal $d_{\max} = \sqrt{w^2 + h^2}$ ($2400\,\text{px}$ for $1440 \times 1920$ EMDB~1 videos):
\begin{equation}
    e_{\text{clamped}}(t) = \frac{1}{N_t} \sum_{j=1}^{N_t} \min\bigl(\|\hat{\mathbf{p}}_{t,\pi(j)} - \mathbf{p}_{t,j}\|_2,\; d_{\max}\bigr),
    \label{eq:clamped_joint_error}
\end{equation}
where $\pi(j)$ denotes the Hungarian assignment for ground-truth joint $j$, with $\|\hat{\mathbf{p}}_{t,\pi(j)} - \mathbf{p}_{t,j}\|_2 = \infty$ whenever joint $j$ is not found (i.e., when fewer joint blobs than ground-truth joints are detected in frame $t$). Per-frame errors are averaged over all annotated frames of a clip. Since an unmodified input video contains no joint markers, the \emph{identity} baseline receives the maximum error of $d_{\max} = 2400\,\text{px}$ on every frame. As additional metrics, the evaluator reports the percentage of correct keypoints (PCK@0.1), i.e., the fraction of ground-truth joints with a matched detection within $10\%$ of the larger side of the person's bounding box (unmatched joints count as incorrect), and the matched mean joint error over joints matched within $30\,\text{px}$.

\paragraph{Experiment details}
In the terminology of \cref{tab:design_space}, the main pose estimation experiments in \cref{fig:pose_estimation} use the ``separate files, spatial, inpainting'' setting. The query is a side-by-side composite video with the original clip on the left and a solid neon-yellow box of the same size on the right, which the model is prompted to inpaint with the joint overlay (see prompts below). Each of the $k \in \{0, \dots, 5\}$ examples is passed as a separate reference image showing a side-by-side [input $|$ solution] frame pair. We generate 1 attempt per query and shot count ($510$ requested videos per model). For \omnione{}, $473$ videos were generated (covering 79 of the 85 clips); for \seedancetwofive{} on the face-blurred clips, $492$ videos were generated ($82$ per shot count, covering 82 of the 85 clips), with $5\,\text{s}$ outputs at $720\text{p}$. Ungenerated videos are excluded from the analysis.

\begin{promptbox}{Prompts for pose estimation ($k \in \{0, \dots, 5\}$)}
\textnormal{\textbf{0-shot ($k = 0$):}}\\
@video0 shows 2 parts side-by-side. The left part of @video0 shows the query input, and the right part of @video0 is currently a solid neon yellow box.\\[0.5ex]
Edit only the right part (neon yellow area) of @video0, replacing it with the solution for the query input on its left.\\[0.5ex]
Overlay solid purple circles on the 12 major body joints (left and right shoulders, elbows, wrists, hips, knees, and ankles) of the person, tracking their motion across frames with no connecting lines. The circles should not disappear. Other than the overlaid circles, nothing changes in the input video.

\medskip
\textnormal{\textbf{1-shot ($k = 1$):}}\\
Here is a side-by-side [input $|$ solution] example: @image0. In @video0, the left part shows the query input, and the right part is currently a solid neon yellow box.\\[0.5ex]
Edit only the right part (neon yellow area) of @video0, replacing it with the solution for the query input on its left.\\[0.5ex]
Overlay solid purple circles on the 12 major body joints [...] (as above).

\medskip
\textnormal{\textbf{Few-shot ($k \geq 2$):}}\\
Here are side-by-side [input $|$ solution] examples: @image0, \dots, @image$\langle k-1\rangle$. In @video0, the left part shows the query input, and the right part is currently a solid neon yellow box.\\[0.5ex]
Edit only the right part (neon yellow area) of @video0, replacing it with the solution for the query input on its left.\\[0.5ex]
Overlay solid purple circles on the 12 major body joints [...] (as above).
\end{promptbox}

\paragraph{Additional results}
On PCK@0.1, \omnione{}'s keypoint error rate ($1 - \text{PCK@0.1}$) in the main setting decreases from $0.62$ ($0$-shot IQM) to $0.49$ at $k = 5$, but without a significant trend over $k \geq 1$ ($\tau = -0.04$, $p = 0.456$). For comparison, dedicated 2D pose estimators such as ViTPose~\citep{xu2022vitpose} and specialized human video models~\citep{wang2026thfm} achieve $<0.10$ keypoint error ($>90\%$ PCK@0.1) on in-the-wild human videos by regressing keypoint heatmaps directly rather than inpainting joint markers in RGB video space. Note also that because the 12 evaluated body joints exclude head keypoints, blurring faces for \seedancetwofive{} (required because the Replicate API safety filter rejected $449/510$ unblurred EMDB~1 requests) leaves the ground-truth joint locations and evaluator untouched while raising completion to $492/510$ ($82/85$ clips at every $k \in \{0, \dots, 5\}$).

\subsection{Video object segmentation}
\label{app:segmentation}

\paragraph{Data}
We evaluate on the validation split of DAVIS~2017~\citep[][released under the \href{https://creativecommons.org/licenses/by-nc/4.0/deed.en}{CC~BY-NC~4.0 license}]{pont20172017}. Full-resolution RGB frames are resized with Lanczos resampling to $1280 \times 720$ ($720\text{p}$) and encoded as MP4 query videos at $24\,\text{fps}$. Following the semi-supervised DAVIS~2017 protocol, the dataset provides the ground-truth segmentation mask of the first frame as context; we resize this first-frame mask with nearest-neighbor interpolation to $1280 \times 720$ and render it in the official PASCAL/DAVIS RGB color palette (black background, distinct color per object) as the reference image for $k=1$. Official $480\text{p}$ palette-indexed ground-truth annotation PNGs serve as evaluation targets.

\paragraph{Evaluation}
Generated videos are paired frame-by-frame by index with the ground truth (up to the shorter of the two sequences), resized (nearest-neighbor) to the $480\text{p}$ resolution of the official annotations, and each pixel is assigned the object ID of the nearest valid DAVIS palette color (background or one of the sequence's foreground objects) by squared RGB Euclidean distance. Predictions are scored with the official DAVIS~2017 implementation of region similarity $\mathcal{J}$ and boundary accuracy $\mathcal{F}$, excluding the first (reference) frame. Consistent with the other computer vision tasks in \cref{fig:all_tasks_4x4}\textbf{A}--\textbf{D}, \cref{fig:segmentation_davis} normalizes the segmentation error $1 - \mathcal{J}\&\mathcal{F}$ frame-by-frame by the no-op (\emph{identity}) baseline; because $1 - \mathcal{J}\&\mathcal{F}$ is bounded in $[0, 1]$, we report the empirical mean across sequences and attempts rather than IQM (\cref{app:eval-statistics}). In raw (unnormalized) mean $1 - \mathcal{J}\&\mathcal{F}$ across sequences and attempts, the \emph{identity} baseline that scores the unmodified input video incurs $91.3\%$ error, and a \emph{first-frame} baseline that repeats the first-frame ground-truth mask for every frame incurs $77.3\%$ ($\approx 84.7\%$ of the identity baseline); both models substantially reduce unnormalized mean error from $0$-shot to $1$-shot (\omnione{}: $83.2\% \rightarrow 61.8\%$; \seedancetwofive{}: $83.5\% \rightarrow 50.8\%$), clearly outperforming the first-frame copy baseline at $k=1$ (while remaining well above dedicated VOS specialists such as Cutie~\citep{cheng2024putting} and SAM~2~\citep{ravi2024sam}, which achieve $1 - \mathcal{J}\&\mathcal{F} \approx 9\%$--$11\%$ on DAVIS~2017 val).

\paragraph{Experiment details}
In the terminology of \cref{tab:design_space}, the video object segmentation experiments in \cref{fig:segmentation_davis} use the \emph{separate--output-only--editing} setting. Because the semi-supervised setting provides only the annotated first frame of the query sequence itself as context, we evaluate $k \in \{0, 1\}$, passing the colorized first-frame mask as a separate reference image (\texttt{@image0}) at $k=1$ and prompting the model to edit the query video into a video segmentation. Because \seedancetwofive{} in editing mode requires input videos of at least $4\,\text{s}$, sequences shorter than $4\,\text{s}$ are freeze-padded on their final frame to $4\,\text{s}$ before inference.

\begin{promptbox}{Prompts for video object segmentation ($k \in \{0, 1\}$)}
\textnormal{\textbf{0-shot ($k = 0$):}}\\
Edit the video to segment the objects. Each object should get its own color. Make the background black.

\medskip
\textnormal{\textbf{1-shot ($k = 1$):}}\\
Edit the video to segment the objects exactly as shown in @image0, while making the background black.
\end{promptbox}

\subsection{Maze solving}
\label{app:maze_solving}
\paragraph{Data}
We reuse three of the four maze splits (5$\times$5, 7$\times$7, and 9$\times$9 rectangular grid mazes, 50 mazes per split) from \citet{wiedemer2026video} with permission. The underlying 2D schematic mazes are generated with the \texttt{maze-dataset} library~\citep[][MIT~License]{mazedataset} via randomized depth-first search and depict black walls, white corridors, a red start circle, and a green goal circle, paired with a ground-truth target image in which the shortest solution path is drawn in blue.

Following visual prompt engineering~\citep{geirhos2026visual}, we transform each 2D schematic maze (padded to a $16{:}9$ aspect ratio) into a photorealistic top-down aerial hedge maze at 1376$\times$768 resolution using \nanobanana{}~\citep[\texttt{gemini-3.1-flash-image-preview};][]{NanoBanana2025}. Conditioned on the schematic layout and a style reference image, \nanobanana{} renders the black walls as dense dark-green hedges, the white corridors as light gravel paths, the red start marker as a person dressed in red wearing a red hat, and the green goal marker as a bright cyan spray-painted circle on the gravel (see prompt below). To guarantee topological fidelity (no hallucinated openings or blocked corridors), we generate multiple seeded candidates per maze and manually curate the final 50 test inputs per split.

We first conduct the zero-shot experiments with \omnione{} on each split, then manually curate eleven correct examples (trimmed to when the person first reaches the goal) as examples for few-shot experiments, thus guaranteeing visual consistency and in-distribution animations. The eleven examples allow us to conduct experiments up to $k=10$ without overlapping queries and examples.

\begin{promptbox}{Prompt for visual prompt engineering of hedge mazes with \nanobanana{}}
Transform the maze layout in $\langle\text{input\_image}\rangle$ into a direct-overhead shot of a hedge maze matching the textures in $\langle\text{reference\_image}\rangle$. Anything that is currently black in $\langle\text{input\_image}\rangle$ should be turned into solid, uninterrupted dark-green hedges as in $\langle\text{reference\_image}\rangle$; white are light gravel paths; a person dressed in red with a red hat takes the place of the red circle; a bright cyan spray-painted circle on the gravel path replaces the green circle. Do not introduce any paths that are not in $\langle\text{input\_image}\rangle$.
\end{promptbox}

\paragraph{Evaluation}
After cropping the query cell from the generated composite grid video and rescaling it to the original query dimensions, our automated evaluator extracts the ground-truth corridor mask, start and goal markers, ground-truth solution path $G$, and corridor width $w$ (estimated as twice the median Euclidean distance transform along the ground-truth path skeleton; $w \approx 100\,\text{px}$, $70\,\text{px}$, and $50\,\text{px}$ for 5$\times$5, 7$\times$7, and 9$\times$9 mazes, respectively) directly from the 2D schematic input and target images. All spatial tolerances are parameterized in units of $w$ so that evaluation remains scale-free across maze sizes.

For each frame of the generated video, the evaluator segments the red runner and trail in HSV color space. To compensate for any camera motion, it detects ORB keypoints on the static background (masking out the red runner and trail), fits a 2D affine transform (translation, rotation, and uniform scale) to the initial query frame via RANSAC, and warps the extracted red mask into the query coordinate frame. On each stabilized frame, the evaluator then measures two quantities in units of $w$: First, walking the ordered skeleton of $G$ from start to goal, each skeleton point is marked as covered if a red pixel lies within $0.5\,w$ of it. The \emph{uncovered gap} is the maximum between the longest contiguous uncovered arc-length along the skeleton and the Euclidean distance from the red mask to the goal center; subtracting a $0.5\,w$ tolerance (which absorbs minor trail flicker) yields the missing solution length $a \in [0, L]$, where $L$ is the initial unbridged length of $G$ beyond the start marker. Second, red pixels outside the ground-truth solution corridor---excluding a $0.5\,w$-radius disc around the runner's body (located at the peak of the red mask's distance transform, so a runner standing near a junction is not penalized)---are grouped into connected components. Each component is scored by its maximum geodesic distance from $G$ (computed via Dijkstra search where stepping onto hedge walls incurs an $8\times$ cost penalty relative to gravel corridors). Components extending beyond a $0.5\,w$ tolerance count as wrong turns, and their depths sum to the total off-path excursion length $b \ge 0$. From the covered solution length $L - a$, missing length $a$, and off-path excursion length $b$, each frame is scored by its 1D length-based intersection-over-union,
\[
\text{Path IoU} = \frac{L - a}{(L - a) + a + b} = \frac{L - a}{L + b} \in [0, 1],
\]
which equals $0$ on an unmodified initial frame ($a = L$, $b = 0$) and $1$ if and only if the trail connects start to goal within tolerance with no wrong turns ($a = 0$, $b = 0$). Iterating through the video chronologically, the evaluator selects the frame maximizing $\text{Path IoU}$ (stopping early on the first frame that achieves $\text{Path IoU} = 1$) and reports the path error ($1 - \text{Path IoU}$). Since the metric is bounded in $[0, 1]$, we use the mean for aggregating results.

\paragraph{Experiment details}
In the terminology of \cref{tab:design_space}, all maze solving experiments use the ``spatial, output-only, editing'' setting. The $k \in \{0, \dots, 5\}$ example videos and a static video of the query input image (placed in the bottom-right cell) are tiled into a 1280$\times$720 ($16{:}9$) spatial grid (1$\times$1 for $k = 0$, 1$\times$2 for $k = 1$, 2$\times$2 for $k \in \{2, 3\}$, and 2$\times$3 for $k \in \{4, 5\}$). The model is prompted to edit the query cell in-place so that the person in red walks through the maze to the cyan circle, stops and spins in place, and leaves a red trail behind. Across all four runs (three splits for \omnione{}, one split for \seedancetwofive{}), we evaluate all 50 test mazes per split across $k \in \{0, \dots, 5\}$ with 2 attempts per query and shot count ($600$ generated videos per run).

\begin{promptbox}{Prompts for hedge mazes ($k \in \{0, \dots, 5\}$)}
\textnormal{\textbf{0-shot ($k = 0$, 1$\times$1 grid):}}\\
@video0 shows the query input.\\[0.5ex]
Edit @video0, replacing it with the solution for the query input.\\[0.5ex]
The person in red makes their way through the maze to the cyan circle, stops there, and spins in place. A red line trails behind the person showing the path they took. Continuous shot, no cuts. Static camera, no pan, no zoom, no dolly.

\medskip
\textnormal{\textbf{1-shot ($k = 1$, 1$\times$2 side-by-side grid):}}\\
@video0 shows 2 parts side-by-side. The left part of @video0 shows an example solution, and the right part of @video0 shows a copy of the query input.\\[0.5ex]
Edit only the right part of @video0, replacing it with the solution for the query input.\\[0.5ex]
The person in red makes their way through the maze to the cyan circle, stops there, and spins in place. A red line trails behind the person showing the path they took. Continuous shot, no cuts. Static camera, no pan, no zoom, no dolly.

\medskip
\textnormal{\textbf{Few-shot ($k \in \{2, 3, 4, 5\}$; $\langle N\rangle = k + 1$ parts in a $\langle\text{grid}\rangle \in \{\texttt{2x2}, \texttt{2x3}\}$ grid):}}\\
@video0 shows $\langle N\rangle$ parts in a $\langle\text{grid}\rangle$ grid. The first $k$ parts of @video0 show example solutions, and the bottom-right part of @video0 shows a copy of the query input.\\[0.5ex]
Edit only the bottom-right part of @video0, replacing it with the solution for the query input.\\[0.5ex]
The person in red makes their way through the maze to the cyan circle, stops there, and spins in place. A red line trails behind the person showing the path they took. Continuous shot, no cuts. Static camera, no pan, no zoom, no dolly.
\end{promptbox}

\begin{figure*}[htbp]
    \centering
    \begin{subfigure}[b]{\twofigwidth\textwidth}
        \centering
        \caption{\opensans 5$\times$5 hedge maze}
        \includegraphics[width=\textwidth]{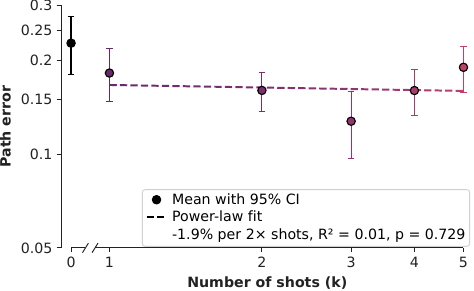}
    \end{subfigure}
    \hfill
    \begin{subfigure}[b]{\twofigwidth\textwidth}
        \centering
        \caption{\opensans 7$\times$7 hedge maze}
        \includegraphics[width=\textwidth]{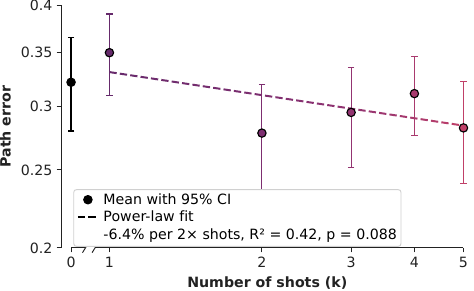}
    \end{subfigure}
    \caption{\textbf{Maze solving on 5$\times$5 and 7$\times$7 hedge mazes} with \omnione{}, complementing the 9$\times$9 results in \cref{fig:maze_solving}. The minimum path error here is also achieved for $k>0$.}
    \label{fig:maze_solving_appendix}
\end{figure*}

\paragraph{Additional results}
Complementing the 9$\times$9 results in \cref{fig:maze_solving}, \cref{fig:maze_solving_appendix} shows results for \omnione{} on 5$\times$5 and 7$\times$7 mazes. On both splits, the minimum mean path error is achieved at a $k>0$, though the results are noisy and there are no clear trends, possibly owing to overall path error already being close to the model's performance ceiling, where errors are dominated not by an inability to plan the correct route, but by general scene-consistency issues (e.g., changing maze layout, duplicating maze runners) unrelated to the specific task.

\subsection{VPCT}
\label{app:vpct}
\paragraph{Data}
We use the visual physics comprehension test (VPCT), published under the MIT License~\citep{camelcase2025vpct}. Each scene is a 2D sketch showing a ball suspended above an arrangement of angled line-obstacles and three target containers at the bottom.

The original dataset contains only the task images, no few-shot examples. To produce these, we implement a Python pipeline that extracts the scene geometry and simulates the ball's 2D rigid-body trajectory at $24\,\text{fps}$ in \texttt{pymunk}~\citep{blomqvist2007pymunk}. The simulation parameters are listed in \cref{tab:vpct_sim_params}. Under this simulation, we find that in three of the 100 original samples (indices \texttt{024}, \texttt{055}, and \texttt{091}), the ball does not land in the ground-truth target container (stopping on a shallow plank in \texttt{024}, overshooting the containers onto the floor in \texttt{055}, and bouncing off the target rim into an adjacent container in \texttt{091}). We exclude these three scenes and use \vpctsketch{} to refer to the 97-sample version of the original dataset with our generated examples.

Following \citet{geirhos2026visual}, we also build a more photorealistic version of the dataset to better estimate \omnione{}'s physical simulation capability, as the model exhibits a strong realism bias. We achieve this by rendering the extracted scene geometry in Blender~\citep{blender} (lines as wooden planks, thickened away from the interaction surface with the ball; the ball as a glossy red ball; the containers as glass bowls; adding a plain wall and floor and lighting that casts subtle shadows), dubbing the resulting version \vpctblender{}.

\begin{figure*}[htbp]
    \centering
    \begin{tikzpicture}[x=1pt, y=1pt, trim left=0pt, trim right=\linewidth]
        \pgfmathsetmacro{\figw}{\linewidth}
        \pgfmathsetmacro{\labelw}{58.0}
        \pgfmathsetmacro{\nframes}{8}
        \pgfmathsetmacro{\cellw}{(\figw - \labelw) / \nframes}
        \pgfmathsetmacro{\cellh}{\cellw * 3.0 / 4.0}
        \pgfmathsetmacro{\vframegap}{2.8}

        \def\framenums{0,15,31,46,62,77,93,108}
        \def\sketchframes{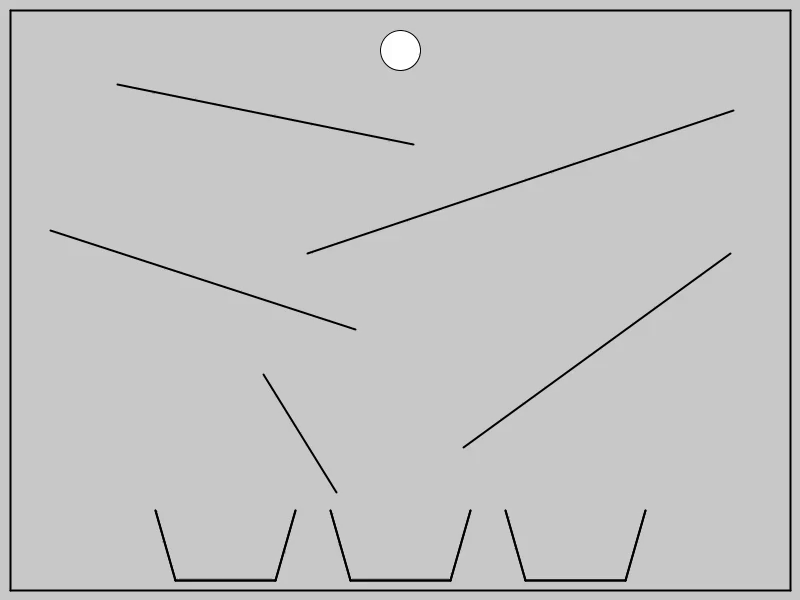,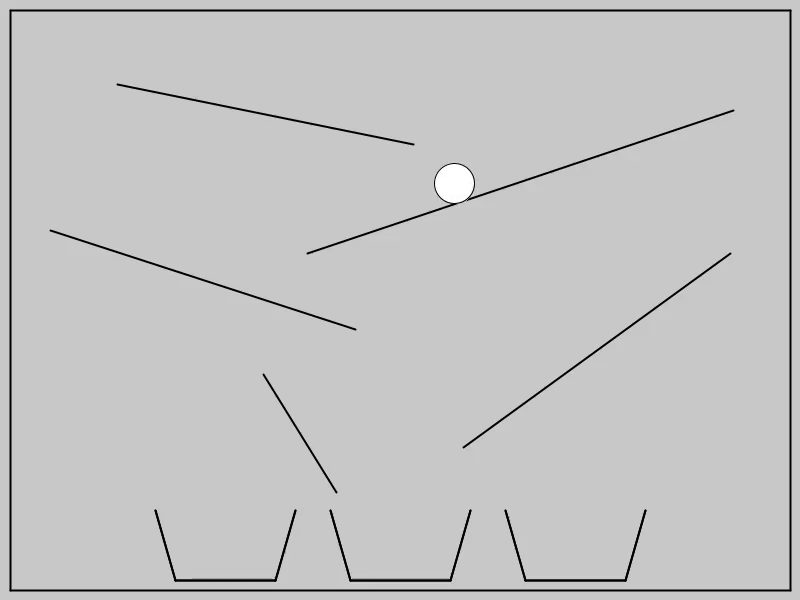,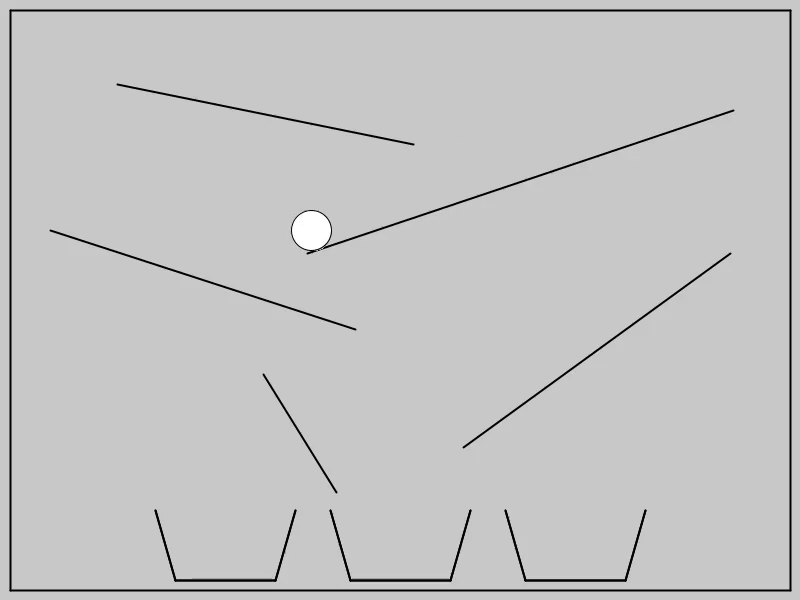,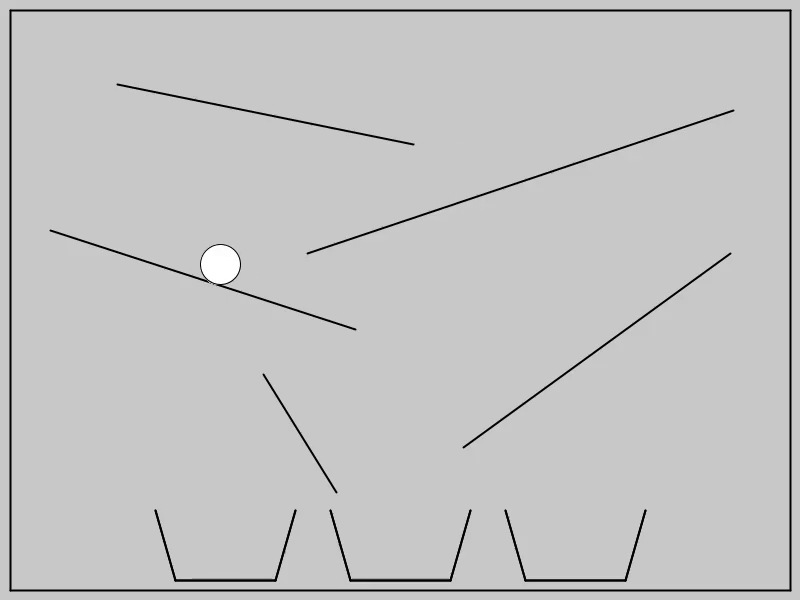,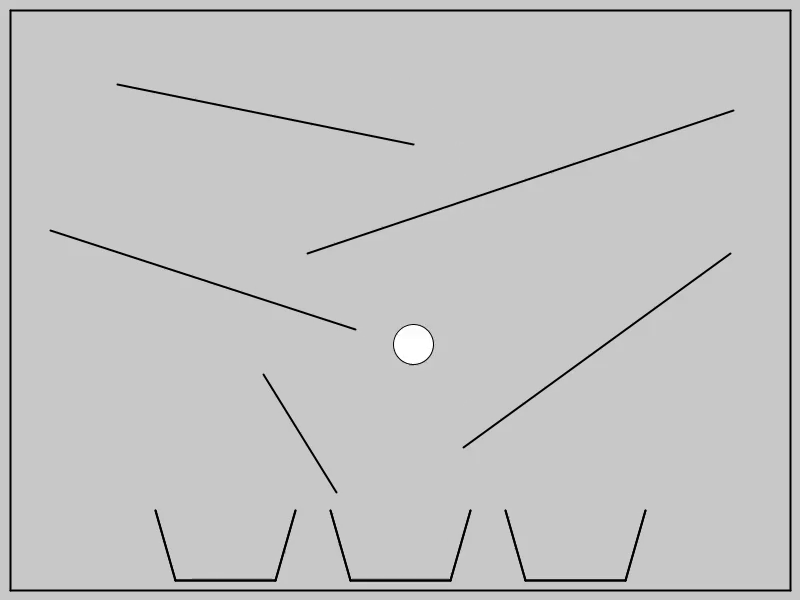,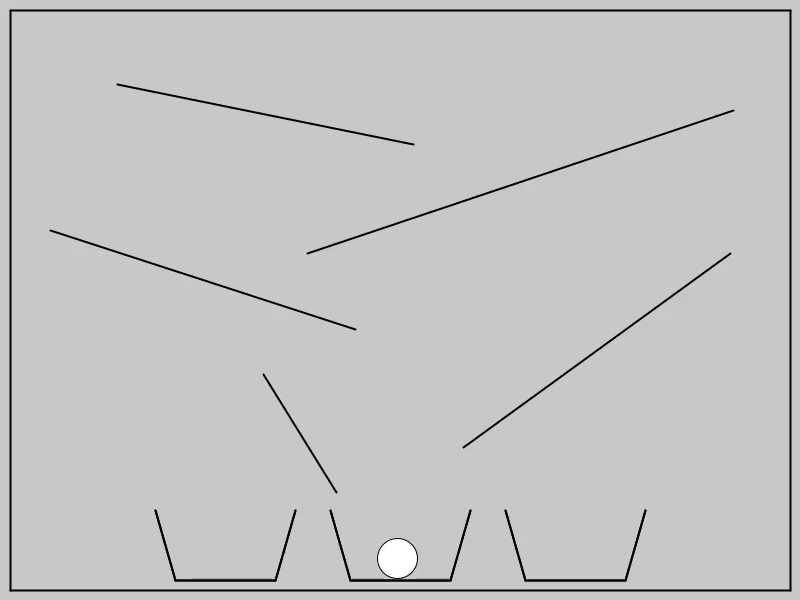,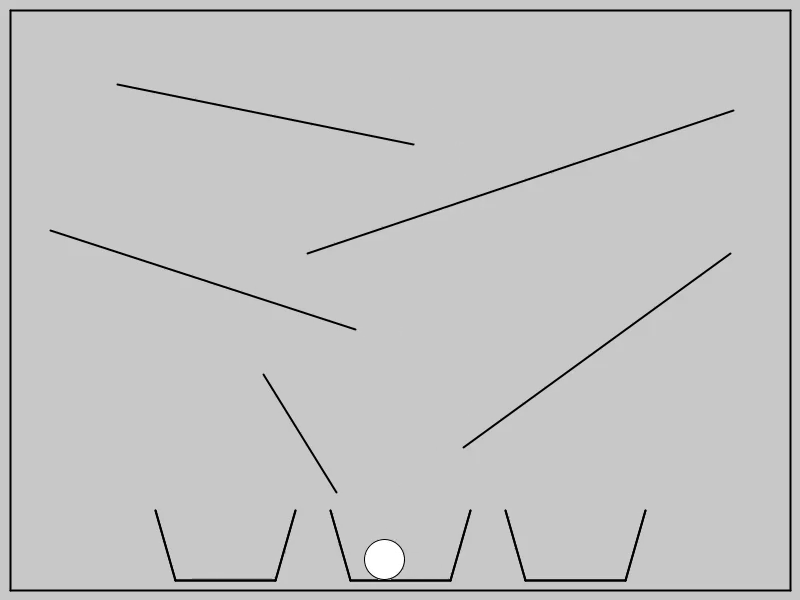,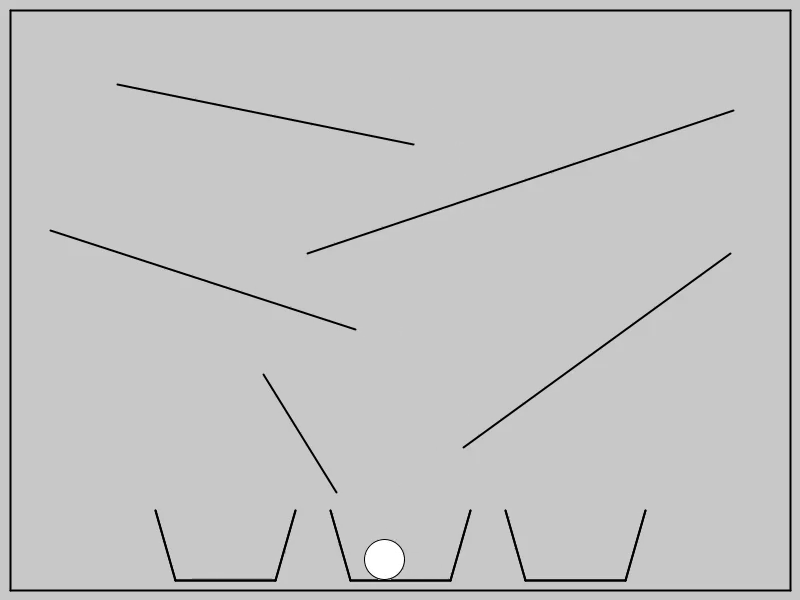}
        \def\blenderframes{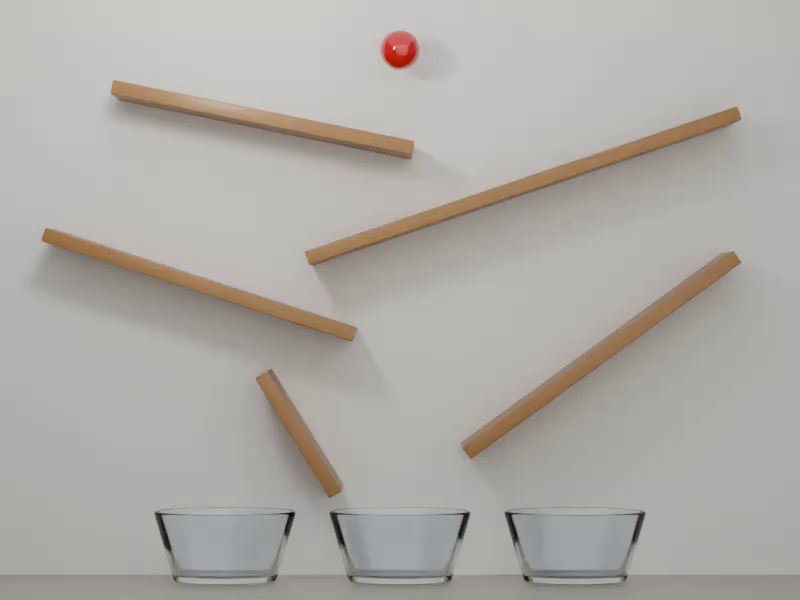,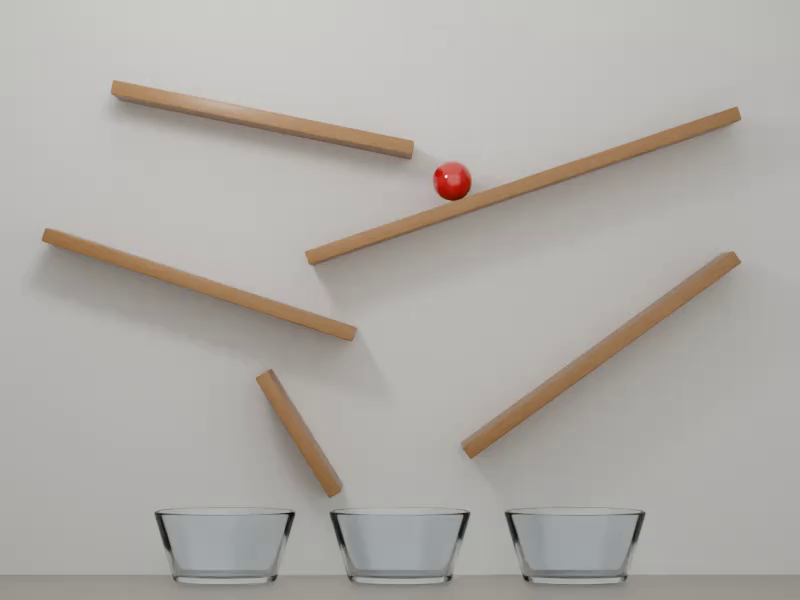,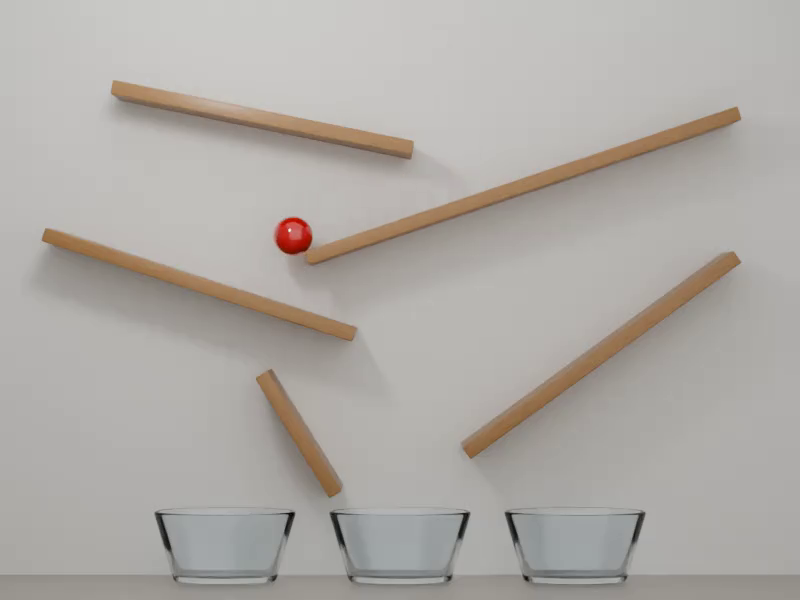,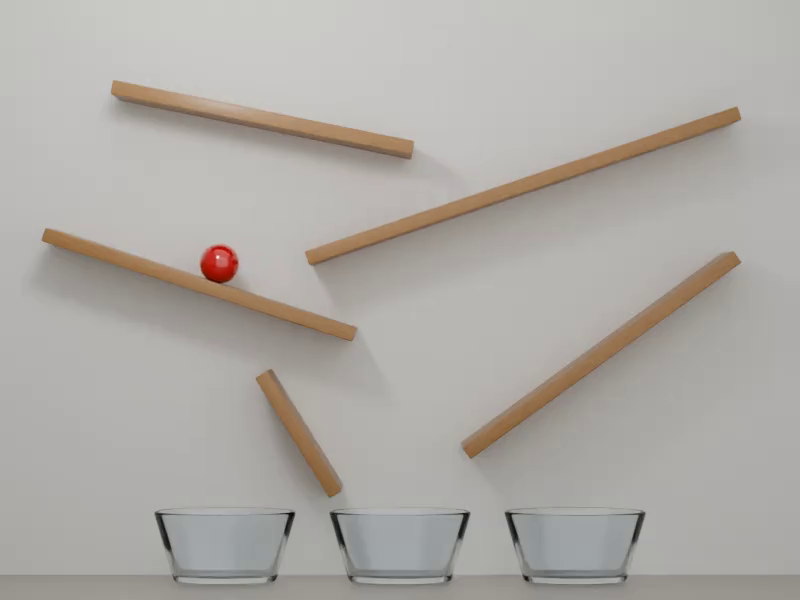,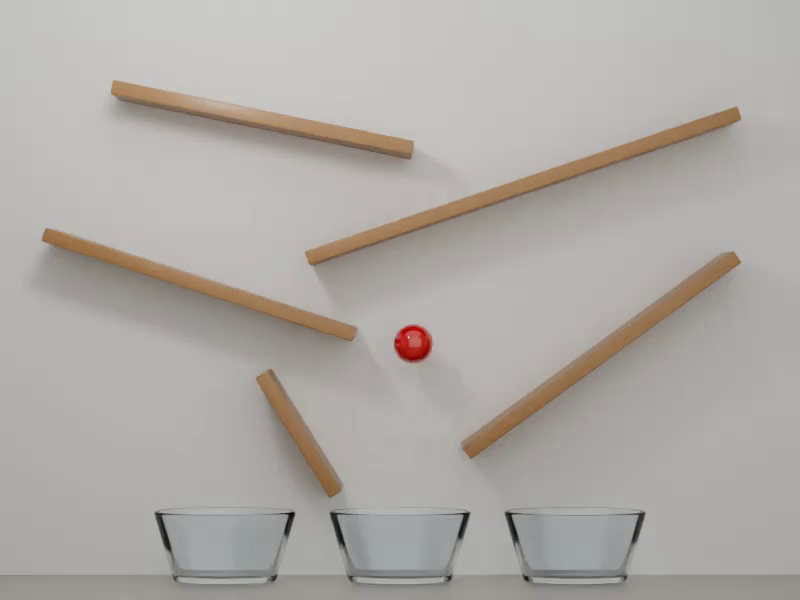,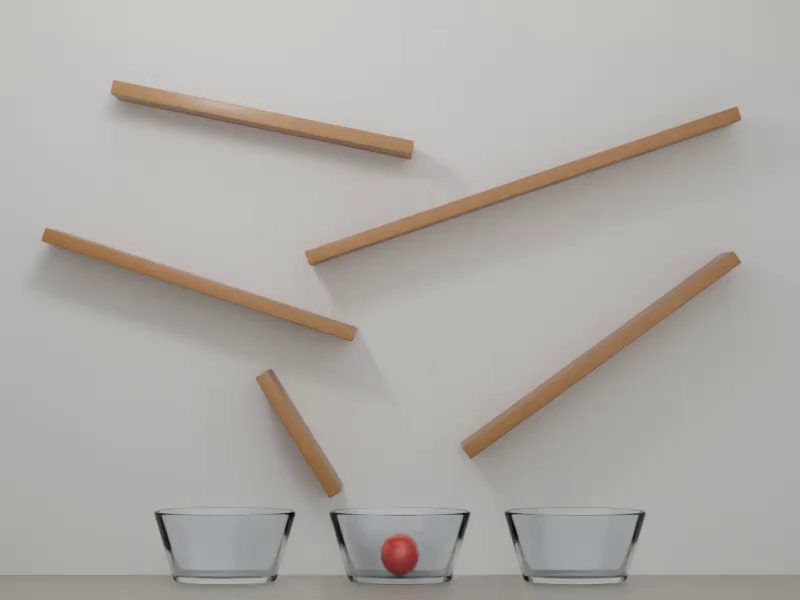,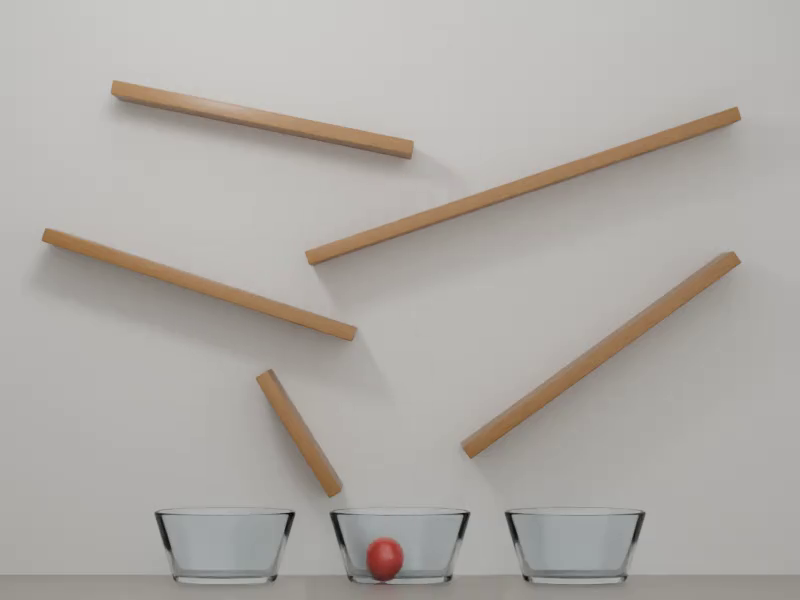,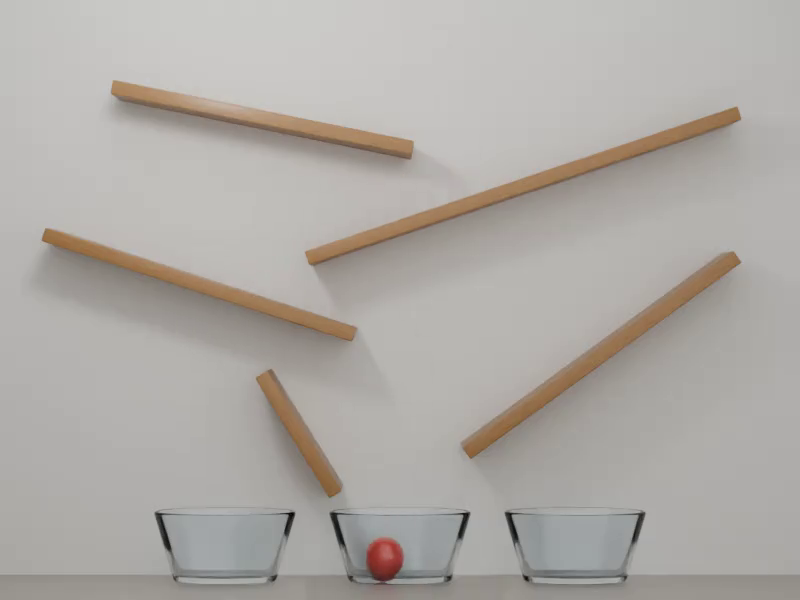}
        \def\genframes{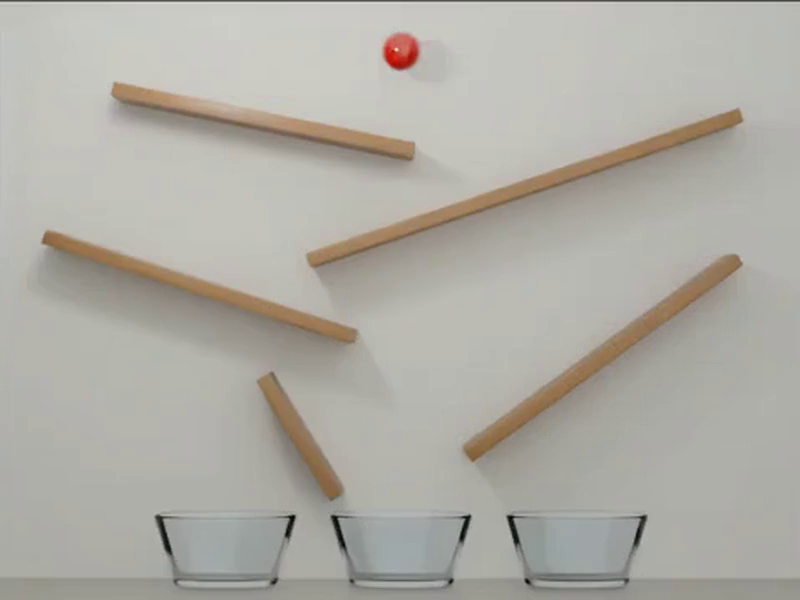,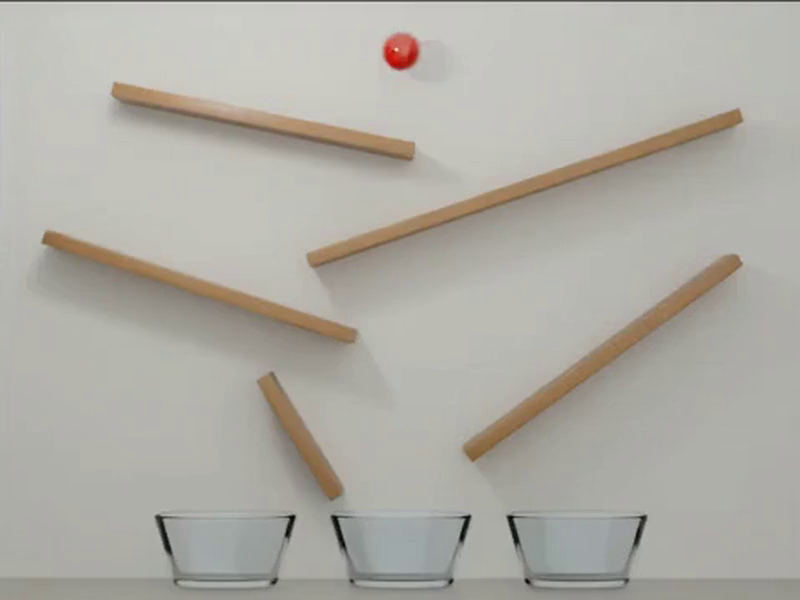,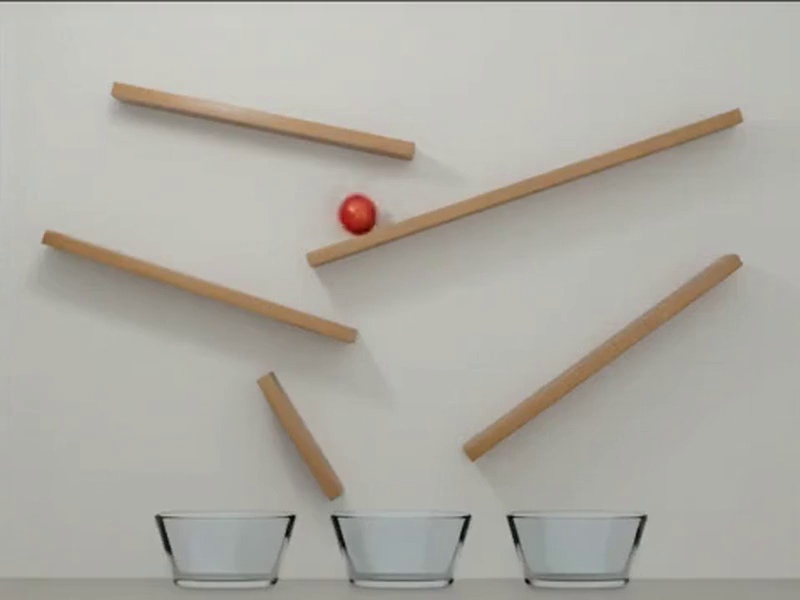,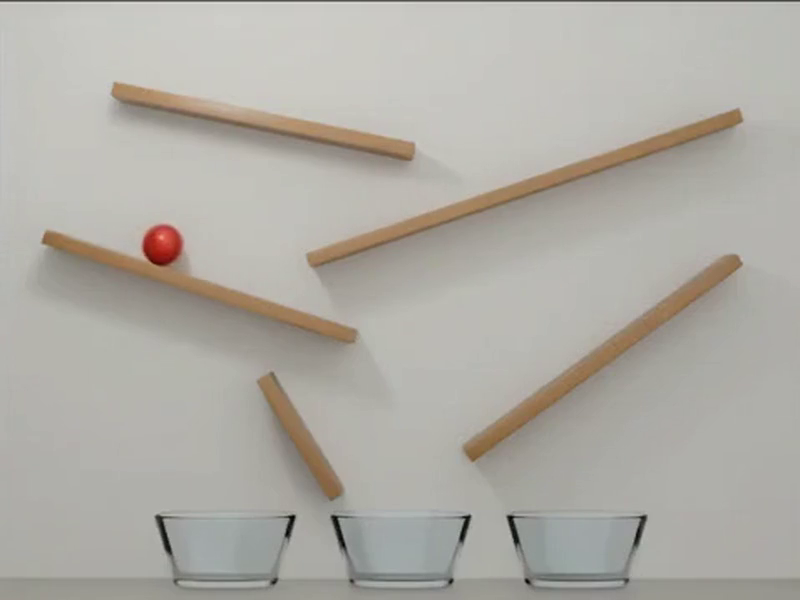,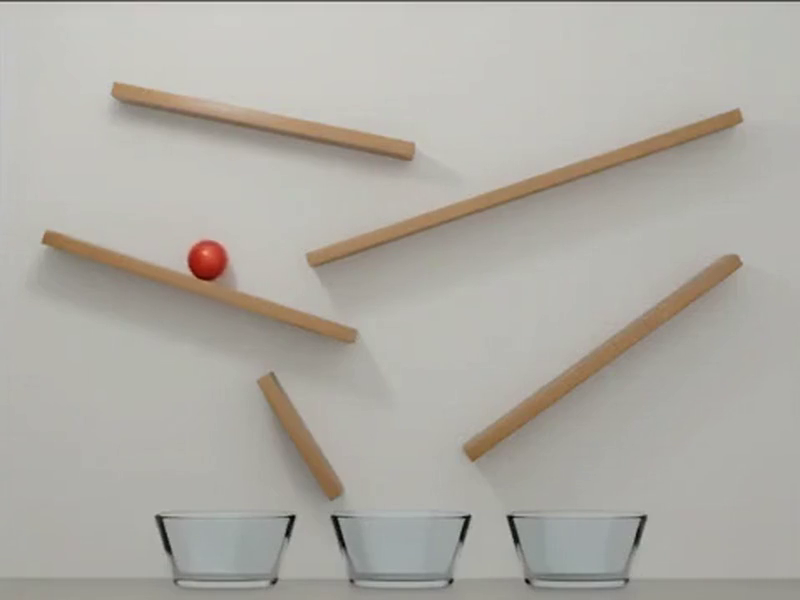,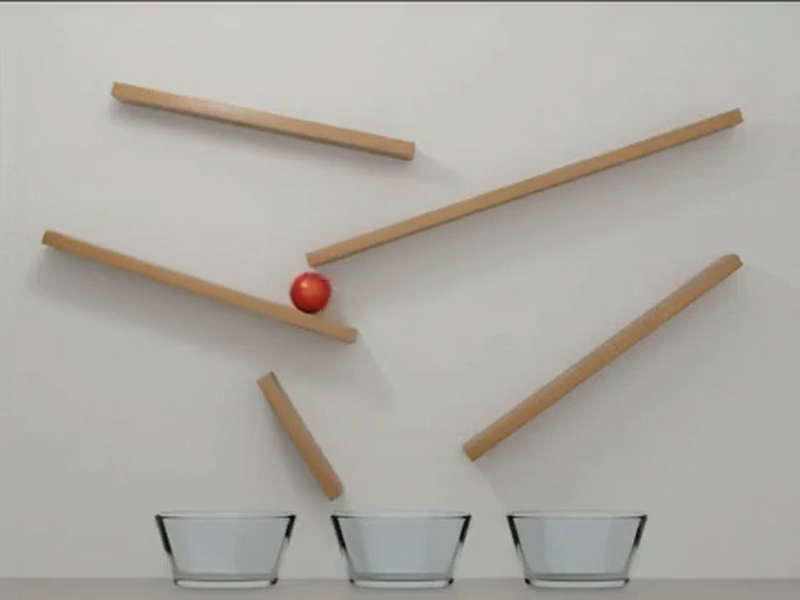,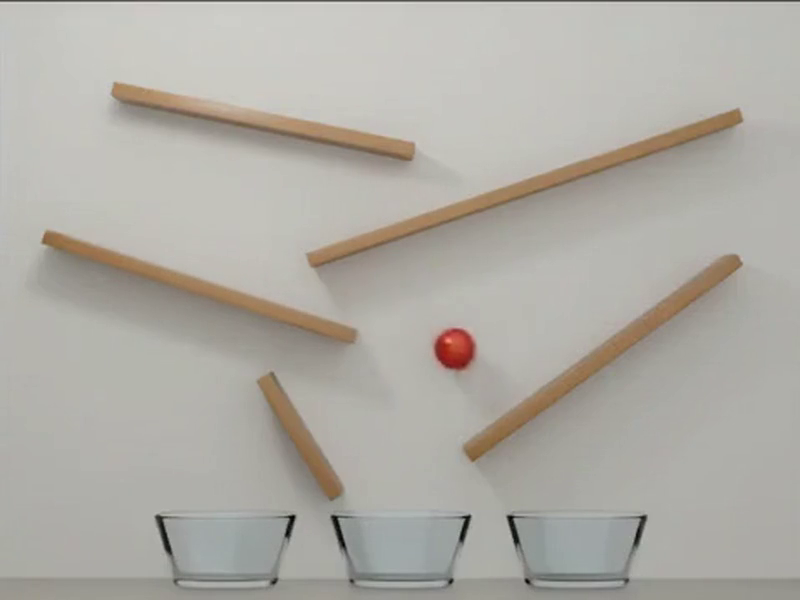,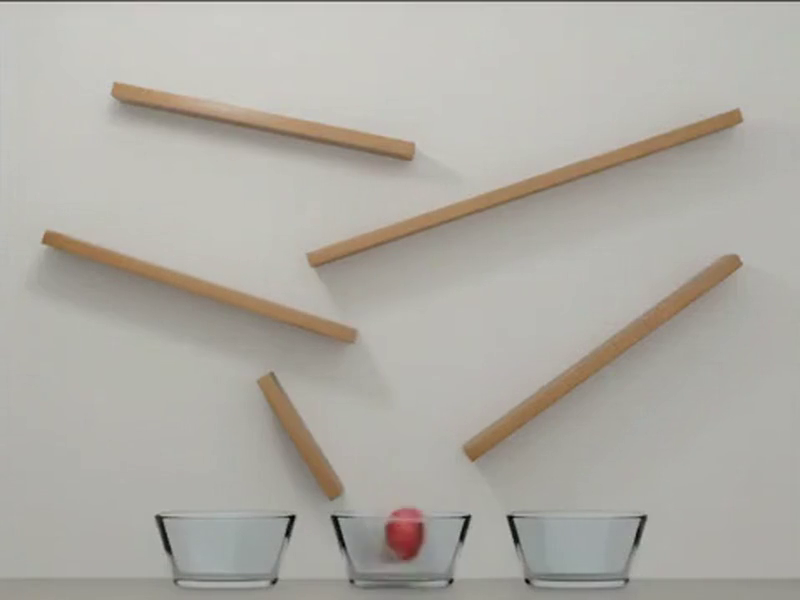}

        \foreach \fnum [count=\col from 0] in \framenums {%
            \node[anchor=south west, inner sep=0pt, outer sep=0pt, font=\opensans\fontsize{8pt}{8pt}\selectfont, text=figtextcolor] at (\labelw + \col*\cellw, 3.5) {Frame \fnum};%
        }%

        \pgfmathsetmacro{\rowoney}{0.0}
        \node[anchor=west, inner sep=0pt, outer sep=0pt, font=\opensans\fontsize{8pt}{8pt}\selectfont, text=figtextcolor] at (0, \rowoney - 0.5*\cellh) {\vpctsketch{}};%
        \foreach \img [count=\col from 0] in \sketchframes {%
            \frameboxsixteennine{\labelw + \col*\cellw, \rowoney}{\cellw pt}{\cellh pt}{\img}%
        }%

        \pgfmathsetmacro{\rowtwoy}{-\cellh - \vframegap}
        \node[anchor=west, inner sep=0pt, outer sep=0pt, font=\opensans\fontsize{8pt}{8pt}\selectfont, text=figtextcolor] at (0, \rowtwoy - 0.5*\cellh) {\vpctblender{}};%
        \foreach \img [count=\col from 0] in \blenderframes {%
            \frameboxsixteennine{\labelw + \col*\cellw, \rowtwoy}{\cellw pt}{\cellh pt}{\img}%
        }%

        \pgfmathsetmacro{\rowthreey}{-2.0 * (\cellh + \vframegap)}
        \node[anchor=west, inner sep=0pt, outer sep=0pt, font=\opensans\fontsize{8pt}{8pt}\selectfont, text=figtextcolor] at (0, \rowthreey - 0.5*\cellh) {\omnione{}};%
        \foreach \img [count=\col from 0] in \genframes {%
            \frameboxsixteennine{\labelw + \col*\cellw, \rowthreey}{\cellw pt}{\cellh pt}{\img}%
        }%
    \end{tikzpicture}
    \caption{\textbf{Example trajectory from \vpctsketch{}, \vpctblender{}, and \omnione{} ($k = 2$)}. Eight evenly-spaced frames from scene \texttt{097} up to the frame where the ball enters the target container in the trajectory generated by \omnione{} (bottom row), alongside frames at the corresponding timestamps from the simulated ground-truth in \vpctsketch{} (top row) and \vpctblender{} (middle row). While \omnione{} arrives at the correct solution, the velocity of the ball throughout the video differs substantially from the ground truth, indicating it has not fully inferred physical properties from the examples.}
    \label{fig:vpct-examples}
\end{figure*}

An example scene from both dataset versions alongside a generated trajectory from \omnione{} is shown in \cref{fig:vpct-examples}.

\begin{table*}[ht]
    \centering
    \small
    \caption{\textbf{2D rigid-body simulation parameters for VPCT in \texttt{pymunk}~\citep{blomqvist2007pymunk}}. All spatial quantities are given in pixel coordinates on the $800 \times 600$ scene canvas ($100\,\text{px} = 1.0\,\text{m}$ in \vpctblender{}).}
    \label{tab:vpct_sim_params}
    \begin{tabular}{ll}
        \toprule
        \textbf{Parameter} & \textbf{Value} \\
        \midrule
        Frame rate \& time step & $24\,\text{fps}$, $10$ substeps/frame ($\Delta t = 1/240\,\text{s}$), $20$ solver iterations \\
        Gravity & $(0, 2400)\,\text{px}/\text{s}^2$ (downward) \\
        Global velocity damping & $0.79\,\text{s}^{-1}$ (in-container per-frame damping: $0.85$) \\
        Ball properties & Mass $1.0$, friction $0.028$, elasticity $0.20$, moment scale $2.3\times$ \\
        Obstacle properties & Friction $1.0$, elasticity $1.0$, bevel radius $0.5\,\text{px}$, polyline tol.\ $2.0\,\text{px}$ \\
        Stopping criterion & Speed $< 5.0\,\text{px}/\text{s}$ in container for $0.5\,\text{s}$ ($+6$ hold frames; max $9.5\,\text{s}$) \\
        \bottomrule
    \end{tabular}
\end{table*}

\paragraph{Evaluation}
Both \vpctsketch{} and \vpctblender{} are evaluated with the same deterministic color-tracking pipeline: After cropping the query cell from the generated grid video back to its original aspect ratio, the evaluator converts each frame to HSV color space and searches the lower $40\%$ of the frame for the ball by thresholding on its color (white for sketch, red for Blender). External contours in the resulting mask are filtered by area and circularity ($\ge 0.5$), and the centroid $(x_\%, y_\%)$ of the largest valid circular contour is recorded in normalized percentage coordinates. Iterating through the video chronologically, the evaluator assigns the first container $\{1, 2, 3\}$ whose horizontal bounds contain $x_\%$ (left: $[0\%, 41.5\%)$, center: $[41.5\%, 58.5\%)$, right: $[58.5\%, 100\%]$) once the ball reaches the container height band ($y_\% \ge 84\%$, with a relaxed vertical threshold on the final frame). If the ball never enters a container, the trial is scored as incorrect. A trial is accurate if the predicted container matches the ground-truth container, and we report the error rate ($1 - \text{accuracy}$). Since the metric is bounded in $[0,1]$, we use the mean for aggregating results.

\paragraph{Experiment details}
In the terminology of \cref{tab:design_space}, all VPCT runs use the ``spatial, output-only, editing'' setting. The $k \in \{0, \dots, 5\}$ example videos and a static video of the query input image (placed in the bottom-right cell) are tiled into a $1280 \times 720$ ($16{:}9$) spatial grid ($1 \times 1$ for $k = 0$, $1 \times 2$ for $k = 1$, $2 \times 2$ for $k \in \{2, 3\}$, and $2 \times 3$ for $k \in \{4, 5\}$), and the model is prompted to edit the query cell in-place to animate the ball rolling into one of the three containers. We average over two attempts per query and shot count for \omnione{} ($1{,}164$ videos per run) and one attempt per query and shot count for \seedancetwofive{} ($582$ videos).

\begin{promptbox}{Prompts for \vpctsketch{} and \vpctblender{} ($k \in \{0, \dots, 5\}$)}
\textnormal{\textbf{0-shot ($k = 0$, $1\times 1$ grid):}}\\
@video0 shows the query input.\\[0.5ex]
Edit @video0, replacing it with the solution for the query input.\\[0.5ex]
The ball falls and rolls down in a physically plausible way, interacting with any obstacles in its path, and ends up in one of the three containers at the bottom. Static shot, no zoom no pan no dolly.

\medskip
\textnormal{\textbf{1-shot ($k = 1$, $1\times 2$ side-by-side grid):}}\\
@video0 shows 2 parts side-by-side. The left part of @video0 shows an example solution, and the right part of @video0 shows a copy of the query input.\\[0.5ex]
Edit only the right part of @video0, replacing it with the solution for the query input.\\[0.5ex]
The ball falls and rolls down in a physically plausible way, interacting with any obstacles in its path, and ends up in one of the three containers at the bottom. Static shot, no zoom no pan no dolly.

\medskip
\textnormal{\textbf{Few-shot ($k \in \{2, 3, 4, 5\}$; $\langle N\rangle = k + 1$ parts in a $\langle\text{grid}\rangle \in \{\texttt{2x2}, \texttt{2x3}\}$ grid):}}\\
@video0 shows $\langle N\rangle$ parts in a $\langle\text{grid}\rangle$ grid. The first $k$ parts of @video0 show example solutions, and the bottom-right part of @video0 shows a copy of the query input.\\[0.5ex]
Edit only the bottom-right part of @video0, replacing it with the solution for the query input.\\[0.5ex]
The ball falls and rolls down in a physically plausible way, interacting with any obstacles in its path, and ends up in one of the three containers at the bottom. Static shot, no zoom no pan no dolly.
\end{promptbox}

\begin{figure*}[htbp]
    \centering
    \begin{subfigure}[b]{\twofigwidth\textwidth}
        \centering
        \caption{\opensans \vpctblender{} with blender examples}
        \includegraphics[width=\textwidth]{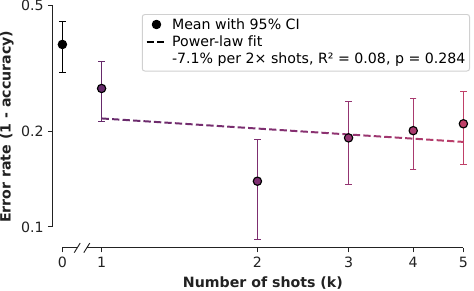}
    \end{subfigure}
    \hfill
    \begin{subfigure}[b]{\twofigwidth\textwidth}
        \centering
        \caption{\opensans \vpctblender{} with sketch examples}
        \includegraphics[width=\textwidth]{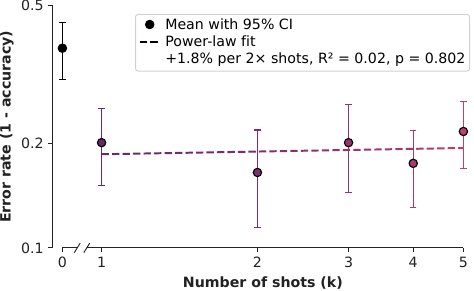}
    \end{subfigure}\par\medskip
    \begin{subfigure}[b]{\twofigwidth\textwidth}
        \centering
        \caption{\opensans \vpctsketch{} with sketch examples}
        \includegraphics[width=\textwidth]{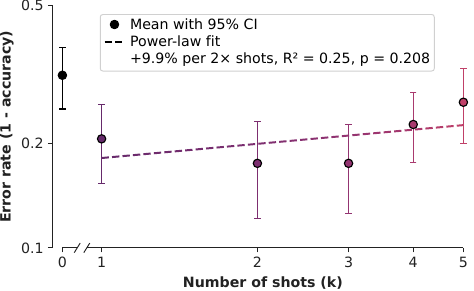}
    \end{subfigure}
    \hfill
    \begin{subfigure}[b]{\twofigwidth\textwidth}
        \centering
        \caption{\opensans \vpctsketch{} with blender examples}
        \includegraphics[width=\textwidth]{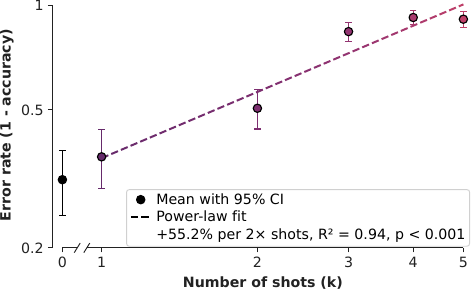}
    \end{subfigure}
    \caption{\textbf{\omnione{} error rate on \vpctsketch{} and \vpctblender{} with sketch or blender examples}. Examples in the same style as the query reduce the error on both \vpctsketch{} and \vpctblender{} (up to $k=2$ or $k=3$). Sketch examples also reduce the error on blender queries, but not the other way around.}
    \label{fig:vpct_appendix}
\end{figure*}

\paragraph{Additional results}
\cref{fig:vpct} shows results for \omnione{} and \seedancetwofive{} on \vpctblender{}. We additionally present results for \omnione{} on \vpctsketch{} in \cref{fig:vpct_appendix}, similarly observing a significant error reduction with multiple examples (here, up to three examples are beneficial). However, this trend does not continue for larger $k$, as the model starts to mix the layout of multiple scenes.

Additionally, we consider whether examples are informative \emph{across dataset versions}~(\cref{fig:vpct_appendix}). While sketch examples are conducive to \omnione{} on \vpctblender{}, leading to a significant error reduction even at $k=1$ (and even more pronounced than with blender examples), blender examples for \vpctsketch{} lead to higher error. A possible explanation is that more blender examples lead \omnione{} to more frequently adopt the blender style, including the red ball, which would lead examples to be scored incorrect, even if the ball motion is otherwise correct.

\subsection{VBVR-Pro}
\label{app:vbvr_pro}
\paragraph{Data}
We evaluate on the 50 in-domain tasks of the VBVR-Pro benchmark~\citep{xu2026vbvr}, which curates and equips tasks from the Very Big Video Reasoning (VBVR) suite~\citep{wang2026very} with deterministic, rule-based verifiers. Both the dataset and evaluation toolkit (\texttt{vbvr-pro-bench}) are released under the Apache~2.0~License. The 50 in-domain tasks span five visual reasoning categories (\emph{abstraction}, \emph{knowledge}, \emph{perception}, \emph{spatiality}, and \emph{transformation}), with five held-out test queries per task ($250$ test queries in total) and separate training exemplars per task from which in-context examples are sampled. Each instance provides an initial query image (\texttt{first\_frame.png}), a text prompt (\texttt{prompt.txt}), a ground-truth solution video (\texttt{ground\_truth.mp4}), and a ground-truth final frame (\texttt{final\_frame.png}). We use the benchmark inputs and targets without visual modification, applying only the temporal and resolution formatting required by each prompting setup and model backend as detailed below.

\paragraph{Evaluation}
Generated videos are scored using the official deterministic, rule-based per-task evaluators from \texttt{vbvr-pro-bench}~\citep{xu2026vbvr}, where each of the 50 tasks has a dedicated computer-vision verifier that tracks task-specific objects, trajectories, or final-frame states against the ground truth. For each sample, the task verifier produces a continuous $\text{task score} \in [0, 1]$ measuring task-specific completion quality (such as trajectory alignment, intermediate state validity, and final-frame accuracy); in the main paper (\cref{fig:vbvr_pro}), we report the ungated continuous \emph{task error} ($1 - \text{task score}$) averaged across evaluated samples at each shot count $k$.

\begin{figure*}[htbp]
    \centering
    \begin{subfigure}[b]{\twofigwidth\textwidth}
        \centering
        \caption{\opensans Failure rate}
        \includegraphics[width=\textwidth]{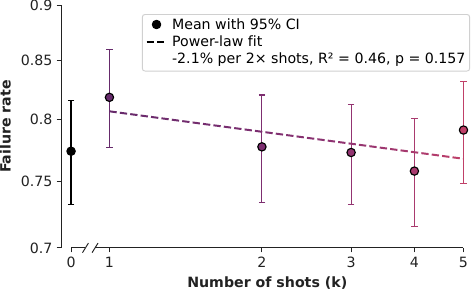}
    \end{subfigure}
    \hfill
    \begin{subfigure}[b]{\twofigwidth\textwidth}
        \centering
        \caption{\opensans Consistency error}
        \includegraphics[width=\textwidth]{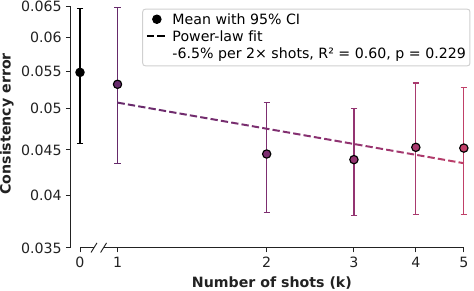}
    \end{subfigure}
    \caption{\textbf{Failure rate and consistency error on VBVR-Pro~\citep{wang2026very, xu2026vbvr} for \omnione{}}. Binary failure rate ($1 - \text{pass rate}$, left; a sample passes when both task score $\ge 0.70$ and consistency score $\ge 0.60$) and continuous consistency error ($1 - \text{consistency score}$, right) across $k \in \{0, \dots, 5\}$ examples on the 50 in-domain tasks, complementing the task error results in \cref{fig:vbvr_pro}.}
    \label{fig:vbvr_pro_appendix}
\end{figure*}

The evaluator additionally computes a first-frame \emph{consistency score} $\in [0, 1]$ measuring structural and perceptual preservation between the initial query image and the first frame of the generated video. The benchmark also defines a consistency gate ($\text{consistency score} \ge 0.60$) that zeros out the task score when violated, and a binary pass criterion requiring both $\text{consistency score} \ge 0.60$ and $\text{task score} \ge 0.70$. To separate task reasoning from first-frame preservation, \cref{fig:vbvr_pro} uses the ungated task score (omitting the consistency gate), while \cref{fig:vbvr_pro_appendix} reports the continuous \emph{consistency error} ($1 - \text{consistency score}$) alongside the gated binary \emph{failure rate} ($1 - \text{pass rate}$).

\paragraph{Experiment details}
In the terminology of \cref{tab:design_space}, the main \omnione{} experiment (\cref{fig:vbvr_pro}, left, and \cref{fig:vbvr_pro_appendix}) uses the ``separate files, output-only, inpainting'' setting across $k \in \{0, \dots, 5\}$ examples on all 50 in-domain tasks ($5$ queries per task, $250$ queries total) with $1$ attempt per query and shot count ($1{,}500$ generations in total). Each example is passed as a separate output-only reference video resampled to $2.0\,\text{s}$. The query input image is converted into a $3.0\,\text{s}$ video to be edited by holding the initial query frame static for the first $1.0\,\text{s}$ followed by $2.0\,\text{s}$ of solid neon-yellow fill frames that the model inpaints with the predicted reasoning rollout.

For \seedancetwofive{} results (\cref{fig:vbvr_pro}, right), we evaluate two queries per task across $k \in \{0, \dots, 5\}$ with $1$ attempt ($600$ generations in total) using the ``separate files, output-only, editing'' setting, where the query input image is converted into a $4.0\,\text{s}$ static video at $24\,\text{fps}$ and each output-only example video is resampled to $2.0\,\text{s}$ at $24\,\text{fps}$ and resized to $1024 \times 1024$ to satisfy backend resolution constraints.

\begin{promptbox}{Prompts for VBVR-Pro (\omnione{}, $k \in \{0, \dots, 5\}$)}
\textnormal{\textbf{0-shot ($k = 0$, inpainting):}}\\
@video0 shows the query input for the first 1s and a solid neon yellow box after that.\\[0.5ex]
Keep seconds 0-1 of @video0 unchanged. Edit seconds 1-3 (neon yellow area) of @video0, replacing it with the solution for the query input shown in seconds 0-1.\\[0.5ex]
$\langle\text{task\_prompt}\rangle$

\medskip
\textnormal{\textbf{1-shot ($k = 1$, inpainting):}}\\
Here is an example of a solved task: @video1. @video0 shows the query input for the first 1s and a solid neon yellow box after that.\\[0.5ex]
Keep seconds 0-1 of @video0 unchanged. Edit seconds 1-3 (neon yellow area) of @video0, replacing it with the solution for the query input shown in seconds 0-1.\\[0.5ex]
$\langle\text{task\_prompt}\rangle$

\medskip
\textnormal{\textbf{Few-shot ($k \in \{2, 3, 4, 5\}$, inpainting):}}\\
Here are examples of solved tasks: @video1, \dots, @video$\langle k\rangle$. @video0 shows the query input for the first 1s and a solid neon yellow box after that.\\[0.5ex]
Keep seconds 0-1 of @video0 unchanged. Edit seconds 1-3 (neon yellow area) of @video0, replacing it with the solution for the query input shown in seconds 0-1.\\[0.5ex]
$\langle\text{task\_prompt}\rangle$
\end{promptbox}

\begin{promptbox}{Prompts for VBVR-Pro (\seedancetwofive{}, $k \in \{0, \dots, 5\}$)}
\textnormal{\textbf{0-shot ($k = 0$, editing):}}\\
@video0 shows the query input.\\[0.5ex]
Edit @video0, replacing it with the solution for the query input.\\[0.5ex]
$\langle\text{task\_prompt}\rangle$

\medskip
\textnormal{\textbf{1-shot ($k = 1$, editing):}}\\
Here is an example of a solved task: @video1. @video0 shows the query input.\\[0.5ex]
Edit @video0, replacing it with the solution for the query input.\\[0.5ex]
$\langle\text{task\_prompt}\rangle$

\medskip
\textnormal{\textbf{Few-shot ($k \in \{2, 3, 4, 5\}$, editing):}}\\
Here are examples of solved tasks: @video1, \dots, @video$\langle k\rangle$. @video0 shows the query input.\\[0.5ex]
Edit @video0, replacing it with the solution for the query input.\\[0.5ex]
$\langle\text{task\_prompt}\rangle$
\end{promptbox}

\paragraph{Additional results}
Complementing the continuous task error in \cref{fig:vbvr_pro}, \cref{fig:vbvr_pro_appendix} reports the binary failure rate ($1 - \text{pass rate}$, left) and continuous consistency error ($1 - \text{consistency score}$, right) for the main \omnione{} experiment across $k \in \{0, \dots, 5\}$ examples.

\subsection{Cellular automata}
\label{app:cellular_automata}
\paragraph{Data}
We construct a synthetic 2D binary cellular automata dataset from scratch. Each sample evolves a binary state grid $S_t \in \{0, 1\}^{9 \times 16}$ (where $1$ denotes an active cell rendered in blue ($\text{RGB} = (30, 100, 220)$), and $0$ denotes an inactive cell rendered in white). Each sample progresses from initial state $S_0$ over $7$ synchronous discrete transitions to produce an $8$-state trajectory $(S_0, S_1, \dots, S_7)$. Out-of-bounds neighbors outside the $9 \times 16$ grid are treated as permanently inactive.

The dataset comprises five deterministic local transition rules designed to span diverse spatial dynamics (rigid translation, monotonic flood-fill, propagating wavefronts, orthogonal parity branching, and mixed orthogonal-birth / diagonal-survival dynamics), summarized in \cref{tab:cellular_automata_rules}. For each rule, we procedurally generate $10$ test queries and $10$ disjoint training examples ($50$ test queries and $50$ example videos in total) using rule-specific initial seed distributions (e.g., sparse interior points or small seed clusters biased toward the upper-left quadrant for directional rules). Every generated trajectory is validated to ensure that no two consecutive states are identical, and that the active cell count remains between $3$ and $95$ (out of $144$ cells) at every step, preventing both extinction and grid saturation.

Each state $S_t$ is rendered onto a $1280 \times 720$ ($16{:}9$) canvas using a $72\,\text{px}$ cell pitch ($1152 \times 648$ active board area centered within a $64\,\text{px}$ left/right and $36\,\text{px}$ top/bottom white outer margin) with $2\,\text{px}$ black grid lines ($\text{RGB} = (20, 20, 20)$) separating $70 \times 70\,\text{px}$ cell interiors. A complete rollout video lasts $2.0\,\text{s}$ ($48$ frames at $24\,\text{fps}$), holding each of the $8$ discrete states $(S_0, \dots, S_7)$ for exactly $6$ frames ($0.25\,\text{s}$). For each test query, the input video is a $2.0\,\text{s}$ ($48$-frame) static video holding the initial state $S_0$.

\begin{table*}[ht]
    \centering
    \small
    \caption{\textbf{Transition rules in the 2D cellular automata benchmark}. For a cell $(r, c)$ at step $t$, $s = S_t[r, c] \in \{0, 1\}$ denotes its current state, $u = S_t[r-1, c]$, $d = S_t[r+1, c]$, $l = S_t[r, c-1]$, $ri = S_t[r, c+1]$, and $ul = S_t[r-1, c-1]$ denote its neighboring states (with out-of-bounds cells fixed to $0$), $N_{\text{ortho}} = u + d + l + ri \in \{0, \dots, 4\}$ is the active orthogonal neighbor count, and $N_{\text{diag}} \in \{0, \dots, 4\}$ is the active diagonal neighbor count.}
    \label{tab:cellular_automata_rules}
    \begin{tabular}{llccc}
        \toprule
        \textbf{Rule} & \textbf{Transition update $S_{t+1}[r, c]$} & \textbf{Support} & \textbf{Monotonic} & \textbf{Cell death} \\
        \midrule
        \texttt{diag\_shift} & $ul = 1$ & $1$ cell & No & Yes \\
        \texttt{se\_growth} & $(s = 1) \lor (u = 1) \lor (l = 1)$ & $3$ cells & Yes & No \\
        \texttt{diag\_wave} & $(u = 1) \lor (l = 1)$ & $2$ cells & No & Yes \\
        \texttt{ortho\_sprout} & $(s = 1) \lor (N_{\text{ortho}} = 1)$ & $5$ cells & Yes & No \\
        \texttt{ortho\_diag\_bridge} & $\bigl(s = 0 \land N_{\text{ortho}} = 1\bigr) \lor \bigl(s = 1 \land N_{\text{diag}} \in \{1, 2\}\bigr)$ & $9$ cells & No & Yes \\
        \bottomrule
    \end{tabular}
\end{table*}

\paragraph{Evaluation}
Given a generated video, our automated evaluator first resizes frames to the canonical $1280 \times 720$ resolution if needed and partitions the sequence into $T = 8$ equal temporal windows corresponding to steps $t \in \{0, \dots, 7\}$ (for a $48$-frame video, $6$ frames per step). To avoid video codec or transition-boundary interpolation artifacts at the exact step boundaries, the evaluator discards the first and last frame of each window and samples only the interior frames (e.g., frames $1, 2, 3, 4$ within each $6$-frame window). Within each interior frame and for each cell $(r, c) \in \{0, \dots, 8\} \times \{0, \dots, 15\}$, the evaluator crops the central $50\%$ of the $70 \times 70\,\text{px}$ cell interior (an $18\,\text{px}$ margin from the $72\,\text{px}$ pitch boundary, completely excluding the black grid lines), computes the mean RGB values $(r, g, b)$, and classifies the cell as active ($1$) if and only if $(r < 180 \land g < 180) \lor (b - r > 40)$. A cell's predicted binary state $\hat{S}_t[r, c] \in \{0, 1\}$ at step $t$ is obtained by majority vote ($\ge 50\%$ active) across the window's interior frames.

At each transition step $t \in \{1, \dots, 7\}$, we compare the predicted binary grid $\hat{S}_t$ against the ground-truth grid $S_t$ via the intersection-over-union of active cells,
\[
\text{IoU}(\hat{S}_t, S_t) = \frac{|\hat{S}_t \cap S_t|}{|\hat{S}_t \cup S_t|} \in [0, 1],
\]
with $\text{IoU} = 1$ if both grids are empty. Averaging the step-wise error over all $7$ evolved steps yields the \emph{rollout error},
\[
\text{Rollout error} = \frac{1}{7} \sum_{t=1}^{7} \bigl(1 - \text{IoU}(\hat{S}_t, S_t)\bigr) \in [0, 1],
\]
which we report in \cref{fig:cellular_automata}. For reference, a no-op baseline that leaves the initial state unchanged across all steps incurs a mean rollout error of $0.840$. Since the metric is bounded in $[0, 1]$, we use the mean for aggregating results across samples and attempts.

\paragraph{Experiment details}
In the terminology of \cref{tab:design_space}, the cellular automata experiments in \cref{fig:cellular_automata} use the ``separate files, output-only, editing'' setting across $k \in \{0, \dots, 5\}$ examples on all $5$ rules ($10$ queries per rule, $50$ test queries in total) with $2$ attempts per query and shot count ($600$ target videos per model; $596$ evaluated videos for \omnione{} after $4$ generation failures at $k = 5$, and $600$ evaluated videos for \seedancetwofive{}). Each example is passed as a separate $2.0\,\text{s}$ ($48$-frame) output-only rollout video from the same transition rule as the query, alongside the static query input video showing $S_0$ ($2.0\,\text{s}$ for \omnione{}; extended to $4.0\,\text{s}$ for \seedancetwofive{} to satisfy backend duration constraints, with evaluation performed on the first $2.0\,\text{s}$ / $48$ frames). Across all shot counts $k \in \{0, \dots, 5\}$ (including $0$-shot), the text prompt explicitly provides the exact natural-language rule description $\langle\text{rule\_description}\rangle$ for the active rule, as shown below.

\begin{promptbox}{Prompts for cellular automata ($k \in \{0, \dots, 5\}$)}
\textnormal{\textbf{0-shot ($k = 0$):}}\\
@video0 shows the query input.\\[0.5ex]
Edit @video0, replacing it with the solution for the query input.\\[0.5ex]
To produce the solution from an input grid: animate the 9x16 cellular automaton grid (blue = active cell, white = inactive cell) from its initial state over 8 discrete states (the initial state followed by 7 step transitions), holding each discrete state for 6 frames (0.25 seconds at 24 fps, for 2.0 seconds total). $\langle\text{rule\_description}\rangle$ Keep the grid position, black grid lines, and white outer border exactly aligned. Do not include any audio in the output.

\medskip
\textnormal{\textbf{1-shot ($k = 1$):}}\\
Here is an example of a solved task: @video0. @video1 shows the query input.\\[0.5ex]
Edit @video1, replacing it with the solution for the query input.\\[0.5ex]
To produce the solution from an input grid: animate the 9x16 cellular automaton grid (blue = active cell, white = inactive cell) from its initial state over 8 discrete states (the initial state followed by 7 step transitions), holding each discrete state for 6 frames (0.25 seconds at 24 fps, for 2.0 seconds total). $\langle\text{rule\_description}\rangle$ Keep the grid position, black grid lines, and white outer border exactly aligned. Do not include any audio in the output.

\medskip
\textnormal{\textbf{Few-shot ($k \in \{2, 3, 4, 5\}$):}}\\
Here are examples of solved tasks: @video0, \dots, @video$\langle k{-}1\rangle$. @video$\langle k\rangle$ shows the query input.\\[0.5ex]
Edit @video$\langle k\rangle$, replacing it with the solution for the query input.\\[0.5ex]
To produce the solution from an input grid: animate the 9x16 cellular automaton grid (blue = active cell, white = inactive cell) from its initial state over 8 discrete states (the initial state followed by 7 step transitions), holding each discrete state for 6 frames (0.25 seconds at 24 fps, for 2.0 seconds total). $\langle\text{rule\_description}\rangle$ Keep the grid position, black grid lines, and white outer border exactly aligned. Do not include any audio in the output.

\medskip
\textnormal{\textbf{Per-rule $\langle\text{rule\_description}\rangle$:}}
\begin{itemize}[leftmargin=1.5em, itemsep=0.25ex, topsep=0.5ex]
    \item \texttt{diag\_shift}: In each discrete step, every blue cell shifts 1 cell Down and 1 cell Right (a cell is blue in the next step if and only if its Up-Left diagonal neighbor was blue in the current step; cells that were previously blue turn white unless fed from Up-Left).
    \item \texttt{se\_growth}: In each discrete step, every existing blue cell stays blue, and a white cell turns blue if and only if its Up neighbor or its Left neighbor was blue in the current step (causing each blue seed to grow continuously Down and Right into a solid triangle).
    \item \texttt{diag\_wave}: In each discrete step, a cell is blue if and only if its Up neighbor or its Left neighbor was blue in the current step (existing blue cells turn white unless re-activated from Up or Left, creating diagonal wavefronts that travel Down and Right).
    \item \texttt{ortho\_sprout}: In each discrete step, every existing blue cell stays blue, and a white cell turns blue if and only if it has EXACTLY 1 blue orthogonal neighbor (Up, Down, Left, or Right; white cells touching 0 or 2 or more blue orthogonal neighbors stay white).
    \item \texttt{ortho\_diag\_bridge}: In each discrete step, a white cell turns blue if and only if it has EXACTLY 1 blue orthogonal neighbor (Up, Down, Left, or Right), while an existing blue cell stays blue if and only if it has 1 or 2 blue diagonal neighbors (otherwise it turns white).
\end{itemize}
\end{promptbox}

\section{Oracle best-of-N and in-context learning}
\label{app:best_of_n_methods}
For \cref{fig:double_scaling}, \omnione{} is evaluated on a total of 5,000 videos, based on the \chromaticscrambleone{} dataset (50 query images $\times$ 10 attempts $\times$ 10 shot settings, $k \in \{1, \dots, 10\}$). Each attempt is scored by the MSE between the final generated frame and the ground-truth target. Oracle Best-of-N is defined as the minimum MSE over $n$ attempts selected using the ground-truth target. For each image and shot count, we estimate its expectation by averaging this minimum over all $\binom{m}{n}$ subsets of $n$ attempts (or over 300 random unique subsets if $\binom{m}{n} > 300$). Aggregation is performed using the mean performance of a group. Error rates are obtained relative to the ``identity baseline'', i.e., not modifying the image at all would score as an error rate of 100\%. This normalization is applied per image: each MSE is divided by the identity-baseline MSE of the same query image before computing Best-of-N and averaging across images.

\begin{figure*}[htbp]
    \centering
    \begin{subfigure}[t]{\threefigwidth\textwidth}
        \caption{\opensans Superadditive scaling:\\In-context examples and Best-of-N}
    \end{subfigure}
    \hfill
    \begin{subfigure}[t]{\threefigwidth\textwidth}
        \caption{\opensans Power law exponents:\\More context $\rightarrow$ steeper power law}
    \end{subfigure}
    \hfill
    \begin{subfigure}[t]{\threefigwidth\textwidth}
        \caption{\opensans Rank statistics:\\Explanation for superadditive scaling}
    \end{subfigure}\par
    \begin{minipage}[b]{\threefigwidth\textwidth}
        \centering
        \includegraphics[width=\textwidth]{figures/double_scaling/funky_transform_v1/297312812_log2_error_rate_vs_log2_best_of_n_by_shots_mean.pdf}
    \end{minipage}
    \hfill
    \begin{minipage}[b]{\threefigwidth\textwidth}
        \centering
        \includegraphics[width=\textwidth]{figures/double_scaling/funky_transform_v1/297312812_error_rate_power_law_exponent_vs_log2_shots_mean.pdf}
    \end{minipage}
    \vspace{0.2cm}
    \hfill
    \begin{minipage}[b]{\threefigwidth\textwidth}
        \centering
        \includegraphics[width=\textwidth]{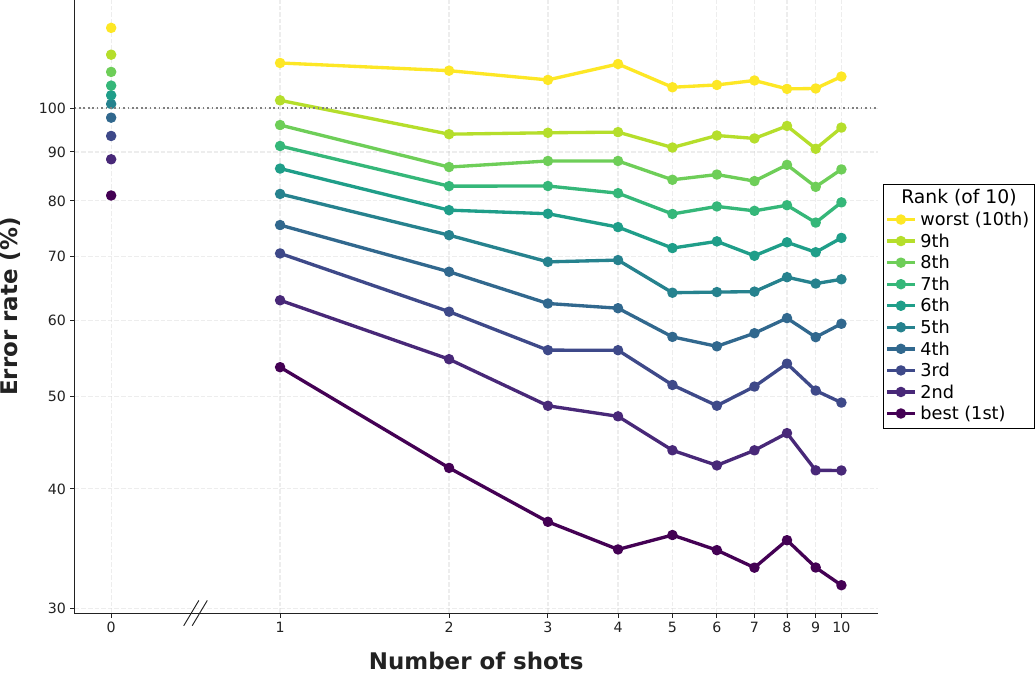}
    \end{minipage}
    \begin{minipage}[b]{\threefigwidth\textwidth}
        \centering
        \includegraphics[width=\textwidth]{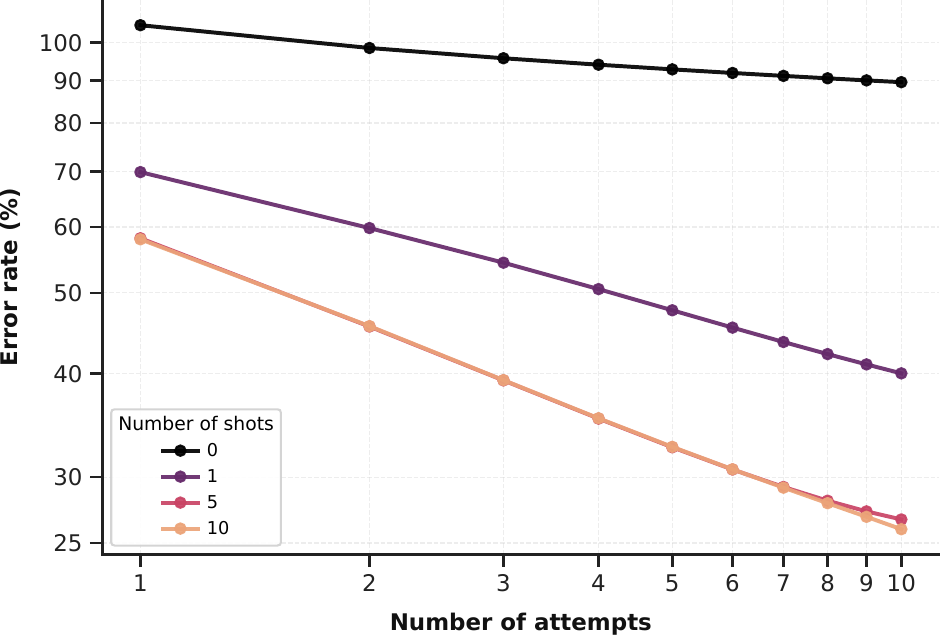}
    \end{minipage}
    \hfill
    \begin{minipage}[b]{\threefigwidth\textwidth}
        \centering
        \includegraphics[width=\textwidth]{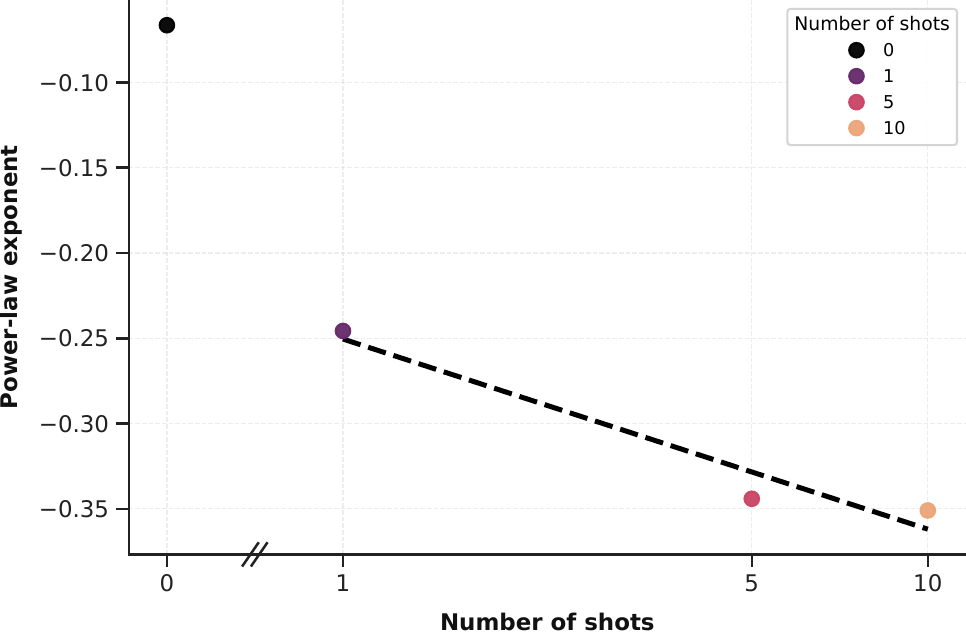}
    \end{minipage}
    \hfill
    \vspace{0.2cm}
    \begin{minipage}[b]{\threefigwidth\textwidth}
        \centering
        \includegraphics[width=\textwidth]{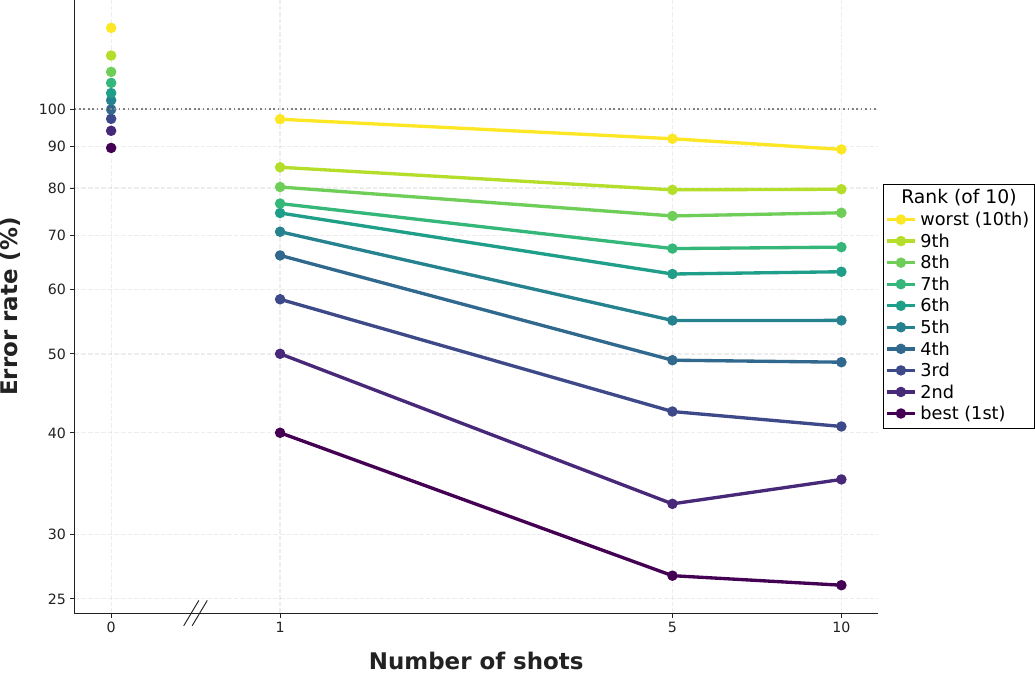}
    \end{minipage}
    \begin{minipage}[b]{\threefigwidth\textwidth}
        \centering
        \includegraphics[width=\textwidth]{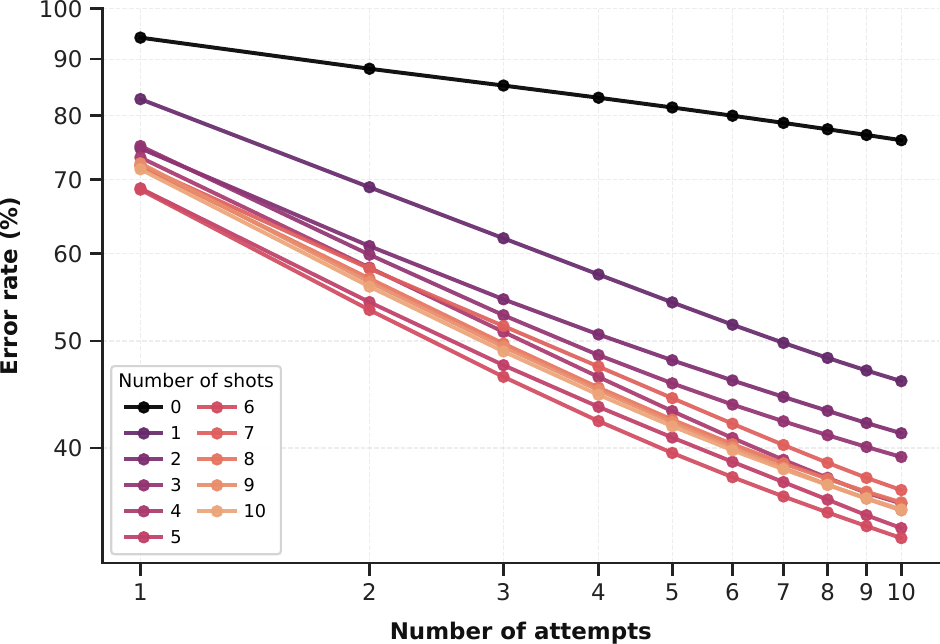}
    \end{minipage}
    \hfill
    \begin{minipage}[b]{\threefigwidth\textwidth}
        \centering
        \includegraphics[width=\textwidth]{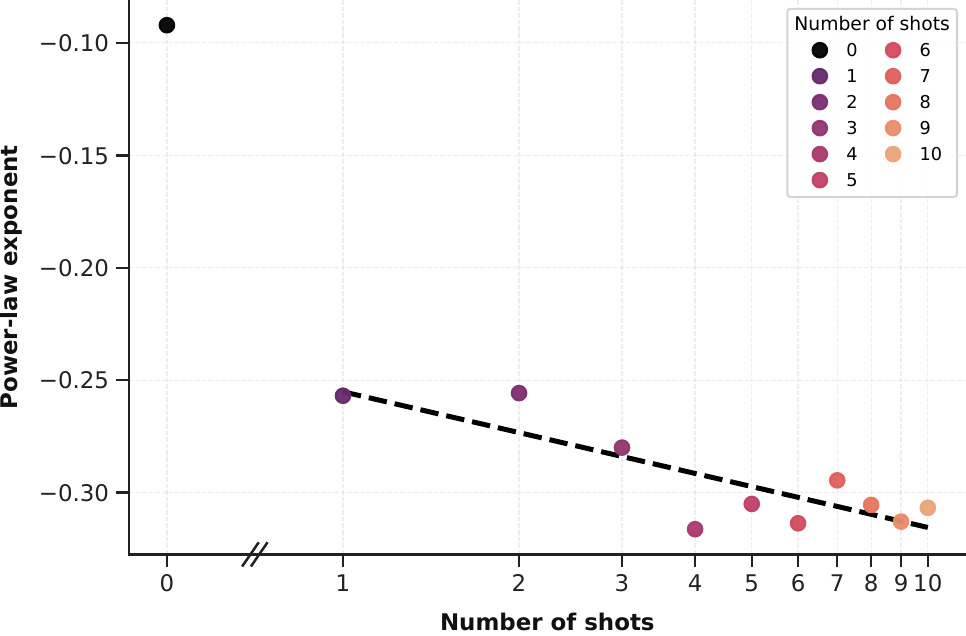}
    \end{minipage}
    \hfill
    \begin{minipage}[b]{\threefigwidth\textwidth}
        \centering
        \includegraphics[width=\textwidth]{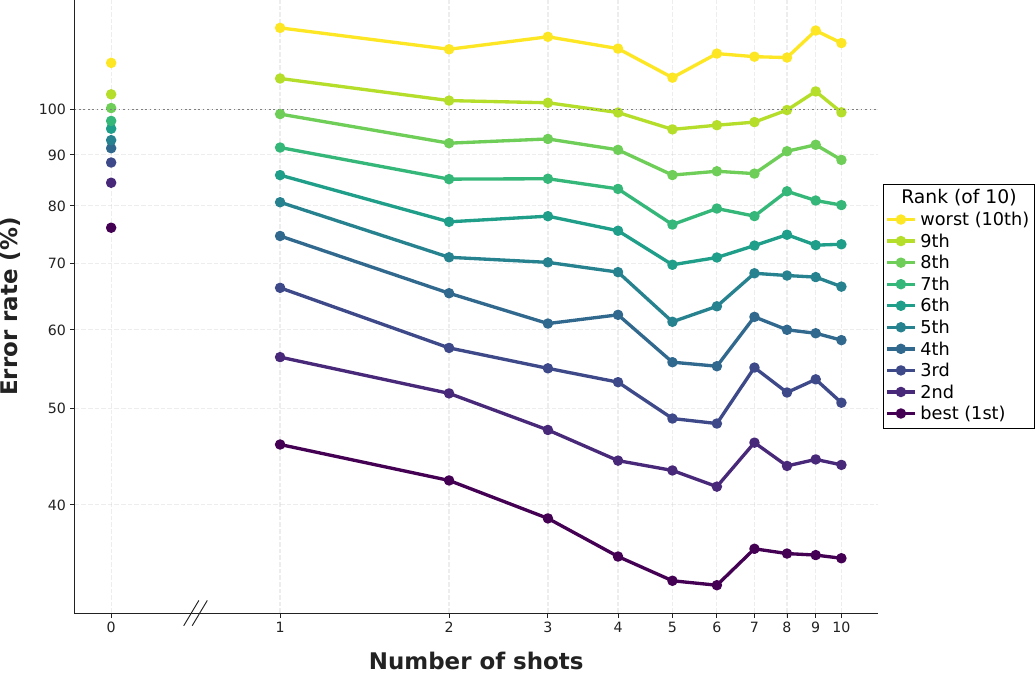}
    \end{minipage}
    \caption{\textbf{Combining in-context learning with test-time scaling (Best-of-N)}. \emph{Top row:} \chromaticscrambleone{}, \omnione{}. \emph{Middle row:} \chromaticscrambleone{}, \seedancetwofive{} on $k\in \{0,1,5,10\}$ shots due to the high cost of Seedance inference. \emph{Bottom row:} \chromaticscrambletwo{}, \omnione{}. \emph{Left and middle columns:} Same as \cref{fig:double_scaling}. \emph{Right column:} Rank statistics across the 10 attempts per query explain the superadditive scaling: while the error rate of the \emph{worst} attempt stays roughly constant across context sizes, the \emph{best} attempts improve substantially with more shots, resulting in superadditive scaling. Note that error rates, in this column, are relative to the per-query identity baseline, i.e., relative to the error achieved by not modifying the query input at all. An error rate larger than 100\% therefore means applying a transformation to the input that is worse than not changing it at all.}
    \label{fig:double_scaling_big_panel}
\end{figure*}

\begin{figure*}[ht]
    \centering
    \begin{subfigure}[t]{0.4040\textwidth}
        \caption{\opensans \omnione{}:\\\chromaticscrambleone{}}
    \end{subfigure}
    \hfill
    \begin{subfigure}[t]{0.1670\textwidth}
        \caption{\opensans \seedancetwofive{}:\\\chromaticscrambleone{}}
    \end{subfigure}
    \hfill
    \begin{subfigure}[t]{0.4040\textwidth}
        \caption{\opensans \omnione{}:\\\chromaticscrambletwo{}}
    \end{subfigure}\par
    \begin{minipage}[b]{0.4040\textwidth}
        \centering
        \includegraphics[width=\textwidth]{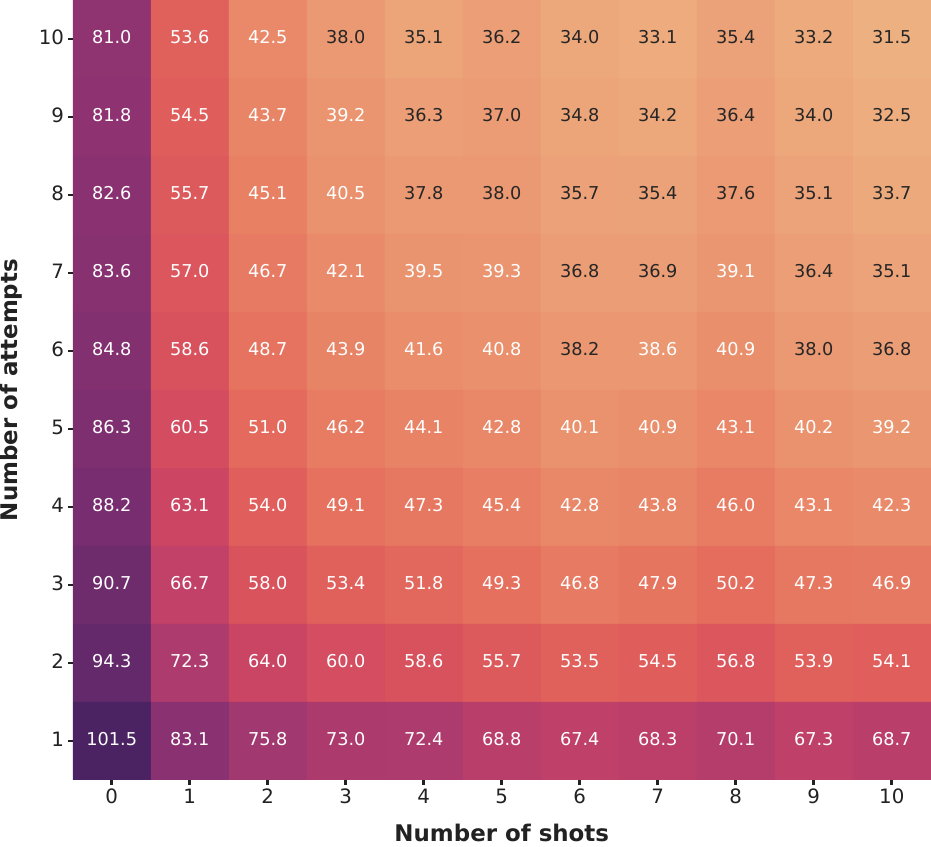}
    \end{minipage}
    \hfill
    \begin{minipage}[b]{0.1670\textwidth}
        \centering
        \includegraphics[width=\textwidth]{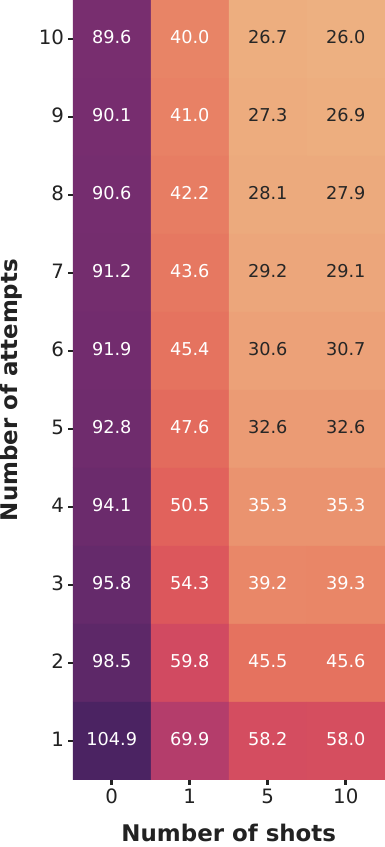}
    \end{minipage}
    \hfill
    \begin{minipage}[b]{0.4040\textwidth}
        \centering
        \includegraphics[width=\textwidth]{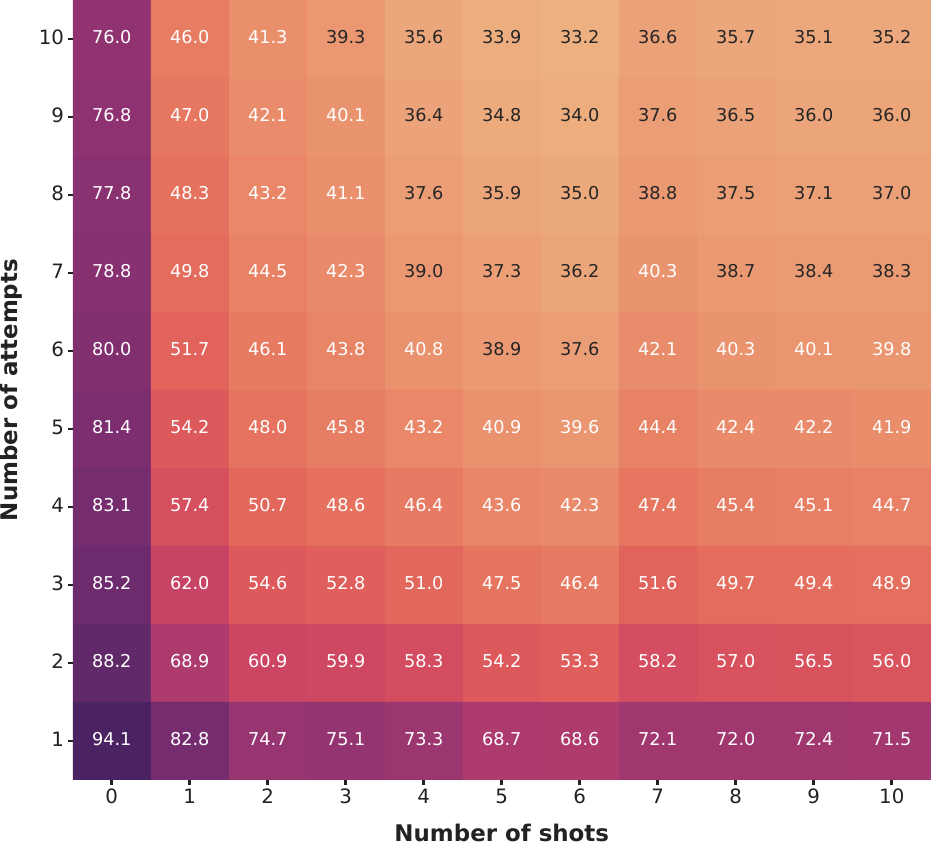}
    \end{minipage}
    \caption{\textbf{Detailed data from \cref{fig:double_scaling_big_panel} (left column), visualized as heatmaps}.}
    \label{fig:double_scaling_heatmap}
\end{figure*}

For \cref{fig:double_scaling_big_panel}, the same experiment is performed not just on \omnione{} and the \chromaticscrambleone{} dataset, but also \seedancetwofive{} and another dataset, \chromaticscrambletwo{}. Since the experiment is very expensive, we evaluate Seedance only on $[0,1,5,10]$ shots. This resulted in 5.5K videos per dataset for Omni ($\approx$ 3K USD) and 2K videos for Seedance (costing $\approx$ 8K USD).

Additionally, an analysis of the rank statistics (rightmost column of \cref{fig:double_scaling_big_panel}) is performed; it shows the error rate of the $r$-th best of the 10 attempts per image as a function of the number of shots. This analysis sheds light on the superadditive scaling we observe: while the error rate of the \emph{worst} attempt stays roughly constant across context sizes, the \emph{best} attempts improve substantially with more shots, resulting in superadditive scaling.

\end{document}